\documentclass{article}

\usepackage[preprint, nonatbib]{neurips_2022}

\usepackage[utf8]{inputenc} % allow utf-8 input
\usepackage[T1]{fontenc}    % use 8-bit T1 fonts
\usepackage{hyperref}       % hyperlinks
\usepackage{url}            % simple URL typesetting
\usepackage{booktabs}       % professional-quality tables
\usepackage{amsfonts}       % blackboard math symbols
\usepackage{nicefrac}       % compact symbols for 1/2, etc.
\usepackage{microtype}      % microtypography
\usepackage{xcolor}         % colors

\usepackage{amsmath}
\usepackage{amssymb}
\usepackage{mathtools}
\usepackage{amsthm}

\usepackage{subfigure}
\usepackage{multirow}
\usepackage{threeparttable}
\usepackage{bm}
\usepackage{wrapfig}
\usepackage{multicol}
\usepackage{algorithm}  
\usepackage{algorithmicx}  
\usepackage{algpseudocode}

\theoremstyle{plain}
\newtheorem{theorem}{Theorem}[section]

\newtheorem{lemma}[theorem]{Lemma}

\theoremstyle{definition}
\newtheorem{definition}[theorem]{Definition}

\theoremstyle{remark}

\newcommand{\bx}{\bm{x}}

\newcommand{\inSample}{n_1}
\newcommand{\outSample}{n_2}
\newcommand{\inNoisePdf}{v_1}
\newcommand{\outNoisePdf}{v_2}

\newcommand{\firstOrderBound}{C}

\title{Anomaly Detection with Test Time Augmentation and Consistency Evaluation}

\author{%
  Haowei He\thanks{Equal contribution.} \\
  IIIS\\
  Tsinghua University\\
  \texttt{hhw19@mails.tsinghua.edu.cn} \\
   \And
   Jiaye Teng\footnotemark[1] \\
   IIIS \\
   Tsinghua University \\
   \texttt{tjy20@mails.tsinghua.edu.cn} \\
   \AND
   Yang Yuan \\
   IIIS \\
   Tsinghua University \\
   \texttt{yuanyang@mail.tsinghua.edu.cn} \\
}

\begin{document}

\maketitle

\begin{abstract}
Deep neural networks are known to be vulnerable to unseen data: they may wrongly assign high confidence scores to out-distribution samples. Recent works try to solve the problem using representation learning methods and specific metrics. In this paper, we propose a simple, yet effective post-hoc anomaly detection algorithm named Test Time Augmentation Anomaly Detection~(TTA-AD), inspired by a novel observation. Specifically, we observe that in-distribution data enjoy more consistent predictions for its original and augmented versions on a trained network than out-distribution data, which separates in-distribution and out-distribution samples. 
Experiments on various high-resolution image benchmark datasets demonstrate that TTA-AD achieves comparable or better detection performance under dataset-vs-dataset anomaly detection settings with a $60\%\sim90\%$ running time reduction of existing classifier-based algorithms.
We provide empirical verification that the key to TTA-AD lies in the remaining classes between augmented features, 
which has long been partially ignored by previous works. Additionally, we use RUNS as a surrogate to analyze our algorithm theoretically. 
\end{abstract}

\section{Introduction}
\label{sec:intro}

Recently, deep neural networks have shown substantial flexibility and valuable practicality in various tasks~\cite{densenet, dl2015}. However, when deploying deep learning in reality, one of the most concerning issues is that deep models are known to be overconfident when exposed to unseen data~\cite{DBLP:journals/corr/AmodeiOSCSM16, DBLP:journals/corr/GoodfellowSS14}. That is, although the deep models generalize well on unseen datasets drawn from the same distribution~(i.e., test data), it incorrectly assigns high confidence to unseen data drawn from another distribution~(i.e., out-distribution data). 
To solve the problem, anomaly detection~\cite{DBLP:books/sp/Aggarwal2013} aims to separate unseen data from training data. 

%\hhw{Specifically, \cite{oodbaseline} observed that a well trained classifier will assign higher softmax probability to in-distribution data than out-distribution data. And a series of following works\cite{ODIN, Ma, generalizedodin} follow this idea and use various algorithms to improve the results. In this work, we find another interesting phenomenon that despite a well trained model will make fake prediction towards out-distribution data, it will make a less robust(consistent) prediction for two augmentations of the input data. Note: maybe we can put this paragraph to Abstract.}

When testing deep neural networks, test time augmentation~(TTA) utilizes the property that a well-trained classifier should have a low variance in predictions across augmentations on most data~\cite{DBLP:journals/jbd/ShortenK19, DBLP:journals/corr/abs-2011-11156}. 
However, this phenomenon is only explored and verified on data drawn from the training distribution~(in-distribution). It is still unclear whether this property generalizes to other different distributions~(out-distribution). 
%	, which may possibly induce people to an illusion that \emph{a well-trained classifier is robust towards augmentations to ALL data.} 
%	\hhw{Or: It is unclear whether this property~(augmentation robustness) also holds for data drawn from other distributions.~(generalizes to other different distributions)}

\begin{wrapfigure}{r}{0.48\textwidth}
	\centering
%	\vspace{-19.1px}
	\includegraphics[width=.95\linewidth]{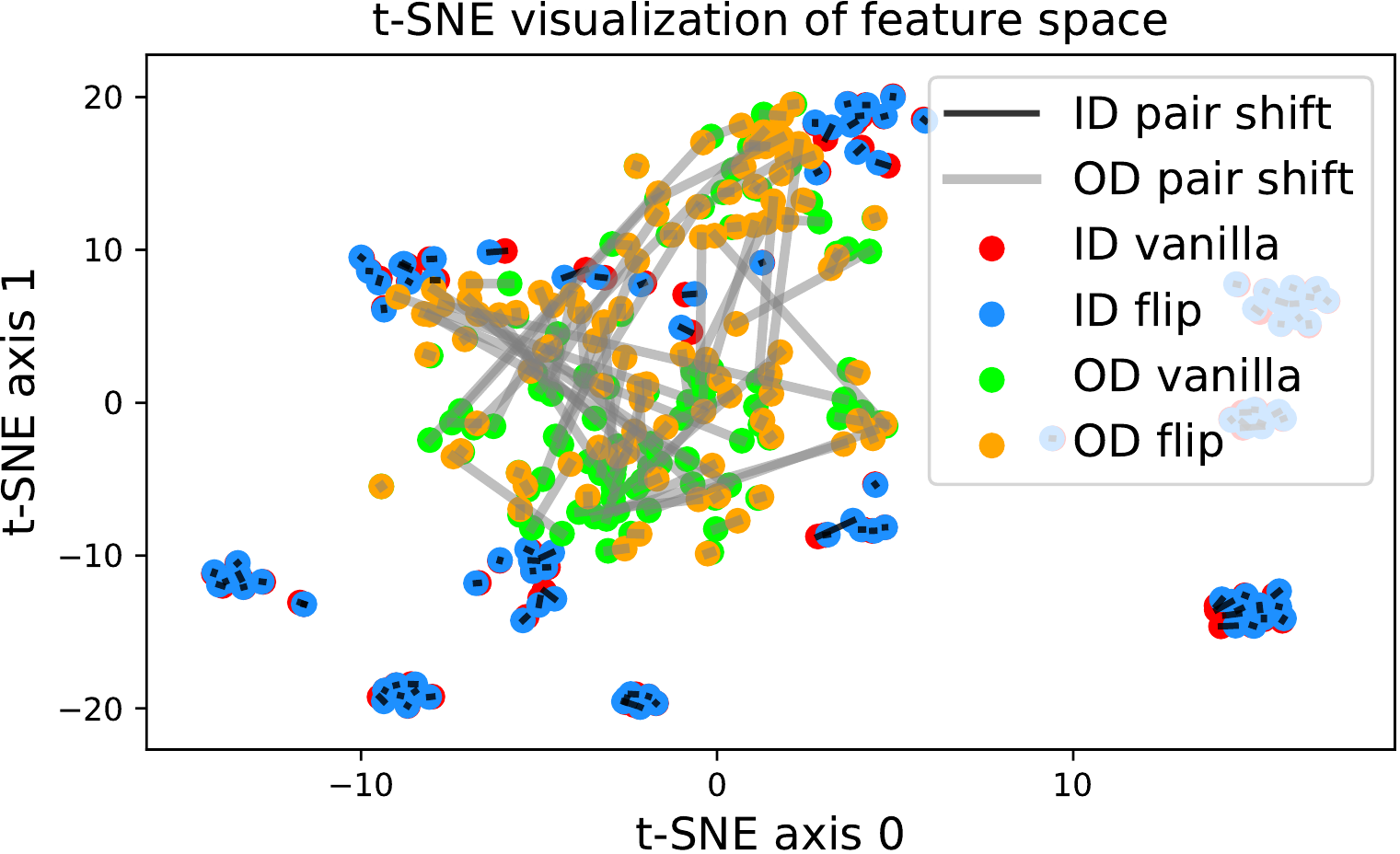}
	\caption{T-SNE feature space visualization. We randomly pick 200 CIFAR-10~(in-distribution, ID) and SVHN samples~(out-distribution, OD). For each sample, we compute its vanilla feature and the feature of its augmented version. Here we use flip. We draw lines between a pair of the vanilla feature point and the augmented feature point. For each pair, the longer the line segment, the greater the feature difference. } 
	\label{fig:feature_shift}
	\vspace{-20px}
\end{wrapfigure}

In fact, the answer is No since the model does not see any augmentations of the out-distribution samples in the training phase.
To verify, we use t-SNE~\cite{tsne} to visualize the projected feature space of a ResNet-34~\cite{resnet} supervisedly trained with CIFAR-10~\cite{cifar} in Figure~\ref{fig:feature_shift}. 
The figure contains test samples from CIFAR-10 and SVHN~\cite{svhn} test sets. For each sample, we plot the raw feature and the feature of its augmented version with a line connecting them. 
It shows that the relative distance between a CIFAR-10 pair~(\textcolor{red}{red}-\textcolor[RGB]{30,144,255}{blue} pairs with black lines) is statistically smaller than that of an SVHN pair~(\textcolor{green}{green}-\textcolor{orange}{orange} pairs with gray lines). Similar observations widely exist across networks, datasets, and augmentation methods~(see Appendix~\ref{app:feature_space}). 
Therefore, a well-trained model is \textit{not} robust towards augmentations of all distributions; Instead, the property only holds for in-distribution data.

%This observation extends the problem scope of TTA~(one distribution) to multiple distributions and inspires us to apply TTA to detect anomalies. Specifically, a well-trained model itself has already mastered enough anomaly detection ability. 
Based on this observation, we develop our algorithm, named TTA-AD~(Test Time Augmentation Anomaly Detection), which has the following characteristics:

\textbf{Performance.} TTA-AD achieves comparable or better detection performance on challenging benchmark vision datasets for both supervised and unsupervised SOTA models.
\textbf{Efficiency.} TTA-AD is a post-hoc algorithm which can be applied directly to a plain classifier. It only computes two forward passes to obtain the anomaly scores, which is far more efficient than most popular classifier-based algorithms. We provide a clock running time comparison in Figure~\ref{fig:runtime}. 
\textbf{Prior-free.} TTA-AD is dataset-agnostic and prior-free. So we do not have to tune the hyper-parameters with out-distribution prior, which may be hard or impossible to obtain during training.
	%	\item \textbf{Black-box.} TTA-AD is a black-box anomaly detection algorithm. It does not require access to internal features and gradients, which is useful when a white box model is unavailable. 
\textbf{Adaptability.} TTA-AD can not only help a plain classification model to detect anomalies, but also adapt to other representation learning anomaly detection methods~\cite{CSI}. Meanwhile, TTA-AD does not require access to internal features and gradients, which is desirable when a white box model is unavailable.

We further analyze the reasons for improved performance through empirical observations and theoretical analysis, which previous works often do not provide.

Empirically, TTA-AD gains benefits from mutual relations between different augmentations of a single sample. We name all the classes except for the maximum class as \textbf{remaining classes}. Specifically, we verify that the mutual relations between the remaining classes of a single sample's different augmentations significantly differ for in- and out-distribution data. 
The idea is different from previous algorithms which focus on maximum probability~\cite{oodbaseline, ODIN, DBLP:conf/nips/HendrycksMKS19}, mutual relations between layers of a sample~\cite{grammatrix}, mutual relations between different samples~\cite{CSI} and class-conditional distributions~\cite{Ma, winkens2020contrastive}.
Section~\ref{sec:empiricalexplain} provides detailed evidence for the huge statistic difference of the remaining classes on in-distribution and out-distribution data.

Theoretically, anomaly detection is essentially a problem of sorting a binary sequence, where the two classes are drawn from in- and out-distributions. In this work, we provide theoretical analysis of TTA-AD's success based on a theoretical tool named RUNS~\cite{wald1940test, barton1957multiple}, which evaluates the degree of confusion of a binary sequence in statistics. A detailed correlation between RUNS and the performance of TTA-AD is discussed in Section~\ref{sec:theoexplain}.
%	Notice that we did not directly prove the increase or decrease of AUROC value, which is a widely used metric to evaluate the anomaly detection performance, since AUROC is theoretically unanalyzable. 
Considering the strong connection between anomaly detection and RUNS, RUNS may serve as a theoretical tool in other anomaly detection works as well.

To conclude, our contributions are: \textbf{1)}~We observe the significant difference of the mutual relations for in- and out-distribution data and propose a new anomaly detection algorithm, TTA-AD, based on the observation. It achieves comparable or better results on challenging benchmark vision datasets for both supervised and unsupervised dataset-vs-dataset anomaly detection. 
\textbf{2)}~We provide both empirical and theoretical evidence to explain how TTA-AD works, which is not addressed by previous algorithms. 
\textbf{3)}~TTA-AD is more practical compared to the previous algorithms. Specifically, it is computationally efficient and free both to out-distribution prior and the model’s internal weights/gradients.

%\begin{itemize}	
%	\item We empirically verify the importance of the remaining classes and propose a new anomaly detection algorithm, TTA-AD. It achieves comparable or better results~(see Table~\ref{table:csi_compare} and Table~\ref{table:imagenetresult}) on challenging vision datasets for both supervised and unsupervised models. 
%	
%	\item TTA-AD is more practical compared to the previous algorithms. Specifically, it is computationally efficient and free both to out-distribution prior and the model’s internal weights/gradients. 
%	
%	\item We find the key to TTA-AD lies in remaining classes empirically, which has been ignored by previous works. And we provide theoretical analysis based on RUNS, which is a powerful analyzing tool not mentioned before in this area as far as we know. 
%	
%	%	\item We analyze the principle of TTA-AD through experiments and provide theoretical analysis based on RUNS, which is a powerful analyzing tool not mentioned before in this area as far as we know. 
%\end{itemize}

%In the following sections, we will first introduce related works in Section~\ref{sec:related} and then state our algorithm in Section~\ref{sec:disentangledata}. In Section~\ref{sec:exp}, we provide detailed anomaly detection performance results. After that, we provide an empirical illustration of our algorithm in Section~\ref{sec:empiricalexplain} and some theoretical analysis in Section~\ref{sec:theoexplain}. The limitation is discussed in Section~\ref{sec:limitation}. Finally, we conclude our paper in Section~\ref{sec:conclusion}.

\section{Related work}
\label{sec:related}

%\paragraph{Anomaly detection} 
%Anomaly detection is a popular research area containing many branches~\cite{AnomalydetectionAsurvey, DeepLearningforAnomalyDetectionAReview}. 
%%Algorithms in this field can be categorized according to different criteria. 
%%From the perspective of application scenarios, it is studied in image classification~\cite{CSI}, segmentation~\cite{DBLP:journals/ijcv/BergmannBFSS21}, video~\cite{abnormaleventdetectioninvideos}, and  language~\cite{oodlanguage} tasks.
%From the perspective of model training, there are supervised/unsupervised classifier-based~\cite{oodbaseline, DBLP:conf/nips/HendrycksMKS19, Ma, ODIN, CSI} and generative-model-based~\cite{anogan, ImageAnomalyDetectionwithGenerativeAdversarialNetworks, DBLP:conf/nips/SchirrmeisterZB20} algorithms.
%From the perspective of anomaly task levels, there are dataset-vs-dataset~\cite{oodbaseline} and one-class-vs-all  settings~\cite{DBLP:conf/nips/GolanE18}.
%Our algorithm focuses on image classification dataset-vs-dataset anomaly detection. 

\paragraph{Dataset-vs-dataset anomaly detection} A series of previous works focus on the dataset-vs-dataset image classification setting. Hendrycks \& Gimpel ~\cite{oodbaseline} observe that correctly classified samples tend to have larger maximum softmax probability, while erroneously classified and out-distribution samples have lower values. The distribution gap allows for the detecting anomalies.
To enlarge the distribution gap, Hendrycks et al.~\cite{DBLP:conf/nips/HendrycksMKS19} enhance the data representations by introducing self-supervised learning and
Liang et al.~\cite{ODIN} propose input pre-processing and temperature scaling. Input pre-processing advocates fine-tuning the input so that the model is more confident in its prediction. Temperature scaling introduces temperature to the softmax function. 
%Formally, the two ingredients are
%\begin{equation}
%\label{eqn:inputpreprocessing}
%\tilde{\bx} = \bx - \epsilon \sign(-\nabla_{\bx} \log S_{\hat{y}} (\bx;t)),
%\end{equation}
%\begin{equation}
%\label{eqn:lastlayertemperature}
%S_i(\bx;t) = \frac{\exp(f_i(\bx)/t)}{\sum_{j=1}^{N}\exp(f_j(\bx)/t)},
%\end{equation}
%where $\bx$ is the input data, $f_i(\bx)$ is the i-th class output of $\bx$, $\epsilon$ is the fun-tuning size and $t$ is the temperature. 
%We denote Eqn~\ref{eqn:inputpreprocessing} as input preprocessing in the rest of the paper.  
Besides maximum probability,  Lee et al.~\cite{Ma} introduce the Mahalanobis score, which computes the class-conditional Gaussian distributions from a pre-trained model and confidence scores. Tack et al.~\cite{CSI} propose contrastive losses and a scoring function to improve anomaly detection performance on high-resolution images. Sastry \& Oore~\cite{grammatrix} use Gram matrices of a single sample's internal features from different layers. 

Each of these classifier-based dataset level algorithms has their own characteristics. Prior out-distribution knowledge is required to tune the parameters by some algorithms~\cite{ODIN, Ma}. Although higher performance is reached, such prior knowledge is rare and unpredictable under certain scenes. Input pre-processing~\cite{generalizedodin, Ma,ODIN} is time-consuming because one needs several times of forwarding and backpropagation. Scoring-based methods are computationally costly since they rely on internal model features~\cite{earlylayer, Ma, grammatrix}. 
Our algorithm utilizes test time augmentation and consistency evaluation to achieve prior-free and computational friendly anomaly detection~(summarized in Table~\ref{table:comparemethods}).

\paragraph{Data augmentations in anomaly detection}
Using data augmentations in anomaly detection has been applied by existing algorithms in both training and evaluating phases. 
Training data augmentation usually aims to learn a better data representation. Previous works employ augmentations from self-supervised learning and contrastive learning in training phase to gain better detection performance~\cite{DBLP:conf/nips/HendrycksMKS19, CSI, zhou2021step}. 
Test data augmentation usually relates to the calculation of anomaly scores. Golan \& El-Yaniv utilize summarized predicted probability over multiple geometric augmentations at test time to detect anomalies~\cite{DBLP:conf/nips/GolanE18}. Wang et al. calculate averaged negative entropy over each augmentation as the detection criterion~\cite{DBLP:conf/nips/WangZLZYXK19}. 

Our TTA-AD uses test data augmentation as well but with two main differences. First, we utilize the relations between augmentations of a single sample, which is not explored by previous works. Second, we empirically verify that the performance improvement is attributed to the remaining classes, while previous works empirically find augmentations are helpful without any reasons. 

%\paragraph{Other anomaly detection algorithms} 
%From the perspective of prior out-distribution knowledge, some works assume no access to out-distribution data at training phase~\cite{oodbaseline,  DBLP:conf/nips/HendrycksMKS19, OCSVM, CSI} while other works use prior out-distribution knowledge to train or tune their anomaly detection algorithms~\cite{Ma,ODIN}. 
%Besides dataset level anomaly detection, there are also class vs. class, one vs. all~\cite{DBLP:conf/nips/GolanE18} and segmentation anomaly detection~\cite{DBLP:journals/ijcv/BergmannBFSS21}. 
Beyond the algorithms we mentioned above, various methods of anomaly detection are well summarized by Chalapathy \& Chawla~\cite{deeplearningforanomalydetectionasurvey}. There are shallow methods such as OC-SVM~\cite{OCSVM}, SVDD~\cite{SVDD}, Isolation Forest~\cite{isolationforest}, and deep methods such as deep SVDD~\cite{deepsvdd} and deep-SAD~\cite{deepsemisupervisedanomalydetection} used to detect anomalies under different anomaly settings. 
%Both \cite{ODIN, Ma}  assume a partial prior knowledge of out-distribution data. In practice, they use $\sim 10\%$ of the out-distribution dataset to select best hyper-parameters such as $T$ and $\epsilon$. 
%\paragraph{Test time augmentation\hhw{TTA in AD}} 
%Typically, data augmentation is performed when training a model to improve generalization ability. At the same time, it can also be used at test time to correct the model predictions~\cite{jin2018deep, DBLP:journals/cacm/KrizhevskySH17}. It is summarized as a property  of neural networks that a well-trained classifier should have a low variance in predictions across augmentations on most data~\cite{DBLP:journals/corr/abs-2011-11156, DBLP:journals/jbd/ShortenK19}. 
%Our algorithm digs deep into the feature space of TTA. Instead of focusing on improving prediction accuracy, we analyze the consistency degree of the models outputs towards original inputs and augmented inputs as shown in Figure~\ref{fig:feature_shift}. 
%\paragraph{Model uncertainty} 
We also notice that model uncertainty is highly related to anomaly detection. In fact, anomaly detection is exactly a task that assigns uncertainty (or anomaly) scores to samples, so the final goals are similar. The main factor that distinguishes them is how such scores are generated~\cite{DBLP:conf/nips/HendrycksMKS19,  DBLP:journals/tmi/SeebockOSWBKLS20}.

\section{Data augmentation and consistency evaluation}
\label{sec:disentangledata}

We present our simple yet effective algorithm TTA-AD, which is purely based on test time data augmentation and consistency evaluation. The whole computation pipeline is shown in Algorithm~\ref{alg:Framwork}.
\begin{algorithm}[htb] 
	\caption{Computation pipeline of TTA-AD.} 
	\label{alg:Framwork} 
	\begin{algorithmic}
		\State {\bfseries Input:}
		A model $f$ pre-trained on in-distribution data;
		A predetermined data augmentation method $T$ drawn from function space $\mathcal{T}$; 
		A set of test data $x_1, x_2,...,x_n$ containing in-distribution and out-distribution data;
		\State {\bfseries Output:} Anomaly scores $S(x_1), S(x_2),...S(x_n)$;
		\For{$i=1$ {\bfseries to} $n$} 
		\State Augment $x_i$ with $T$ to obtain $T(x_i)$;
		\State Feed $x_i$ and $T(x_i)$ into $f$, and compute $f(x_i)$ and $f(T(x_i))$ through forward passes;
		\State $S(x_i)=1-\langle f(x_i), f(T(x_i)) \rangle$; 
		%		\STATE \quad 4. {\bfseries return:} $S(x_i)$; 
		\EndFor
	\end{algorithmic} 
	
\end{algorithm}

Our algorithm adapts to both supervised and unsupervised pre-trained models. The performance in both cases is covered in experiments. Below, we first explain the two cores of our algorithm in detail.

%\begin{figure*}[htbp]
%	\centering
%	\begin{minipage}[t]{0.9\linewidth}
%		\centering
%		\includegraphics[width=3.5in]{figure/page3_crop.pdf}
%		\caption{\label{fig:disentangling}Computational pipeline: test time data augmentation and consistency evaluation.}
%	\end{minipage}%
%\end{figure*}

%The intuition behind our algorithm is that, if a deep learning model has learned the training data, i.e. the in-distribution data, then it should be more robust to augmentations of in-distribution data. On the contrary, though the model will give wrong but confidential prediction to out-distribution data, the consistency value of different augmentations of out-distribution data outputs should be relatively low. 

%\subsection{Notation\hhw{move to proof, ignore}}
%We denote the in-distribution dataset as $X^{in}$, and the out-distribution dataset as $X^{out}$. $X^{in}$ and $X^{out}$ are drawn from different distribution. Given data $X=\{\bx_1, \bx_2, \bx_3,...\bx_n\}$, a disentangling function $D$ drawn from function space $\mathcal{D}$, disentangling process can be described as 
%$$\bx_i \rightarrow D(\bx_i)$$

\subsection{Data augmentation and feature space distance}
We use data augmentation techniques without backpropagation at test time, which is similar to TTA, to help enlarge the gap between in-distribution and out-distribution samples. Formally, given test data $X=\{\bx_1, \bx_2, \bx_3,...,\bx_n\}$, a transformation function $T$ drawn from function space $\mathcal{T}$, the augmentation process can be described as 
$\bx_i \rightarrow T(\bx_i)$.
%As mentioned before, the transformation is inspired by test time augmentation. The main difference between our transformation and contrastive learning is that training label information is gathered in semi-supervised anomaly detection tasks. The representation of in-distribution data is supposed to be well learned by the classifier. Because of this, the classifier will make more consistent outputs of raw input and its transformed version for in-distribution data than out-distribution data.
In our algorithm, the function space $\mathcal{T}$ contains Fast Fourier Transformation~(FFT) and Horizontal Flip~(HF) because they satisfy the sensitivity criterion. Specifically, recent studies find that neural networks are less error-prone to low-frequency components~\cite{highfreq}, and deep neural networks and datasets are sensitive to visual chirality~\cite{lin2020visual}. TTA-AD extends these conclusions by revealing that while the neural networks are known to be sensitive to high frequency components and visual chirality~(horizontal flip), the sensitivity is enhanced in out-distribution data.
%Limited to space, we put some image examples of the proposed augmentations in Appendix~\ref{app:transformationexamples}.
%Overall, the choice of transformations should obey the principle that the semantic meaning should not be changed. This is reasonable since otherwise in-distribution and out-distribution data will have similar difference magnitude in feature space.
%Empirically, we observe that the linear transformations, which are equivalent to adding a convolution layer at the head of the classifier, are not effective for detecting anomalies--for example, linear blurring and Gaussian blurring. We leave the details and experiments in Appendix~\ref{app:othermethods}. In practice, we mainly use two kinds of efficient transformation methods in our experiments: Fast Fourier Transformation~(FFT) and Horizontal Flip~(HF). 

FFT is a widely used efficient image processing technique~\cite{bracewell1986fourier, oppenheim1997signals}. FFT converts an image from the space domain to the frequency domain using a linear operator. Then the frequency representation can be inversely converted into the original image data by IFFT~(Inverse Fast Fourier Transformation). In TTA-AD, we propose to cut off partial sensitive high-frequency signals. 
Our ablation study on the filter radius in Section~\ref{sec:ablation} shows that the anomaly detection performance fluctuates little within a wide range of filter radius. In the following text, we will use $\text{FFT}_{100}$ to represent the FFT and IFFT transformation with a 100-pixel filter radius. 
Since discovering visual chirality~\cite{lin2020visual}, there have not been many studies exploring how to exploit the property. We introduce it into anomaly detection, and according to our observations, it indeed enlarges the distribution gap. We verify through extensive experiments~(Figure~\ref{fig:feature_shift} and Appendix~\ref{app:feature_space}) to show that the sensitivity is more significant in out-distribution data with both transformations. Specifically, the distance of paired in-distribution sample features is statistically shorter than that of paired out-distribution samples. Our analysis in Section~\ref{sec:algorithmanalysis} further provides the empirical reasons and theoretical analysis. 
%Despite the widespread use of horizontal flip in training neural networks, it was not until recently that the pervasive visual chirality is discovered~\cite{lin2020visual}. It inspires us to use horizontal flip to enlarge the distribution gap between in-distribution and out-distribution data. 

\paragraph{Discussion} We note that there are also other transformations which may satisfy the sensitivity criterion such as adversarial examples. However, they may bring undesired randomness and high computation burden, which is contrary to the original intention of TTA-AD. Additionally, employing data augmentation is not a new idea in anomaly detection as mentioned in Section~\ref{sec:related}. Unlike previous algorithms, our algorithm focuses on the \textit{relationship} between augmented feature pairs instead of treating each augmented output separately. We next illustrate the measurement of the mutual relationship in the following.

%We also observe that some linear transformation functions are not beneficial to anomaly detection--for example, linear blurring and Gaussian blurring. We leave details and results of those defective transformations in Appendix~\ref{app:othermethods}. These findings may imply that neural networks and vision datasets are not as sensitive as high frequency component and visual chirality to the considered linear transformations and do not have similar properties like "linear chirality". 

%The effect in feature space of applying these transformations at test time is visualized in Figure~\ref{fig:feature_shift} and Appendix~\ref{app:feature_space}.  

\subsection{Consistency Evaluation}

%As a substitution of maximum softmax probability used in \cite{oodbaseline, ODIN}, we propose inner product of the model output.  
Unlike previous works that ignore the interaction between augmentations of a sample, our anomaly score evaluates the relations between augmentations of a single sample in a simple but effective form. 
With a model $f$ trained on the training~(in-distribution) data, we denote the softmax output of $\bx$ as $f(\bx)$. Suppose we fix a transformation method $T$. We define the consistency score of a given data $\bx$ as the inner product of the model output of $\bx$ and $T(\bx)$, which is $\langle f(\bx), f(T(\bx))\rangle$.

%\paragraph{Fact 1}Here we state the basis of our algorithm. For in-distribution data $x_i\sim X_{in}$, the classifier $f$ tends to be confidential about its decision, which means $f(D(x_{i}))$ is nearly the same as $f(x_{i})$. On the contrary, for $x_o \in X_{out}$, the classifier is likely to make divergent decision, thus $S_{sim}(x_o)$ is relatively small. We claim that this is a common phenomenon and we will show the experimental observation in Section~\ref{sec:exp}. \hhw{any more explanation?} \hhw{Or maybe show an experimental example here}

To further enlarge the gap of the model output for in- and out-distribution data, we use a fixed temperature scaling technique. Given a sample $\bx$, we denote the softmax output with temperature scaling parameter $t$ as $f(\bx;t)$. 
%Another thing needs to illustrate is that we label in-distribution data as negative data ($0$) and out-distribution data as positive data ($1$). 
And, define our final anomaly score for a given sample $\bx$ as
\begin{equation}
	\label{eqn:oodscore}
	S(\bx) = 1 - \langle f(\bx;t), f(T(\bx;t))\rangle.
\end{equation}
Based on the observation~(Figure~\ref{fig:feature_shift}), the model $f$ tends to make more consistent outputs for in-distribution data than out-distribution data. So $S(\bx)$ should be close to $0$ for in-distribution data and  $S(\bx)$ should be approximately $1$ for out-distribution data. Such score assigning process is the basis of our theoretical analysis in Section~\ref{sec:theoexplain}.

\section{Experiments}
\label{sec:exp}

\subsection{Compared algorithms and metrics}
In anomaly detection, algorithms with different settings often cannot be measured simultaneously. For example, some algorithms focus on the one-class-vs-all setting~\cite{deepsvdd, DBLP:conf/nips/GolanE18, DBLP:conf/nips/WangZLZYXK19, DBLP:conf/iccv/GongLLSMVH19}, while others focus on the dataset-vs-dataset setting~\cite{ODIN, zhou2021step} only. There are also algorithms that cover multiple settings~\cite{DBLP:conf/nips/HendrycksMKS19, CSI}. Since our algorithm relies on the remaining classes, i.e. multiple classes in the training set, those one-vs-all algorithms are not compared. The SOTA algorithm of advanced dataset is CSI~\cite{CSI}. We also cover MSP~\cite{oodbaseline}, ODIN~\cite{ODIN}, Mahalanobis~\cite{Ma} and Rot~\cite{DBLP:conf/nips/HendrycksMKS19}. 

Following previous works~\cite{DBLP:conf/nips/HendrycksMKS19, Ma, CSI}, we use the AUROC~(Area Under Receiver Operating Characteristic curve) as our evaluation metric. It summarizes True Positive Rate~(TPR) against False Positive Rate~(FPR) at various threshold values. The use of the AUROC metric frees us from fixing a threshold and provides an overall comparison metric. 
%\paragraph{TNR, TPR, FPR, FNR} We treat the out-distribution data as positive samples and use True Negative Rate(TNR), True Positive Rate(TPR), False Positive Rate(FPR) and False Negative Rate(FNR) to evaluate the detection performance.

% Statisticians often use the Area Under ROC(AUROC) curve to reflect the overall performance. 

\subsection{Dataset settings and classifier training}
\label{sec:dataset}

Dataset settings are diverse in anomaly detection. We mainly consider the advanced ImageNet~\cite{deng2009imagenet} settings, including full set settings and two kinds of subset settings. We also verify on CIFAR~\cite{cifar} settings and put the results in Appendix~\ref{app:cifar_results} due to limited space.

%\paragraph{ImageNet based settings} 
%First of all, we evaluate our algorithm on the full dataset and subsets. We will introduce the settings in detail below. 

In the full-size ImageNet setting, the whole validation set is treated as the in-distribution test set. Since the ImageNet training set already contains all-embracing natural scenery, the corresponding out-distribution dataset is chosen to be a publicly available artificial image dataset from Kaggle\footnote{\url{https://www.kaggle.com/alamson/safebooru}}. We use the first 50000 figures since the ImageNet validation dataset is of the same size. Some examples of the dataset are depicted in Figure~\ref{fig:dataset} in Appendix~\ref{app:datasetconstruction}. We will call this setting \textit{ImageNet vs. Artificial}.

%The reason for advocating a new benchmark is to 1) identify the performance of previous nearly perfect algorithms\footnote{The SOTA algorithm reaches 99.9\% AUROC on CIFAR} 2) Verify the effectiveness of our algorithm.
Additionally, two sources of subset settings are included. The \textit{ImageNet-30} subset setting is used in relevant researches~\cite{DBLP:conf/nips/HendrycksMKS19, CSI}. The corresponding out-distribution datasets includes CUB-200~\cite{CUB}, Stanford Dogs~\cite{stanforddogs}, Oxford Pets~\cite{parkhi2012cats}, Places-365~\cite{places} with small images (256*256) validation set, Caltech-256~\cite{griffin2007caltech}, and
Describable Textures Dataset (DTD)~\cite{dtd}. Beyond that, we also validate our algorithm on several public ImageNet subset combinations introduced by other works~\cite{geirhos16, imagenetsubset11} and public sources\footnote{\url{https://github.com/MadryLab/robustness}}.These additional three ImageNet subsets are \textit{Living 9}, \textit{Geirhos 16}, and \textit{Mixed 10}. The out-distribution categories are randomly selected from the complement set. The detailed subset hierarchy and configuration are shown in Appendix~\ref{app:datasetconstruction}. In total, we consider four subset settings.

%In our experiments, we find that some of the previous algorithms can not handle the GPU memory consumption caused by storing high-resolution images' internal features. Therefore we build several ImageNet subsets. The basic principles of constructing subsets are as follows. 
%We construct the in-distribution subsets with publicly available subsets~\cite{geirhos16, imagenetsubset11} and some subset hierarchies from public sources\footnote{\url{https://github.com/MadryLab/robustness}}. 
%The corresponding out-distribution dataset classes are randomly selected from the complement categories. The three ImageNet subsets are \textit{Living 9}, \textit{Geirhos 16}, and \textit{Mixed 10}. When training the classifier, we balance the training and validation datasets to make the whole training size comparable to traditional ones and alleviate the training imbalance problem. The detailed subset hierarchy description and configuration are shown in Appendix~\ref{app:datasetconstruction}. We also evaluate our algorithm with a similar subset, \textit{ImageNet-30}~\cite{DBLP:conf/nips/HendrycksMKS19}, vs. other datasets~\cite{Khosla2012NovelDF,nilsback2006visual,parkhi2012cats,CUB} in Appendix~\ref{app:otherrelevantalgorithms}. 

%\paragraph{Classifier training}
We notice the performance of anomaly detection is highly affected by the trained classifiers. So we train networks with three random seeds per setting and report the mean and variance. The training details are in Appendix~\ref{app:classifier_training}. 

%\paragraph{CIFAR-10 based settings} We run CIFAR-10 experiments following the previous works~\cite{oodbaseline,Ma,ODIN}. Specifically, the CIFAR-10 dataset is treated as in-distribution data, while the SVHN~\cite{svhn}, TinyImageNet~\cite{tinyimagenet}, and LSUN~\cite{lsun} datasets are out-distribution datasets. 

\subsection{Detection results}
\label{sec:imagenet_exp_results}

This section covers the main anomaly detection results including the full ImageNet setting~(Table~\ref{table:imgnet}) and four ImageNet subset settings~(Table~\ref{table:csi_compare}~and \ref{table:imagenetresult}). %, and (3) CIFAR settings~(Table~\ref{table:cifarresult} in Appendix), respectively. 

\begin{table}[!h]
	\caption{\label{table:imgnet}AUROC~(\%) of full \textit{ImageNet vs. Artificial} dataset on the pre-trained torchvision models~\cite{pytorch}.}
	\centering
	\begin{tabular}{lcc}
		\toprule[1pt]
		\multicolumn{1}{c}{\multirow{2}{*}{Algorithm}} & \multicolumn{2}{c}{Architecture}                                      \\ \cline{2-3} 
		\multicolumn{1}{c}{}                           & \multicolumn{1}{l}{ResNet-50}     & \multicolumn{1}{l}{DenseNet-121} \\ \toprule[1pt]
		MSP                                       & $81.05$                           & $79.81$                           \\
		$\text{\textbf{TTA-AD}}_{\text{FFT40}}\text{~(ours)}$                                        & $86.91$ & $89.37$ \\
		$\text{\textbf{TTA-AD}}_{\text{FFT100}}\text{~(ours)}$                                        & \bm{$92.08$} & \bm{$90.84$} \\
		$\text{\textbf{TTA-AD}}_{\text{Flip}}\text{~(ours)}$                                           & $90.20$                           & $89.36$                           \\ \toprule[1pt]
	\end{tabular}
\end{table}

\begin{table*}[h]
	\centering
	\caption{\label{table:csi_compare}AUROC~(\%) of \textit{ImageNet-30} results on unsupervised models. Higher average AUROC score is reached by TTA-AD. On Places-365, our algorithm improves the SOTA algorithm by 16.7\%.}
	\setlength{\tabcolsep}{1mm}
	\begin{tabular}{ccccccc|c}
		\toprule[1pt]
		\multirow{2}{*}{Algorithm} & \multicolumn{6}{c|}{Out-distribution dataset} & \multicolumn{1}{c}{\multirow{2}{*}{AVG.}}       \\ \cline{2-7} 
		& CUB-200 & Dogs & Pets  & Places  & Caltech  &
		DTD & \multicolumn{1}{c}{} \\  \midrule[1pt]
		Rot+Trans &  $74.5_{\pm0.5}$      & $77.8_{\pm1.1}$   & $70.0_{\pm0.8}$    &  $53.1_{\pm1.7}$ & $70.0_{\pm0.2}$  &$89.4_{\pm0.6}$ & $72.5$\\
		CSI Unlabeled                                                                                                                                             & \bm{$90.5_{\pm0.1}$}  & \bm{$97.1_{\pm0.1}$}                 & $85.2_{\pm0.2}$    &  $78.3_{\pm0.3}$ &$87.1_{\pm0.1}$ &\bm{$96.9_{\pm0.1}$} & $89.2$   \\
		%		\begin{tabular}[c]{@{}c@{}}TTA-AD (unsupervised) \end{tabular} & $86.5_{\pm0.1}$ & $95.0_{\pm0.1}$		& \bm{$94.9_{\pm0.1}$}  & \bm{$91.3_{\pm0.2}}$ & \bm{$92.3_{\pm0.1}}$ & $92.5_{\pm0.1}$ \\
		$\text{\textbf{TTA-AD}}_{\text{Flip}}$~(ours) & $86.5_{\pm0.1}$ & $95.0_{\pm0.1}$		& \bm{$94.9_{\pm0.1}$}  & \bm{$91.3_{\pm0.2}$} & \bm{$92.3_{\pm0.1}$} & $92.5_{\pm0.1}$ & \bm{$92.1$}\\
		%		\color[HTML]{C0C0C0} CSI Labeled                                                                                                 & {\color[HTML]{C0C0C0} $94.6$} & {\color[HTML]{C0C0C0} 98.3} & {\color[HTML]{C0C0C0} $97.4$} & {\color[HTML]{C0C0C0} 96.2} \\ 
		\bottomrule[1pt]
	\end{tabular}
\end{table*}

\begin{table*}[h]
	\caption{\label{table:imagenetresult}AUROC~(\%) of ImageNet subset results on supervised models. TTA-AD has a uniform performance improvement on all settings. Best results are in \textbf{bold}. Second best results are \underline{underlined}.}
	\centering
	\begin{threeparttable}
		\begin{tabular}{clccc}
			\toprule[1pt]
			\multicolumn{1}{l}{\multirow{2}{*}{Architecture}} & \multicolumn{1}{c}{\multirow{2}{*}{Algorithm}} & \multicolumn{3}{c}{ImageNet subset settings}                                                                                            \\ \cline{3-5} 
			\multicolumn{1}{l}{}                              & \multicolumn{1}{c}{}                           & Living 9                                 & Geirhos 16                               & Mixed 10                                 \\ \toprule[1pt]
			\multirow{7}{*}{ResNet-50}                        & MSP                                       & $95.32_{\pm0.15}$                       & $90.06_{\pm1.37}$                       & $95.29_{\pm0.06}$                       \\
			%			& OC-SVM                                         & $49.12\pm0.67$                       & $49.65\pm0.59$                       & $47.81\pm2.38$                       \\
			& ODIN                                           & $96.78_{\pm0.36}$                       & $82.35_{\pm0.70}$                       & $95.92_{\pm0.78}$                       \\
			& Mahalanobis\tnote{1}          & $64.21_{\pm3.31}$                       & $64.31_{\pm1.96}$                       & $71.33_{\pm8.11}$                       \\
			& $\text{\textbf{TTA-AD}}_{\text{FFT40}}\text{~(ours)}$                                          & \bm{$98.49_{\pm0.02}$} & \bm{$95.53_{\pm0.39}$} & \bm{$98.36_{\pm0.10}$} \\
			& $\text{\textbf{TTA-AD}}_{\text{FFT100}}\text{~(ours)}$                                         & $97.17_{\pm0.20}$                       & $92.53_{\pm0.67}$                       & $97.25_{\pm0.13}$                       \\
			& $\text{\textbf{TTA-AD}}_{\text{Flip}}\text{~(ours)}$                                           & \underline{$97.29_{\pm0.19}$}                       & \underline{$92.82_{\pm0.68}$}                       & \underline{$97.38_{\pm0.15}$}                       \\ \midrule[0.25pt]
			\multirow{7}{*}{DenseNet-121}                     & MSP                                       & $95.57_{\pm0.07}$                       & $91.76_{\pm0.09}$                       & $94.99_{\pm0.40}$                       \\
			%			& OC-SVM                                         & $47.83\pm0.50$                        & $49.01\pm1.50$                        & $44.70\pm1.30$                         \\
			& ODIN                                           & $96.61_{\pm0.52}$                       & $86.61_{\pm4.36}$                       & $96.49_{\pm0.83}$                       \\
			& Mahalanobis\tnote{2}          & $49.01_{\pm2.95}$                       & $59.26_{\pm5.83}$                       & $65.52_{\pm3.88}$                       \\
			& $\text{\textbf{TTA-AD}}_{\text{FFT40}}\text{~(ours)}$                                          & \bm{$98.50_{\pm1.41}$} & \bm{$95.89_{\pm2.69}$} & \bm{$98.17_{\pm1.38}$} \\
			& $\text{\textbf{TTA-AD}}_{\text{FFT100}}\text{~(ours)}$                                         & $97.21_{\pm0.07}$                       & $93.10_{\pm0.24}$                       & \underline{$96.82_{\pm0.35}$}                       \\
			& $\text{\textbf{TTA-AD}}_{\text{Flip}}\text{~(ours)}$                                           & \underline{$97.26_{\pm0.08}$}                       & \underline{$93.23_{\pm0.22}$}                       & $96.80_{\pm0.33}$                       \\ \toprule[1pt]
		\end{tabular}
		\begin{tablenotes}
			\footnotesize
			\item[1,2] We use the official code and hyper parameter settings of the paper. The implementation details are in Appendix~\ref{app:attempts_to_improve}.
		\end{tablenotes}
	\end{threeparttable}
\end{table*}

\paragraph{ImageNet} For the full \textit{ImageNet vs. Artificial} setting, we only compare our post-hoc TTA-AD with the MSP algorithm due to the computational costs of other algorithms being too large for the full size ImageNet. The performance is shown in Table~\ref{table:imgnet}.

\paragraph{ImageNet subsets} For \textit{ImageNet-30}, we adapt the TTA-AD to the unsupervised CSI model by substituting the predicted probabilities in Eqn.~\eqref{eqn:oodscore} with output features~(512 dimensional tensors). The results in Table~\ref{table:csi_compare} shows that a better average AUROC value is reached compared with Rot+Trans~\cite{DBLP:conf/nips/HendrycksMKS19} and the SOTA algorithm, CSI~\cite{CSI}, with designed scoring functions.

For the other three subset settings, \textit{Living 9}, \textit{Geirhos 16}, and \textit{Mixed 10}, we compare TTA-AD with supervised algorithms listed in Table~\ref{table:imagenetresult}. We can see that TTA-AD is more efficient and outperforms all counterparts. All the compared algorithms are implemented according to the original paper and code except necessary modification of ImageNet figure size. Implementation details are stated in Appendix~\ref{app:attempts_to_improve}.

\paragraph{Discussion} 
Zhou et al.~\cite{zhou2021step} report that severe performance degradation of the Mahalanobis~\cite{Ma} exists when generalizing to other datasets due to validation bias. In order to verify whether TTA-AD also has similar problems, we conduct experiments on CIFAR-10 based settings to test the generalization ability of TTA-AD. Limited to space, the detection results are in Appendix~\ref{app:cifar_results}. Our experiments lead to a similar conclusion as Zhou et al. that simpler algorithms generalize better on different settings. The results also show that TTA-AD can not only achieve better results on advanced datasets, but also has stable generalization performance on low-resolution datasets.

%And some ingeniously designed algorithms indeed reach nearly perfect results. However, we will find that some advanced algorithms are hard to generalize to higher resolution datasets.

\begin{table*}[htbp]
	\caption{\label{table:comparemethods} Training and testing costs of highly related~(classifier-based) algorithms. Running time experiments and detailed explanations are shown in Figure~\ref{fig:runtime},  Section~\ref{sec:algorithms_running_time} and Appendix~\ref{app:computational cost}. TTA-AD requires the least amount of calculation.~(FP: forward pass. BP: backpropagation. Req.: Required.)}
	\centering
	%	Y: required. N: not required. 
	\begin{threeparttable}
		\begin{tabular}{|l|c|c|c|c|}
			\hline
			\multicolumn{1}{|c|}{Algorithm} & \begin{tabular}[c]{@{}c@{}}Access to the \\ internal weights\end{tabular}  & \begin{tabular}[c]{@{}c@{}}Out-distribution  \\   data prior\end{tabular} & \begin{tabular}[c]{@{}c@{}}Parameters\\ search\end{tabular} & \begin{tabular}[c]{@{}c@{}}Computational  \\ cost\end{tabular} \\ \hline
			%			Baseline & \textbf{N}  &  \textbf{N}  & \textbf{N}   & \textbf{1FP}                 \\ \hline
			%		Baseline+                          & Y                          & N                                                                                           & N                                                               &                                     1FP+1BP+1FP                                   \\ \hline
			ODIN~\cite{ODIN}                               & \textbf{Not Req.}                          & Req.                                                                                           & Req.                                                               &                                     1FP+1BP+1FP                                   \\ \hline
			Mahalanobis~\cite{Ma}    & Req.       & Req.    & Req.          &
			\begin{tabular}[c]{@{}c@{}}1FP+k$\times$(1FP+1BP+\\1FP)\tnote{1}  +Regression\tnote{2}    \end{tabular}                           \\ \hline
			\textbf{TTA-AD}~(ours)                               & \textbf{Not Req.}                          & \textbf{Not Req.}  & \textbf{Not Req.}    &  \textbf{2FP}        \\ \hline
		\end{tabular}
		\begin{tablenotes}
			\footnotesize
			\item[1] Compute feature mean and covariance of k selected layers.
			\item[2] Regression on the computed Mahalanobis score.
		\end{tablenotes}
	\end{threeparttable}
\end{table*}
\vspace{-10pt}

\subsection{Algorithms running time}
\label{sec:algorithms_running_time}

We show computational costs of all considered algorithms in this section since running time is important in practical use. We run all experiments on an RTX3090 graphics card. The post-hoc algorithms, ODIN and TTA-AD, do not include a training phase, while the Mahalanobis needs a little bit of time to train. All implementations are based on official code sources~(See Appendix~\ref{app:attempts_to_improve}). Figure~\ref{fig:runtime} shows that TTA-AD only requires $10\%\sim40\%$ running time of advanced algorithms while maintaining comparable or better detection results~(Table~\ref{table:imagenetresult}). The difference in running time is mainly due to the use of different techniques as summarized in Section~\ref{sec:related} and Table~\ref{table:comparemethods}. The number of parameter searching groups comes directly from the original papers and codes. The running time of CSI is missing because introducing contrastive learning is often considered much more time-inefficient than plain training which is the default training methods of the three algorithms in Figure~\ref{fig:runtime}. The details, criteria for choosing comparison algorithms and running times on other network architectures are stated in Appendix~\ref{app:computational cost}.

\begin{wrapfigure}{r}{0.48\textwidth}
	%	\setcaptionwidth{2.2in} 
	\centering
	%	\vspace{-15pt}
	\includegraphics[width=0.80\linewidth]{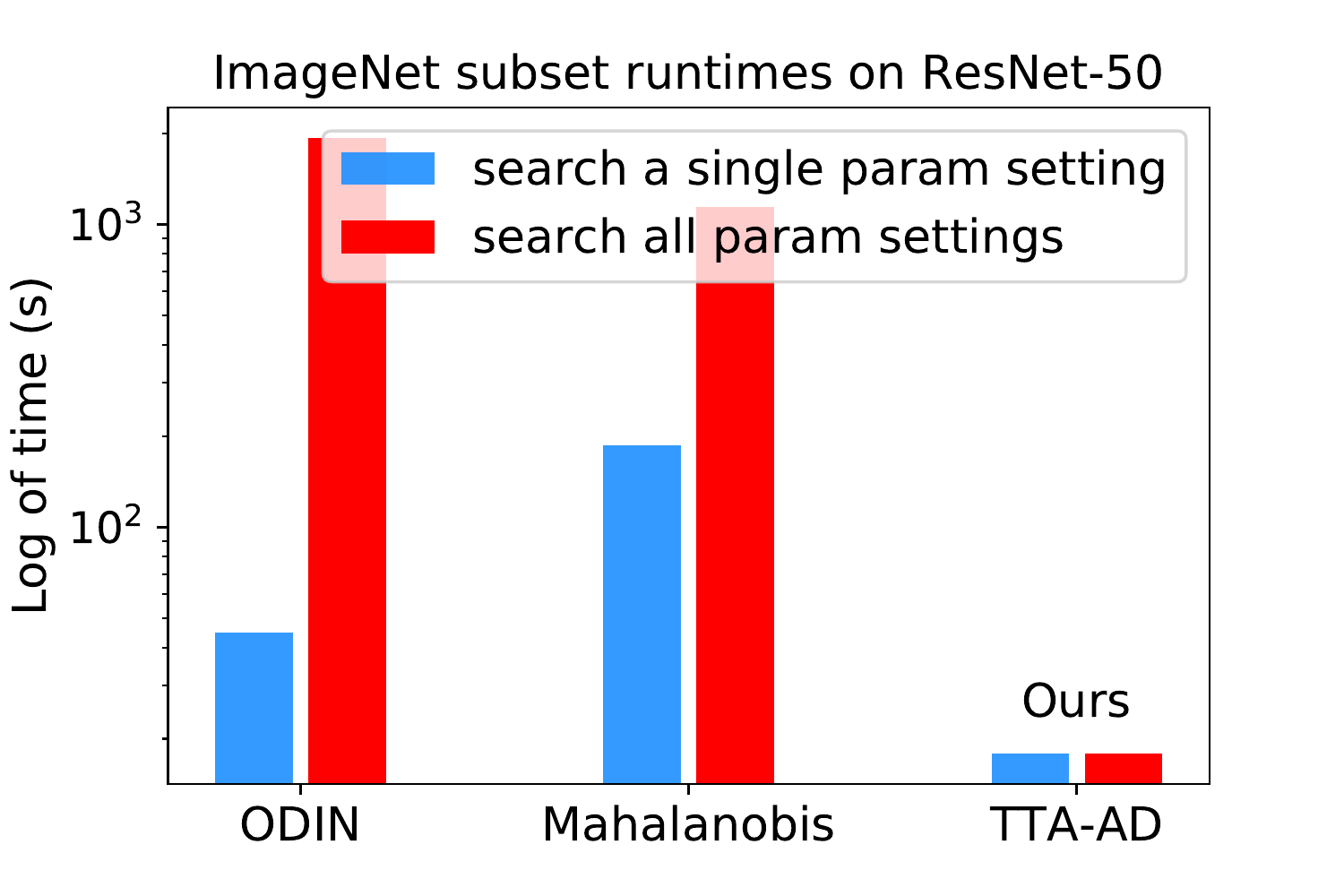}
	\caption{\label{fig:runtime} Average classifier-based algorithm running time on ImageNet subset \textit{Living 9}. For algorithms that require parameter search, we provide the total time of searching all parameter settings and the time of a single search. TTA-AD }
	\vspace{-10pt}
\end{wrapfigure}

% 3090 experiments: 20210929 
%(1) OOD_Baseline_and_ODIN_imagenet Baseline time: 45.69 ODIN settings: time:  TTA-AD time: ind freq 
%(2) OOD_Baseline_and_ODIN_imagenet Ma 
%(3) OOD_Baseline_and_ODIN_imagenet TTA-AD time: ind freq 

%ind baseline method time 2.5516772270202637
%ood baseline method time 5.318024635314941
%
%
%
%cifar10 svhn resnet 34 baseline 37.40 
%ODIN time 1624.2168319225311
%
%ind freq method time 3.163581132888794
%ood freq method time 6.424107551574707
%
%sample mean 19.630412340164185
%magnitude, time 0.002 227.37840700149536

\subsection{Ablation studies for filter radius and temperature}
\label{sec:ablation}
Performance tables in this paper include several different augmentation methods of TTA-AD. Although the AUROC values fluctuate with different augmentation methods and filter radius, the performance is similar. With linear searching across the FFT filter radius and temperature settings, we demonstrate that they have a small effect on the anomaly detection performance within a wide range.

\begin{figure}[htbp]
	\centering
	%	\vspace{-15pt}
	\begin{minipage}[c]{0.45\textwidth} %minipage使之保持同一行，0.2占这行的0.2	
		\centering
		\includegraphics[width=\linewidth]{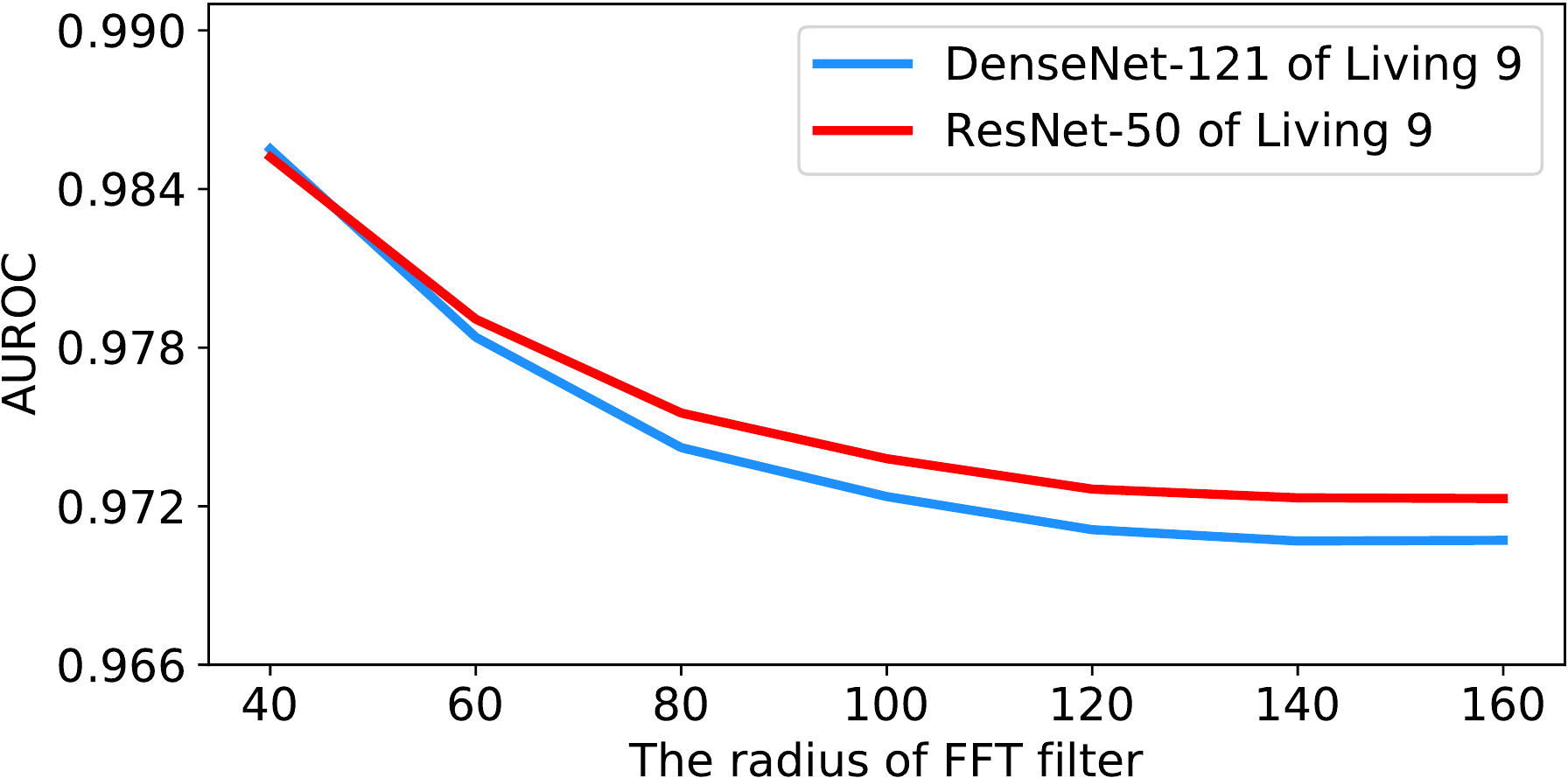}
		\caption{\label{fig:freqsensitivity} Detection AUROC  with different radius of the FFT filter. Flatter curves mean less sensitivity to parameters.}
	\end{minipage}%
	\hspace{10pt}
	\begin{minipage}[c]{0.45\textwidth}
		%		\setcaptionwidth{2.2in} 
		\centering
		\includegraphics[width=\linewidth]{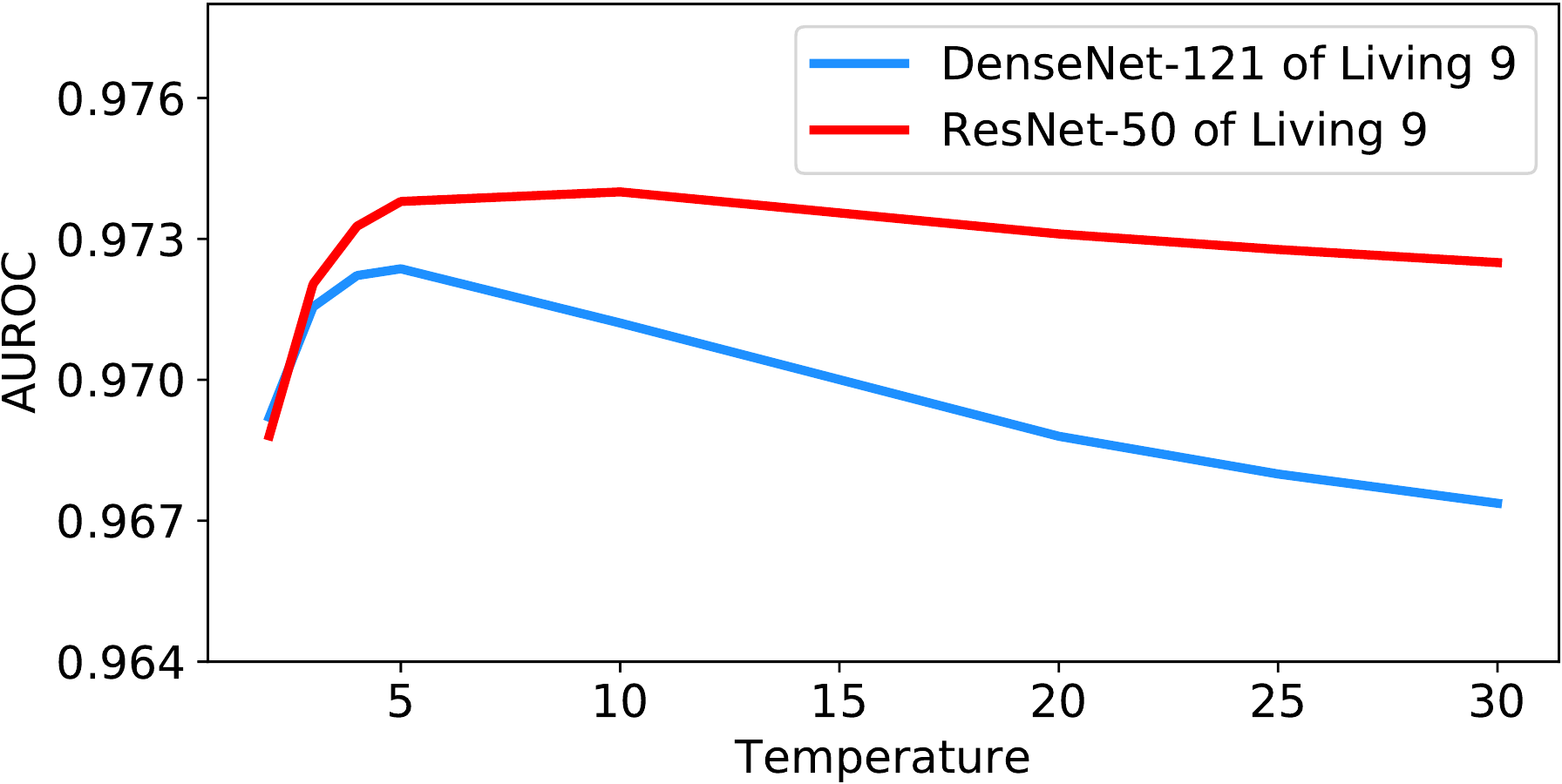}
		\caption{\label{fig:tempsensitivity} Detection AUROC  with different temperature. Flatter curves mean less sensitivity to parameters.}
	\end{minipage}
	%	\vspace{-10px}
\end{figure}

For the FFT filter radius, we provide ablation studies across a wide range of filter radius settings, ranging from 40 to 160 pixels, in Figure~\ref{fig:freqsensitivity}. Among all settings, TTA-AD does not degrade much and still outperforms other algorithms as shown in Table~\ref{table:imagenetresult}, which demonstrates that the performance of TTA-AD is not very sensitive to the filter radius. A possible reason is that the energy is highly concentrated in the low frequency space instead of high frequency parts. More ablations of other settings are in Appendix~\ref{app:sensitivitytofilterradius}.

%\vspace{-8px}

%we empirically reserve about 90\% content of the frequency domain. That is, for the CIFAR-10 dataset, which contains 32x32 RGB images, we roughly preserve the low-frequency part inside a circle with a radius of 20 pixels. 
%For ImageNet, where the figure is resized to 224x224, we pick the frequency filter radius as 100. We also emphasize that the choice of filter frequency is not very sensitive for high-resolution images. We try different frequency filters, ranging from 40 to 160 pixels, and Figure~\ref{fig:freqsensitivity} shows that our algorithm can maintain good detection results in a wide frequency range. 

Unlike the previous algorithms~\cite{Ma,ODIN}, which consider large-scale candidates of temperature scaling~(e.g., from 1 to 1000) and then do the grid-search with prior out-distribution knowledge, we empirically find that the best temperature for our method is stable around 5 and the performance does not decrease much within the considered wide temperature range. 
All our experiments set the temperature parameter to 5 if not specified. Ablation studies for temperature parameter are shown in Figure~\ref{fig:tempsensitivity}, and more results can be found in Appendix~\ref{app:sensitivitytotemperature}. All curves show that TTA-AD is not sensitive to the temperature parameter, thus we do not have to tune it with out-distribution prior.

%\begin{twocolumn}
%	\begin{figure}[htbp]
%		\centering
%		\includegraphics[width=1.5in]{figure/temp/freq_effect_img_ani.pdf}

%		\caption{\label{fig:freqsensitivity}Detection AUROC of ImageNet vs Anime dataset with different radius of FFT filter.}
%	\end{figure}
%
%	\begin{figure}[htbp]
%		\centering
%		\includegraphics[width=1.5in]{figure/temp/temp_effect_living9.pdf}

%		\caption{\label{fig:tempsensitivity}Detection AUROC of Living 9 setting with different temperature.}
%	\end{figure}
%\end{twocolumn}

%\begin{figure*}[htbp]
%	\begin{minipage}[c]{0.46\textwidth} 
%		\centering
%		\includegraphics[width=2.5in]{figure/temp/freq_effect_img_ani.pdf}
%		\caption{\label{fig:freqsensitivity}Detection AUROC  with different radius of FFT filter. Flatter curves mean less sensitivity to parameters.}
%	\end{minipage}%
%	\begin{minipage}[c]{0.46\textwidth}
%		\centering
%		\includegraphics[width=2.6in]{figure/temp/temp_effect_living9.pdf}
%		\caption{\label{fig:tempsensitivity}Detection AUROC  with different temperature.Flatter curves mean less sensitivity to parameters.}
%	\end{minipage}
%\end{figure*}

\section{Algorithm analysis}
\label{sec:algorithmanalysis}
In this section, we will explain why data augmentation and consistency evaluation help to detect anomalies empirically~(Section~\ref{sec:empiricalexplain}) and mathematically~(Section~\ref{sec:theoexplain}). 
%and assumptions and then analyze what is happening in one slot and the whole slots consequently.

\subsection{Empirical analysis of using remaining classes probability}
\label{sec:empiricalexplain}
Our anomaly score~(Eqn~\eqref{eqn:oodscore}) adds up all class probabilities rather than focusing only on the maximum predicted probability~\cite{oodbaseline, DBLP:conf/nips/HendrycksMKS19}. The main difference between consistency evaluation and maximum predicted probability comes from the remaining score $S_{rem}$, defined as
\begin{equation}
	S_{rem} = \langle f(\bx), f(T(\bx)) \rangle - f_j(\bx) f_j(T(\bx)), \label{eqn:Sremain}
\end{equation}
where $j=\arg \max f_j(\bx)$ is the predicted class with the highest probability of the input $\bx$.

\vspace{-10pt}

\begin{figure}[htbp]
	\centering
	\subfigure[\label{fig:max_probability}]{
		\begin{minipage}[t]{0.32\linewidth}
			\centering
			\includegraphics[width=\linewidth]{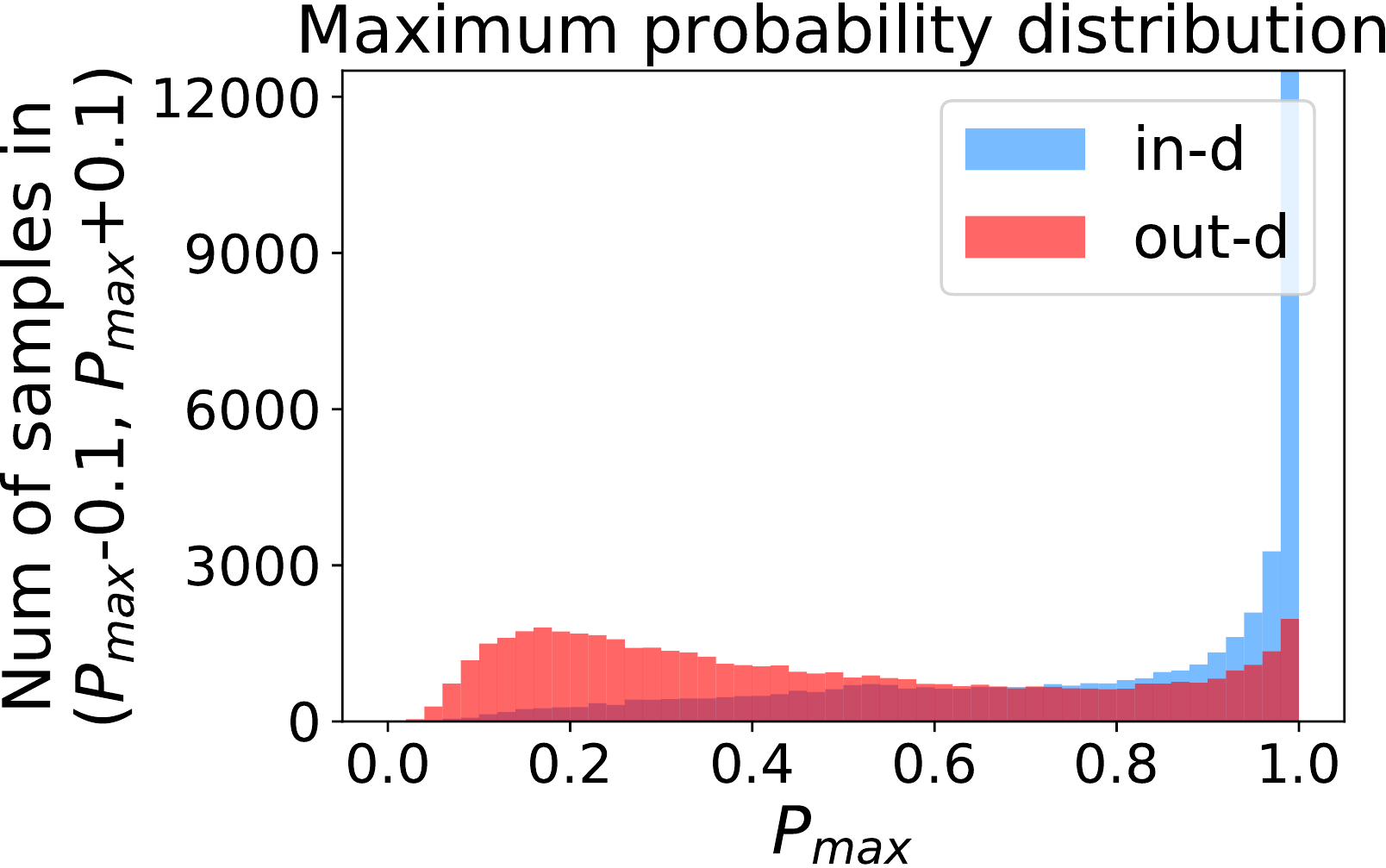}
		\end{minipage}%
	}
		%\caption{fig2}
	\subfigure[\label{fig:slice}]{
		\begin{minipage}[t]{0.30\linewidth}
			\centering
			\includegraphics[width=\linewidth]{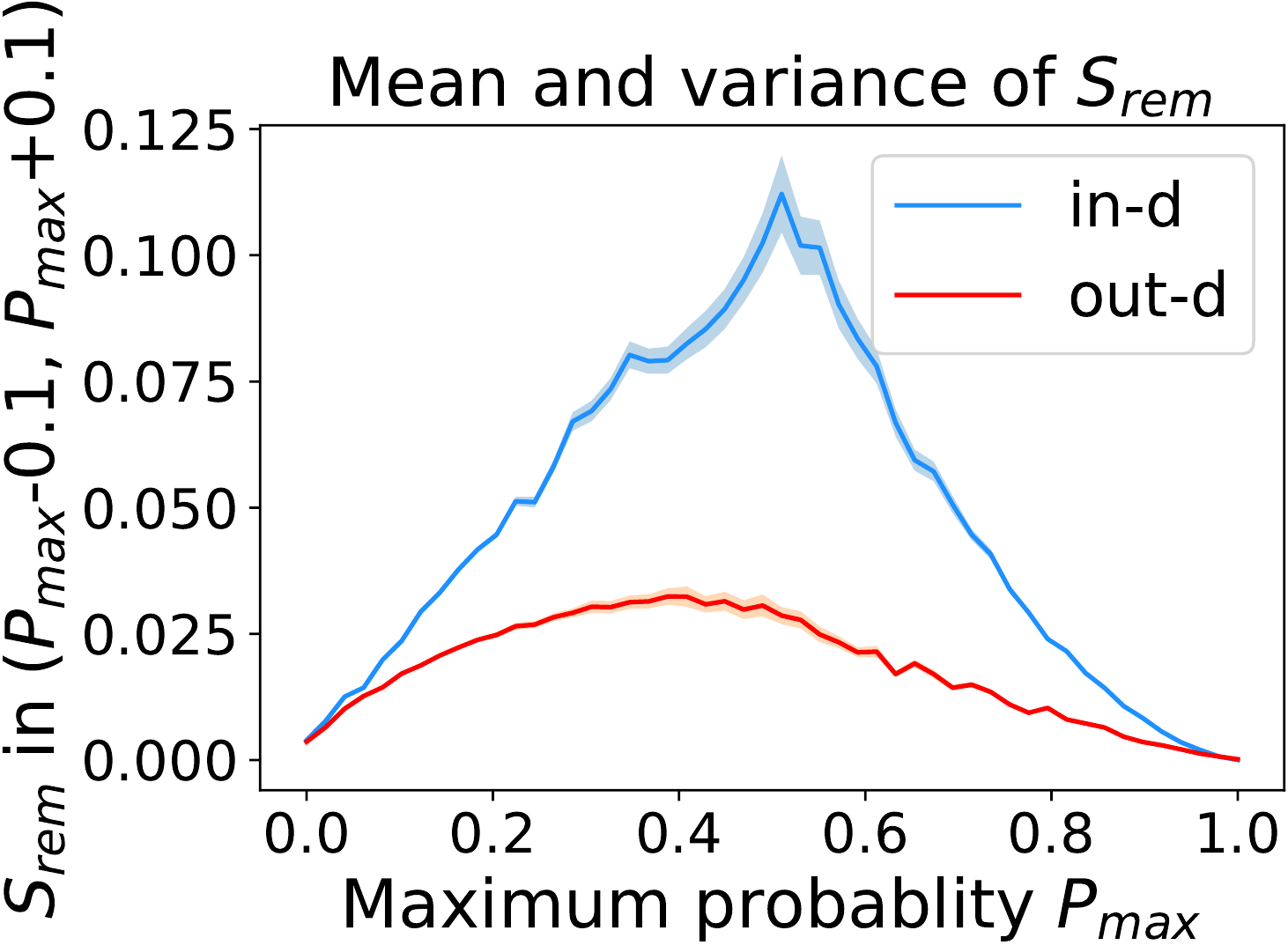}
		\end{minipage}%	
	}
	\subfigure[\label{fig:inner_product}]{
		\begin{minipage}[t]{0.32\linewidth}
			\centering
			\includegraphics[width=\linewidth]{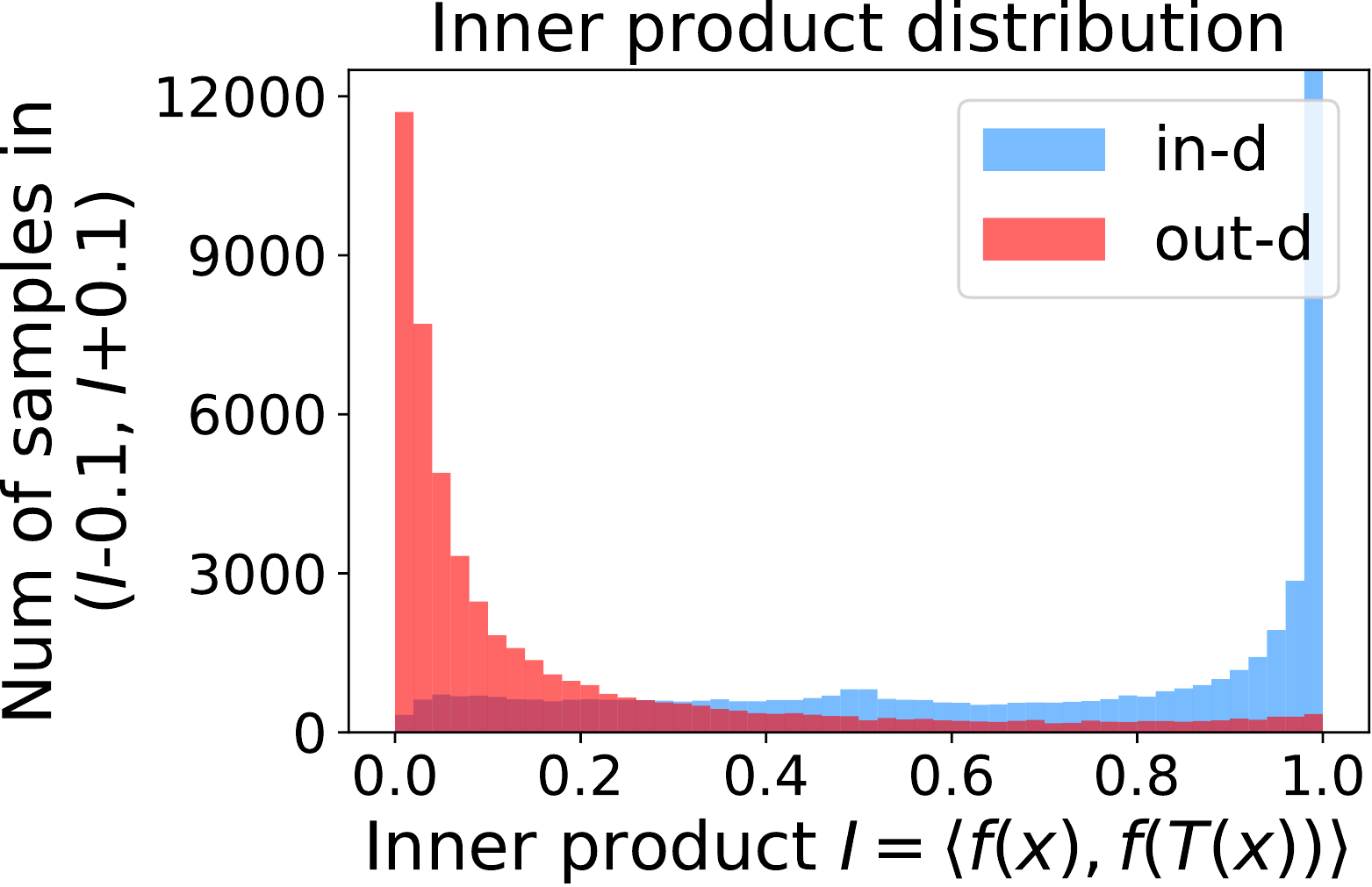}
		\end{minipage}%	
	}
	\centering
	\caption{\label{fig:empirical_evidence} Figures showing the effect of remaining classes of a pre-trained ResNet-50. Full ImageNet validation set and the Artificial dataset~(see Section~\ref{sec:dataset}) are treated as in- and out-distributions respectively. (a) Maximum probability distributions of in- and out-distribution samples. (b) Mean and variance of remaining scores within each slot $(P_{max}-\epsilon, P_{max}+\epsilon)$. (c) Distributions of inner product $I$ of in- and out-distribution samples. Note that our anomaly score $S(x)=1-I$.}
\end{figure}

%\begin{equation}
%	S_{remain} = \langle f(x), f(D(x)) \rangle - [\max_j f_j(x)] [\max_j(f_j(D(x)))] \label{eqn:Sremain}
%\end{equation}

Figure~\ref{fig:empirical_evidence} shows the beneficial effect of the remaining classes. First, Figure~\ref{fig:max_probability} presents the distribution of maximum probability. Then, we calculate the mean and variance of the remaining scores $S_{rem}$ in each evenly divided interval, as Figure~\ref{fig:slice} shows. To draw the Figure~\ref{fig:slice}, we first evenly divide the interval $(0,1)$ in Figure~\ref{fig:max_probability} into 50 parts. Then we pick one interval $(P_{max}-\epsilon, P_{max}+\epsilon)$, select all the samples whose maximum predicted probability is between $(P_{max}-\epsilon, P_{max}+\epsilon)$, and compute the mean and variance of those selected samples' remaining scores $S_{rem}$. For every interval, we process the in- and out-distribution data separately indicated by the red and blue lines. 

Intuitively, samples from both in- and out-distributions with similar $P_{max}$ values are mixed within one same interval~(Figure~\ref{fig:max_probability}). Figure~\ref{fig:slice} shows that for samples within the same interval, the remaining scores of the in-distribution samples are statistically larger than that of out-distribution samples. Therefore, adding the remaining scores to the data within the same interval in Figure~\ref{fig:max_probability} makes the in- and out-distribution samples distinguishable. When this process happens in each interval, we obtain our inner product scores~($1-S(x)$) as depicted in Figure~\ref{fig:inner_product}, which improves the maximum probability method from Figure~\ref{fig:max_probability}. 

Comparing the maximum probability~(Figure~\ref{fig:max_probability}) and consistency evaluation methods~(Figure~\ref{fig:inner_product}), our method is more discriminative. Generally, a larger gap in Figure~\ref{fig:slice} indicates a higher performance improvement. This empirical analysis is consistent with our numerical results in Section~\ref{sec:exp}.  
Similar observations across network architectures and datasets can be found in Appendix~\ref{app:remainingscore}.

\subsection{Theoretical analysis based on runs}
\label{sec:theoexplain}
The most fundamental and common part of anomaly detection is attaching a real-valued score calculated by the algorithm to each sample. We then obtain a sequence by sorting all samples according to their scores from small to large. If we label the out- and in-distribution samples as 0 and 1, respectively, we will finally get a mixing binary sequence. Intuitively, a sequence with good sorting~(like $00\dots011\dots1$) is preferred, but how to quantitatively describe the mixing degree? Empirically, we use AUROC but its discontinuity hinders the development of anomaly detection theory. To analyze our algorithm, we propose to use expected runs~\cite{wald1940test, barton1957multiple} as a continuous surrogate to approximately measure the confusion degree of a binary sequence:

%From the perspective of the calculation, the ultimate goal of the out-distribution samples detection task is to obtain a well sorted sequence of the given samples. The sequence is sorted by the real-valued score calculated by our algorithm~(Eqn~\eqref{eqn:oodscore}) for each sample. And we hope the scores of out-distribution samples are statically different from that of in-distribution samples.

%To quantitatively analyze our algorithm, we use \emph{runs}
%\footnote{\url{https://en.wikipedia.org/wiki/Wald\%E2\%80\%93Wolfowitz_runs_test}}
%to approximately measure the order confusion of a binary~() sequence: 
%Area Under Receiver Operating Characteristic~(AUROC) is commonly used to assess anomaly detection algorithms empirically. Considering the hardness of ROC analysis, we propose to analyze its surrogate: \emph{runs}. 

\begin{definition}[Runs]
	\label{Def: runs}
	The runs number of a binary sequence is the number of maximal non-empty segments consisting of adjacent equal elements. Each segment is called a run.
	%footnote wiki?
\end{definition}

For example, the sequence $\underline{00}$ $\underline{111}$ $ \underline{000}$ $\underline{11}$ $\underline{000}$ has 5 runs. Ideally, the in-distribution and out-distribution samples are completely separable~($\underline{00...0}$ $\underline{11...1}$), so the runs number is 2. On the contrary, the runs number of the worst case, when the two kinds of samples alternate in the sequence, equals the sequence length. 

Assume the in-distribution and out-distribution sample scores are drawn from two distributions. Then the expected runs number evaluates the similarity of the two distributions. The above two examples suggest that more separable distributions lead to a smaller expected runs number. 
%In general, when the sequence length is fixed, a smaller runs number leads to a better ordering of in-distribution and out-distribution samples, which results in the detection of out-distribution samples.
%\textbf{The calculation of the expected runs.}
%Assume the in-distribution and out-distribution samples are drawn from two distributions. 
Now we show how to calculate the expected runs number for any given two distributions in Lemma~\ref{lem:expectedRuns}.

\begin{lemma}[Calculation of the expected runs number]
	\label{lem:expectedRuns}
	Consider two random variables $X_1$ with PDF $f(x)$ and $X_2$ with PDF $g(x)$.
	Assume for $x \in \mathcal{X}$, $f(x)+g(x)>0$, and the length of the support $|\mathcal{X}|$ is upper bounded by $M$. 
	We assume the first-order derivative is bounded by $\firstOrderBound$, namely $f^\prime(x) \leq C$ and $g^\prime(x) \leq C$.
	Then, for $n_1$ samples from distribution $f(x)$ and $n_2$ samples from distribution $g(x)$, if $n_1, n_2 \to \infty$ while $0<n_1/n_2 = \kappa$, we have 
	$$
	\frac{\mathbb{E} R}{\sqrt{n_1n_2}} = \int_x \frac{\sqrt{n_1 n_2} f(x) g(x)}{n_1 f(x)+ n_2 g(x)} dx + o_{\sqrt{n_1n_2}}((MC)^{1/2}).
	$$
\end{lemma}
In the rest of the paper, we only consider the integral term while ignoring the little $o$ term. 

%Runs number is an approximation of the Receiver Operating Characteristic~(ROC) which is used to evaluate the anomaly detection's performance. Lemma~\ref{lem: equ} illustrates this point from the perspective of the maximal runs number. \hhw{wording}
%
%\begin{lemma}[Maximal runs number]
%	\label{lem: equ}
%	Consider two distributions $f(x)$ and $g(x)$, then the expected runs number reaches the maximum when $f(x) = g(x)$ almost surely.
%\end{lemma}
%
%More details can be found in Appendix~\ref{app:lemma2}.
%we derive a fundamental property of the expected runs. Specifically, we prove that when $g(x) = f(x)$, the expected runs reach the maximum. Besides, Lemma~\ref{lem: equ} also validates the fact that the expected runs is a reasonable surrogate to the ROC.

%\textbf{Expected runs on Beta distribution.}
We will derive a sufficient condition based on the expected runs number to identify our algorithm's effectiveness. The supports of $f(x)$ and $g(x)$ are $[0,1]$.
Considering the two distributions in real datasets~(see Figure~\ref{fig:beta_inner_product} in Appendix~\ref{app:lemma3}), we model the two distributions as Beta distributions and denote the runs number as $R (\alpha_1, \beta_1, \alpha_2, \beta_2)$. Lemma~\ref{lem: change} investigates how the Beta distribution parameters $\alpha_1, \beta_1, \alpha_2, \beta_2$ increases/decreases the expected runs number.

% Therefore, it is meaningful to calculate derivations.

\begin{lemma}[Runs number increases/decreases]
	\label{lem: change}
	When $ \alpha_2 - \alpha_1 \leq 0$, we have $\frac{\partial \mathbb{E} R (\alpha_1, \beta_1, \alpha_2, \beta_2)}{\partial \alpha_1} \leq 0$.
	When $ \alpha_2 - \alpha_1 \geq  \sum_{k=0}^{\lfloor \beta_1 \rfloor} \frac{1}{\alpha_1 + k}$ , we have $\frac{\partial \mathbb{E} R (\alpha_1, \beta_1, \alpha_2, \beta_2)}{\partial \alpha_1} \geq 0$.
	By symmetry, $\beta_1, \beta_2, \alpha_2$ performs similarly.
\end{lemma}

%Lemma~\ref{lem: change} shows a sufficient condition to compare the expected runs number between different parameters in Beta distributions. 
For example, when $\alpha_1>\alpha_2$, we see $\mathbb{E} R(\alpha_1, \beta_1, \alpha_2, \beta_2) \geq \mathbb{E} R(\alpha_1+1, \beta_1, \alpha_2, \beta_2)$.
We leave the necessary and sufficient conditions for future work. Limited to space, we put the detailed proof and statements in Appendix~\ref{app:runs}. 

More generally, the expected runs provides a general theoretical surrogate for anomaly detection. In order to analyze the pros and cons of the algorithm performance, we can model the distribution of anomaly scores of in- and out-distribution, compute the runs, and derive runs with respect to distribution parameters. This will let us know how the distributions affect the runs number, which may guide designing and tuning the algorithms.

\section{Limitation}
\label{sec:limitation}
Although TTA-AD is effective and easy to be implemented, it has its inherit limitations. 
%Due to the transformation methods for language and tabular datasets are significantly different from vision datasets, it is unpredictable to implement TTA-AD on those tasks. We emphasize that this is a common case since a series of current works are focusing on vision tasks only~\cite{ODIN, Ma, DBLP:conf/nips/HendrycksMKS19, CSI}. 
TTA-AD may fail under some adversarially designed synthetic datasets, e.g., a transformation-invariant vision dataset.
Additionally, the goal of proposing runs based theoretical analysis is to reduce the gap between theoretical analysis and practical metrics of anomaly detection. There is still some future work to refine this theoretical framework. 
%Meanwhile, there are other vision anomaly detection tasks except for the multi-class settings, but TTA-AD can not adapt to those settings since current version of TTA-AD is designed for multi-class only. 
%In fact, the multi-class vision anomaly detection is popular in existing researches~\cite{AnomalydetectionAsurvey, DeepLearningforAnomalyDetectionAReview, ODIN, Ma, DBLP:conf/nips/HendrycksMKS19, CSI}. 

\section{Conclusion}
\label{sec:conclusion}
In this paper, we propose a simple yet effective dataset-vs-dataset vision post-hoc anomaly detection algorithm TTA-AD, which reaches comparable or better detection performance with current advanced and SOTA algorithms. 
It largely reduces the computational costs, demands no specific network architectures, no access to the internal feature, and no prior out-distribution data knowledge. 
We empirically show that TTA-AD gains performance improvement from the remaining classes, which are ignored by some previous works. 
Meanwhile, we provide a general theoretical framework based on runs to serve as a surrogate for reducing the gap between theoretical analysis and practical applications in anomaly detection.

\bibliography{neurips_2022}
\bibliographystyle{plain}

%%%%%%%%%%%%%%%%%%%%%%%%%%%%%%%%%%%%%%%%%%%%%%%%%%%%%%%%%%%%
\section*{Checklist}

%%%% BEGIN INSTRUCTIONS %%%
%The checklist follows the references.  Please
%read the checklist guidelines carefully for information on how to answer these
%questions.  For each question, change the default \answerTODO{} to \answerYes{},
%\answerNo{}, or \answerNA{}.  You are strongly encouraged to include a {\bf
%	justification to your answer}, either by referencing the appropriate section of
%your paper or providing a brief inline description.  For example:
%\begin{itemize}
%	\item Did you include the license to the code and datasets? \answerYes{See Section~\ref{gen_inst}.}
%	\item Did you include the license to the code and datasets? \answerNo{The code and the data are proprietary.}
%	\item Did you include the license to the code and datasets? \answerNA{}
%\end{itemize}
%Please do not modify the questions and only use the provided macros for your
%answers.  Note that the Checklist section does not count towards the page
%limit.  In your paper, please delete this instructions block and only keep the
%Checklist section heading above along with the questions/answers below.
%%%% END INSTRUCTIONS %%%

\begin{enumerate}

	\item For all authors...
	\begin{enumerate}
		\item Do the main claims made in the abstract and introduction accurately reflect the paper's contributions and scope?
		\answerYes{}
		\item Did you describe the limitations of your work?
		\answerYes{See Section~\ref{sec:limitation}.}
		\item Did you discuss any potential negative societal impacts of your work?
		\answerNA{}
		\item Have you read the ethics review guidelines and ensured that your paper conforms to them?
		\answerYes{}
	\end{enumerate}

	\item If you are including theoretical results...
	\begin{enumerate}
		\item Did you state the full set of assumptions of all theoretical results?
		\answerYes{See Section~\ref{sec:theoexplain} and Appendix~\ref{app:runs}.}
		\item Did you include complete proofs of all theoretical results?
		\answerYes{See Appendix~\ref{app:runs}.}
	\end{enumerate}

	\item If you ran experiments...
	\begin{enumerate}
		\item Did you include the code, data, and instructions needed to reproduce the main experimental results (either in the supplemental material or as a URL)?
		\answerYes{See Appendix~\ref{app:attempts_to_improve}.}
		\item Did you specify all the training details (e.g., data splits, hyperparameters, how they were chosen)?
		\answerYes{See Appendix~\ref{app:attempts_to_improve}.}
		\item Did you report error bars (e.g., with respect to the random seed after running experiments multiple times)?
		\answerYes{Results are reported with mean and variance.}
		\item Did you include the total amount of compute and the type of resources used (e.g., type of GPUs, internal cluster, or cloud provider)?
		\answerYes{See Section~\ref{sec:algorithms_running_time}.}
	\end{enumerate}

	\item If you are using existing assets (e.g., code, data, models) or curating/releasing new assets...
	\begin{enumerate}
		\item If your work uses existing assets, did you cite the creators?
		\answerYes{}
		\item Did you mention the license of the assets?
		\answerYes{}
		\item Did you include any new assets either in the supplemental material or as a URL?
		\answerNA{}
		\item Did you discuss whether and how consent was obtained from people whose data you're using/curating?
		\answerNA{}
		\item Did you discuss whether the data you are using/curating contains personally identifiable information or offensive content?
		\answerNA{}
	\end{enumerate}

	\item If you used crowdsourcing or conducted research with human subjects...
	\begin{enumerate}
		\item Did you include the full text of instructions given to participants and screenshots, if applicable?
		\answerNA{}
		\item Did you describe any potential participant risks, with links to Institutional Review Board (IRB) approvals, if applicable?
		\answerNA{}
		\item Did you include the estimated hourly wage paid to participants and the total amount spent on participant compensation?
		\answerNA{}
	\end{enumerate}

\end{enumerate}

%%%%%%%%%%%%%%%%%%%%%%%%%%%%%%%%%%%%%%%%%%%%%%%%%%%%%%%%%%%%

\newpage

\appendix

\section{Test time augmentation feature space under different data distributions}
\label{app:feature_space}
\begin{figure*}[h!]
	\centering
	\subfigure[$\text{FFT}_{22}$ transformantion on DenseNet-121 for CIFAR-10 vs. SVHN]{
		\begin{minipage}[t]{0.32\linewidth}
			\centering
			\includegraphics[width=1.8in]{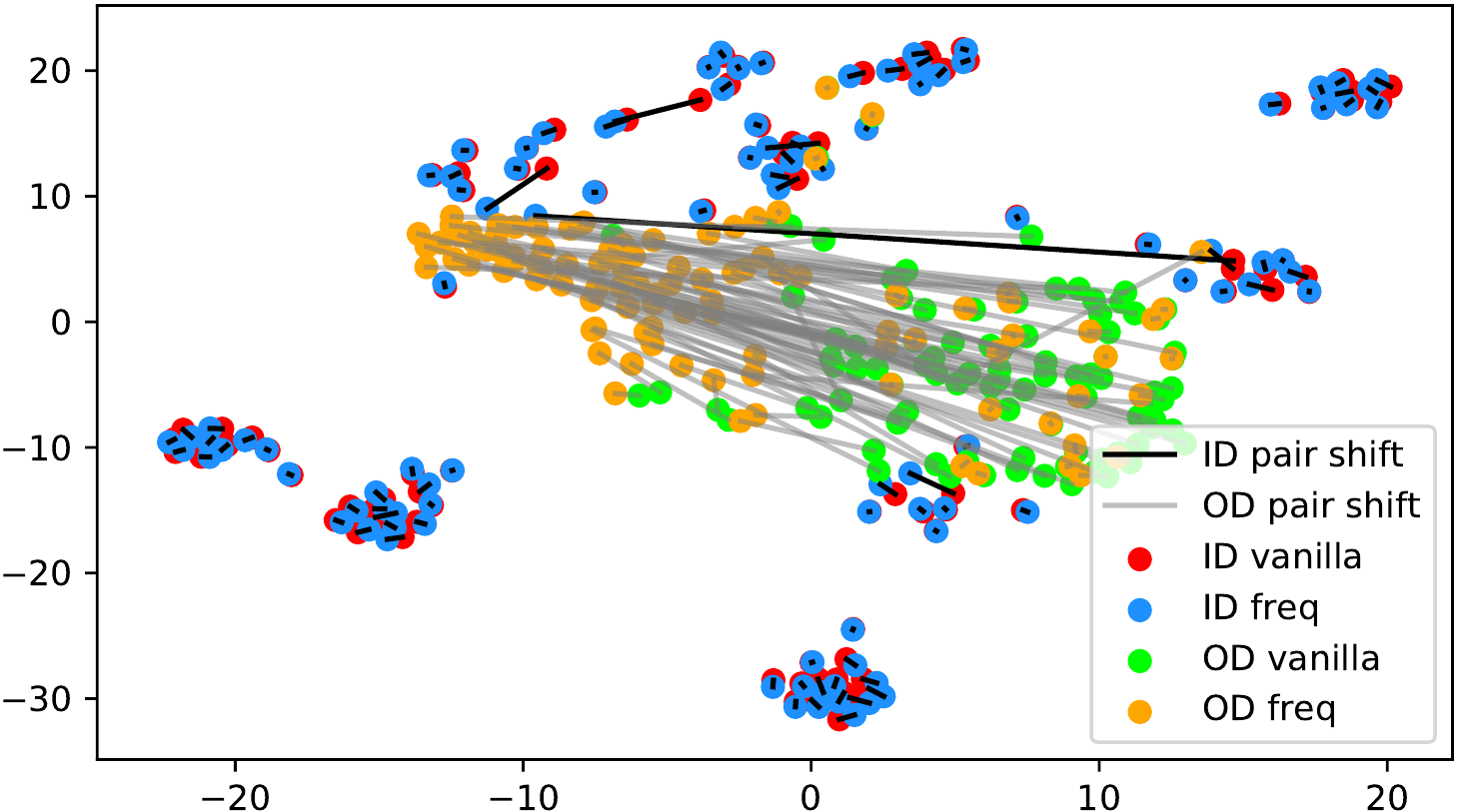}
			%\caption{fig1}
		\end{minipage}%
	}
	\subfigure[Flip transformantion on ResNet-34 for CIFAR-10 vs. SVHN]{
		\begin{minipage}[t]{0.32\linewidth}
			\centering
			\includegraphics[width=1.8in]{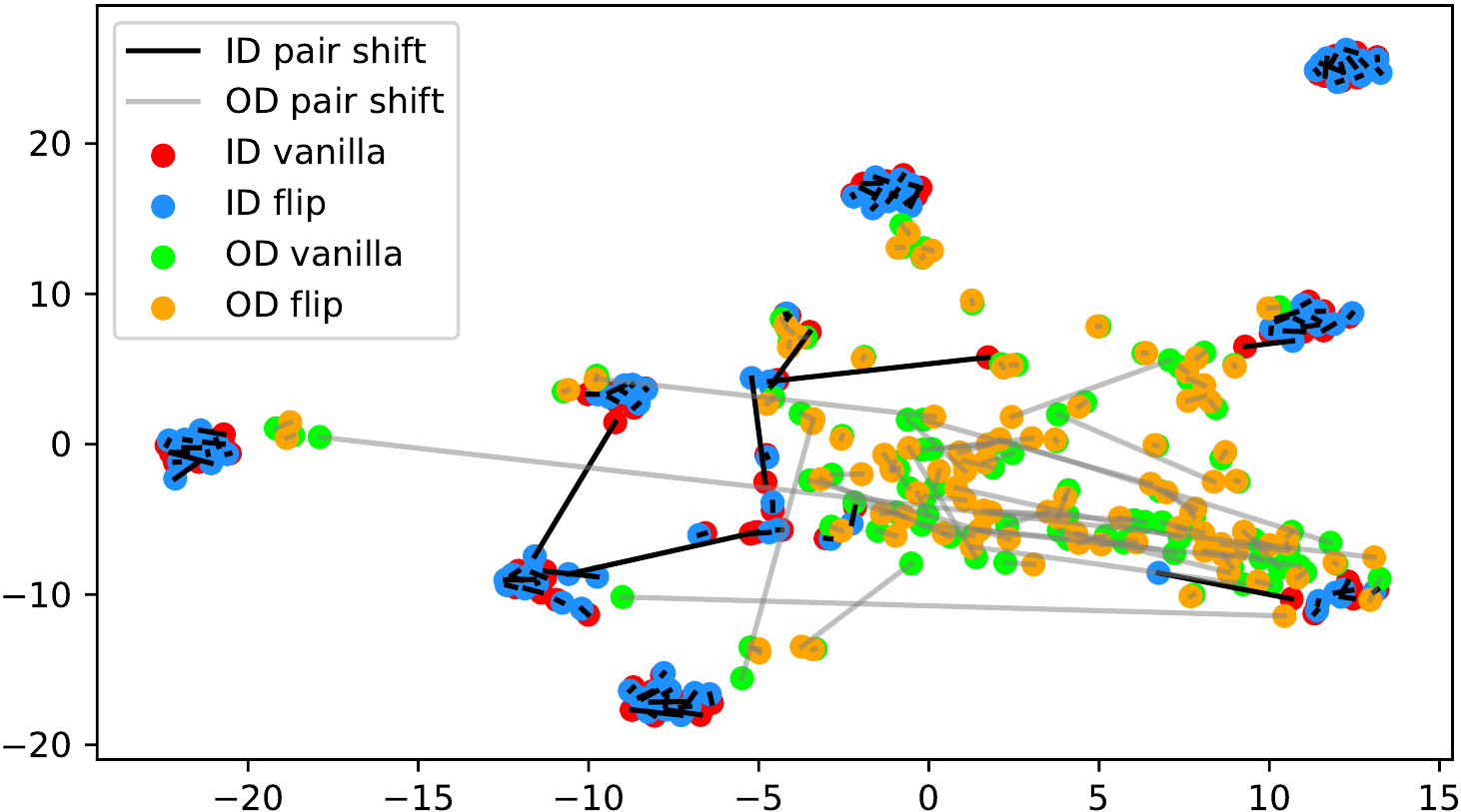}
			%\caption{fig1}
		\end{minipage}%
	}
	\subfigure[Flip transformantion on DenseNet-121 for CIFAR-10 vs. SVHN]{
		\begin{minipage}[t]{0.32\linewidth}
			\centering
			\includegraphics[width=1.8in]{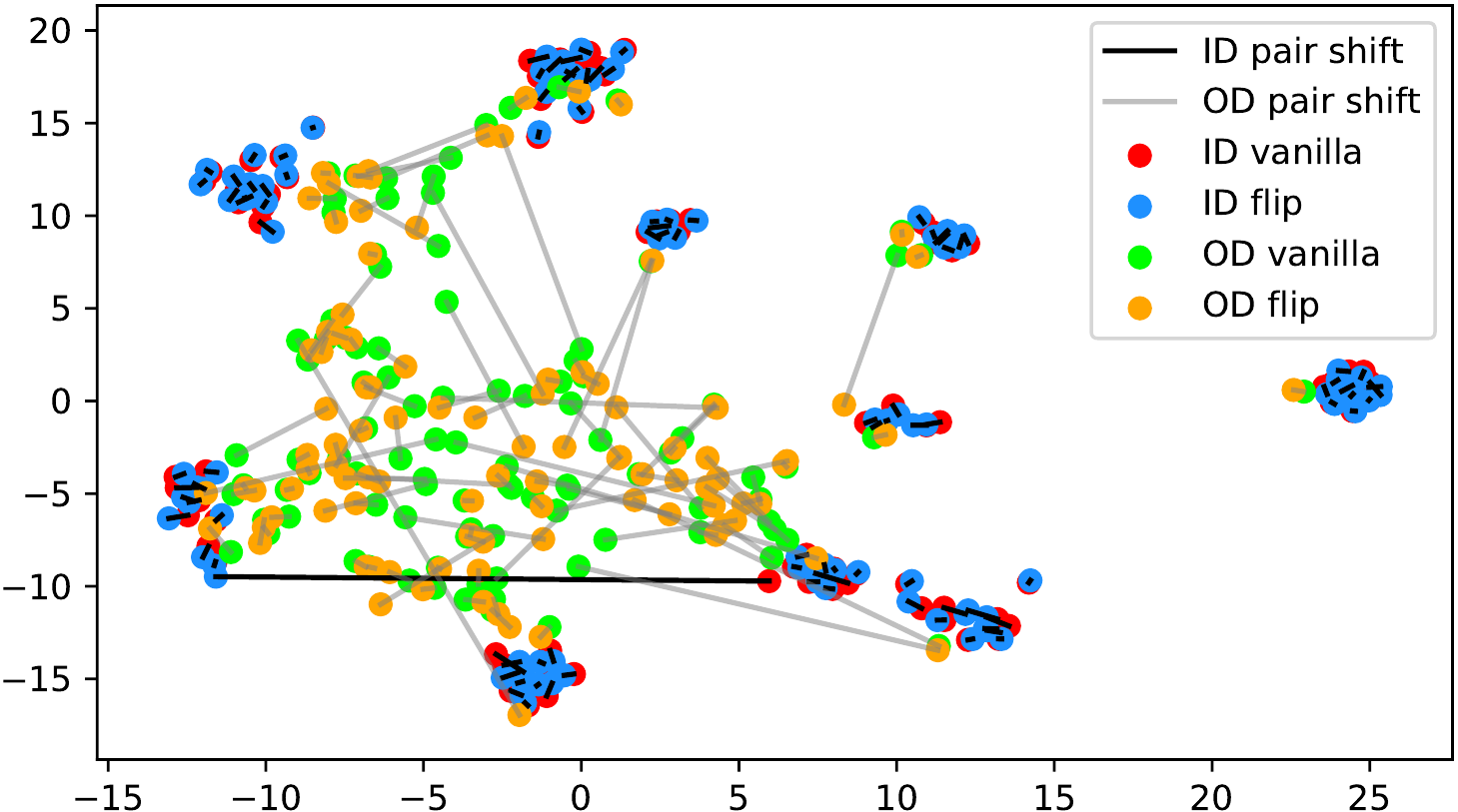}
			%\caption{fig1}
		\end{minipage}%
	}
	\\
	\vspace{-5px}
	\subfigure[$\text{FFT}_{40}$ transformantion on ResNet-50 for ImageNet Living 9 setting]{
		\begin{minipage}[t]{0.32\linewidth}
			\centering
			\includegraphics[width=1.8in]{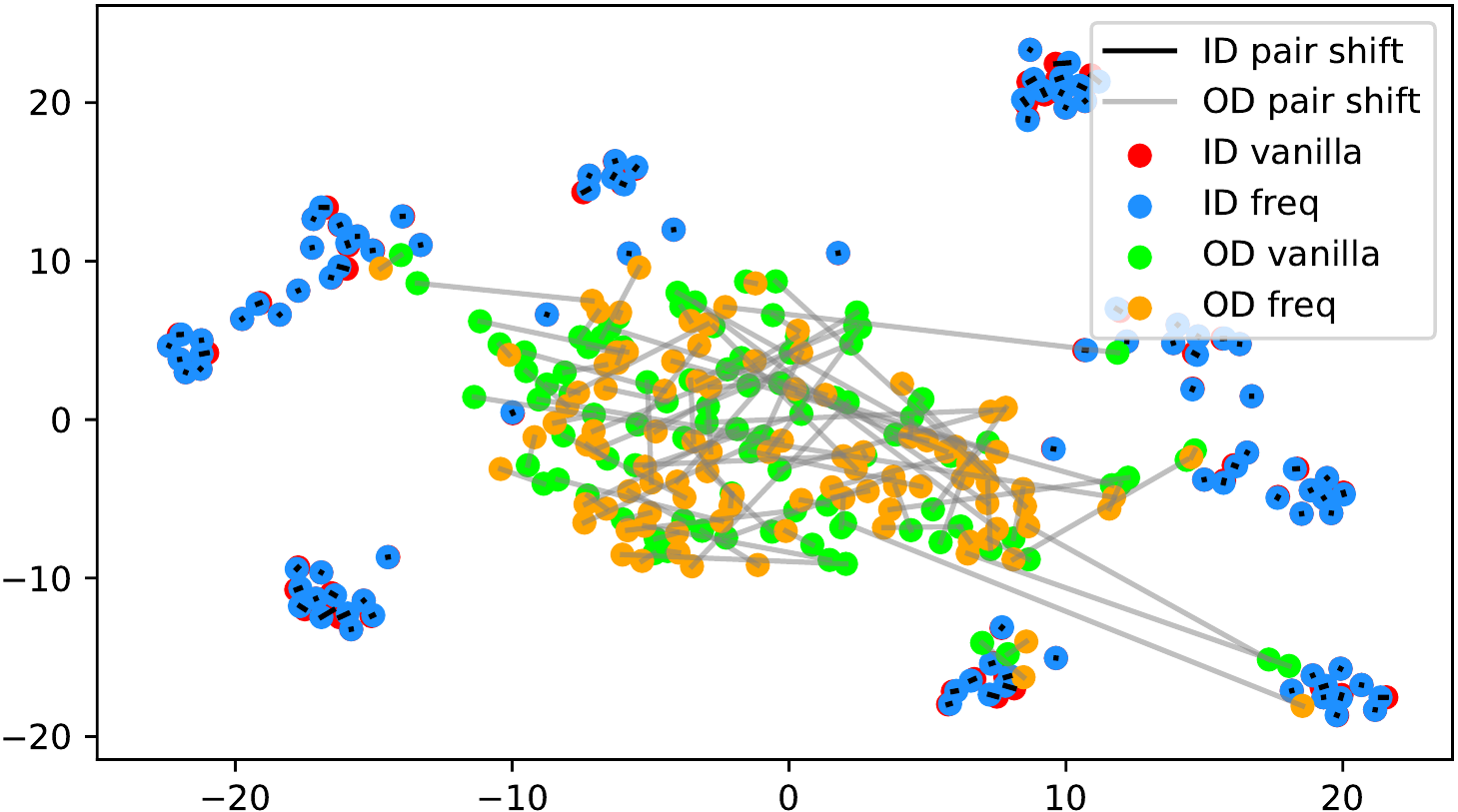}
			%\caption{fig1}
		\end{minipage}%
	}
	\subfigure[$\text{FFT}_{40}$ transformantion on ResNet-50 for ImageNet Geirhos 16 setting]{
		\begin{minipage}[t]{0.32\linewidth}
			\centering
			\includegraphics[width=1.8in]{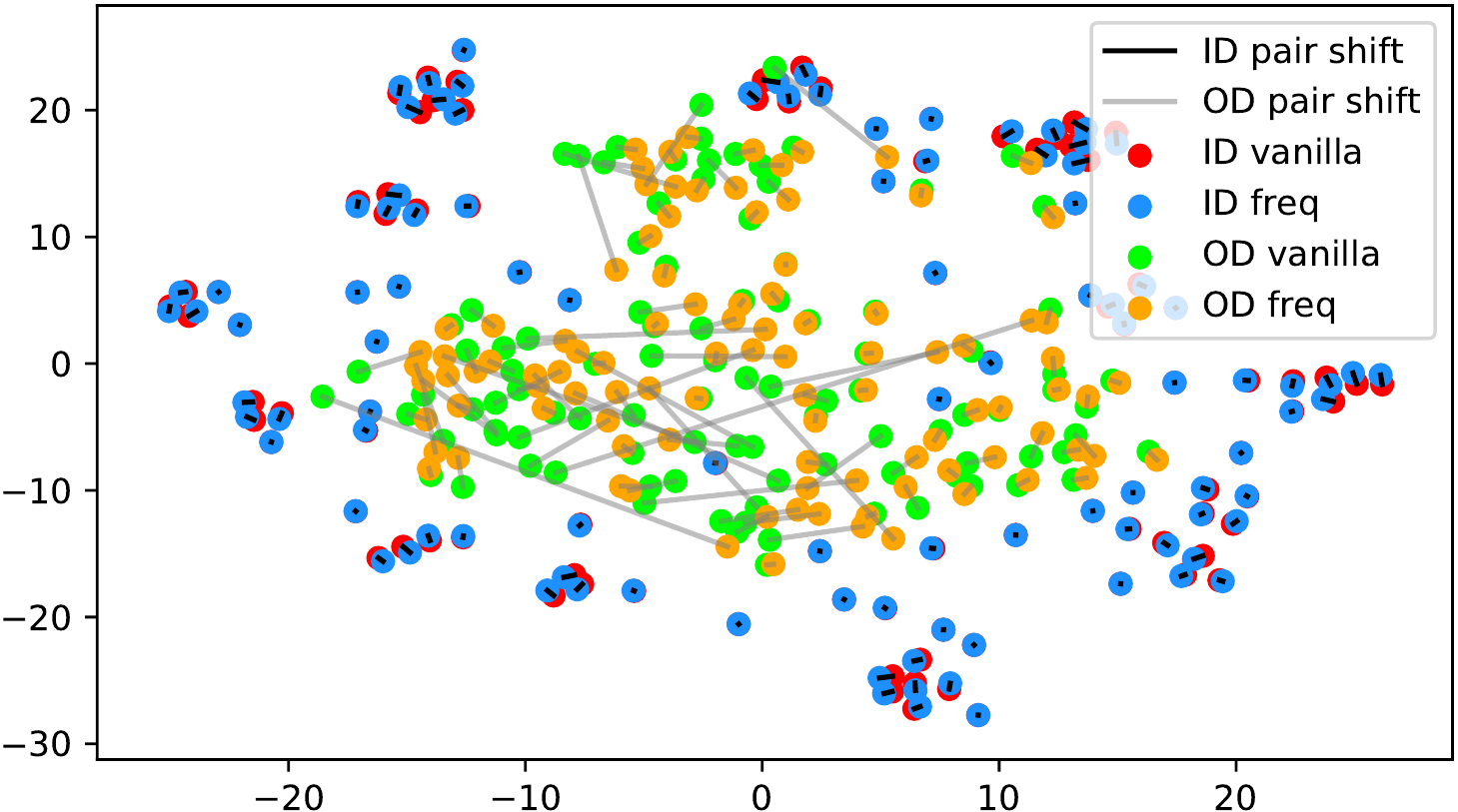}
			%\caption{fig1}
		\end{minipage}%
	}
	\subfigure[$\text{FFT}_{40}$ transformantion on ResNet-50 for ImageNet Mixed 10 setting]{
		\begin{minipage}[t]{0.32\linewidth}
			\centering
			\includegraphics[width=1.8in]{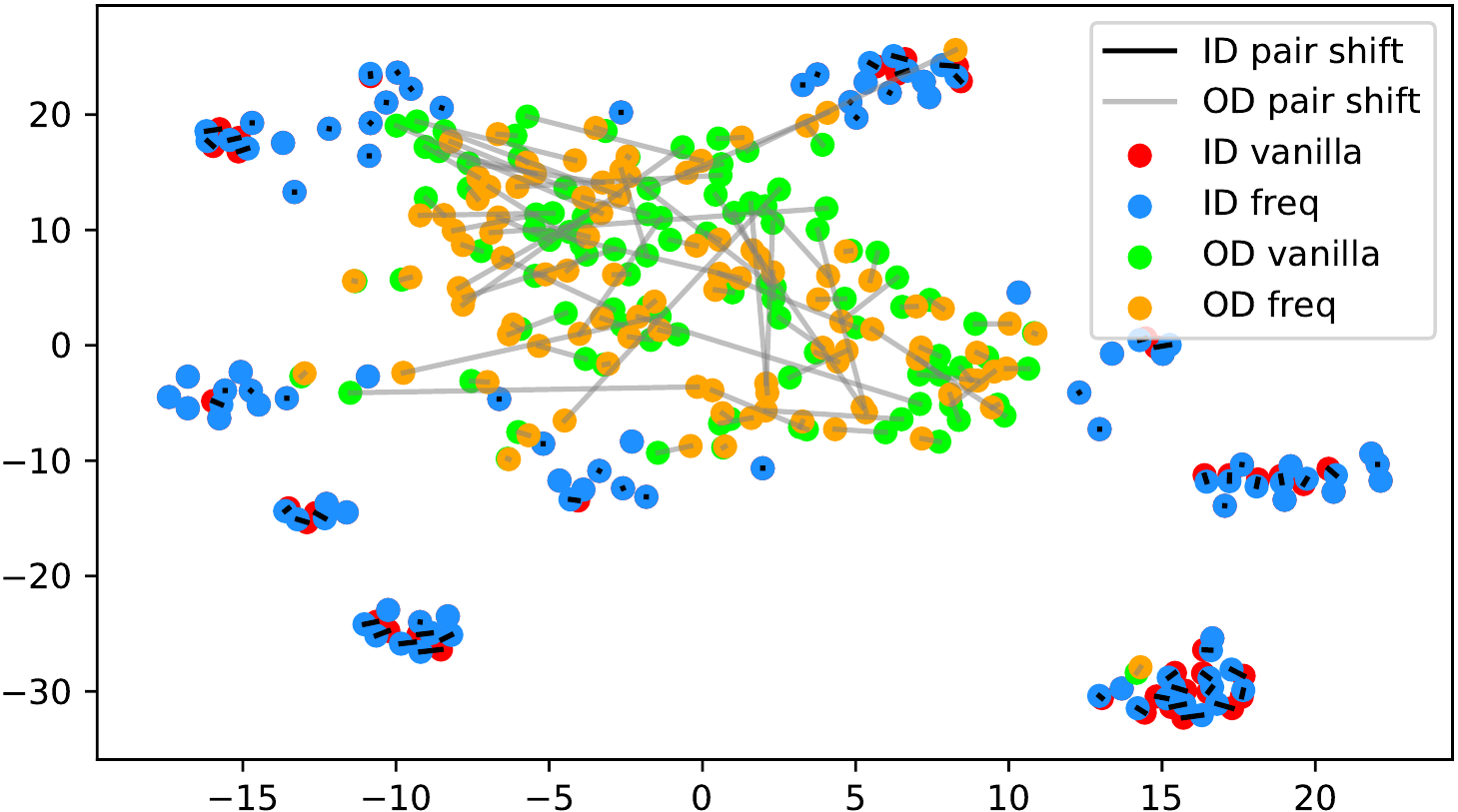}
			%\caption{fig1}
		\end{minipage}%
	}
	\\
	\vspace{-5px}
	\subfigure[Flip transformantion on ResNet-50 for ImageNet Living 9 setting]{
		\begin{minipage}[t]{0.32\linewidth}
			\centering
			\includegraphics[width=1.8in]{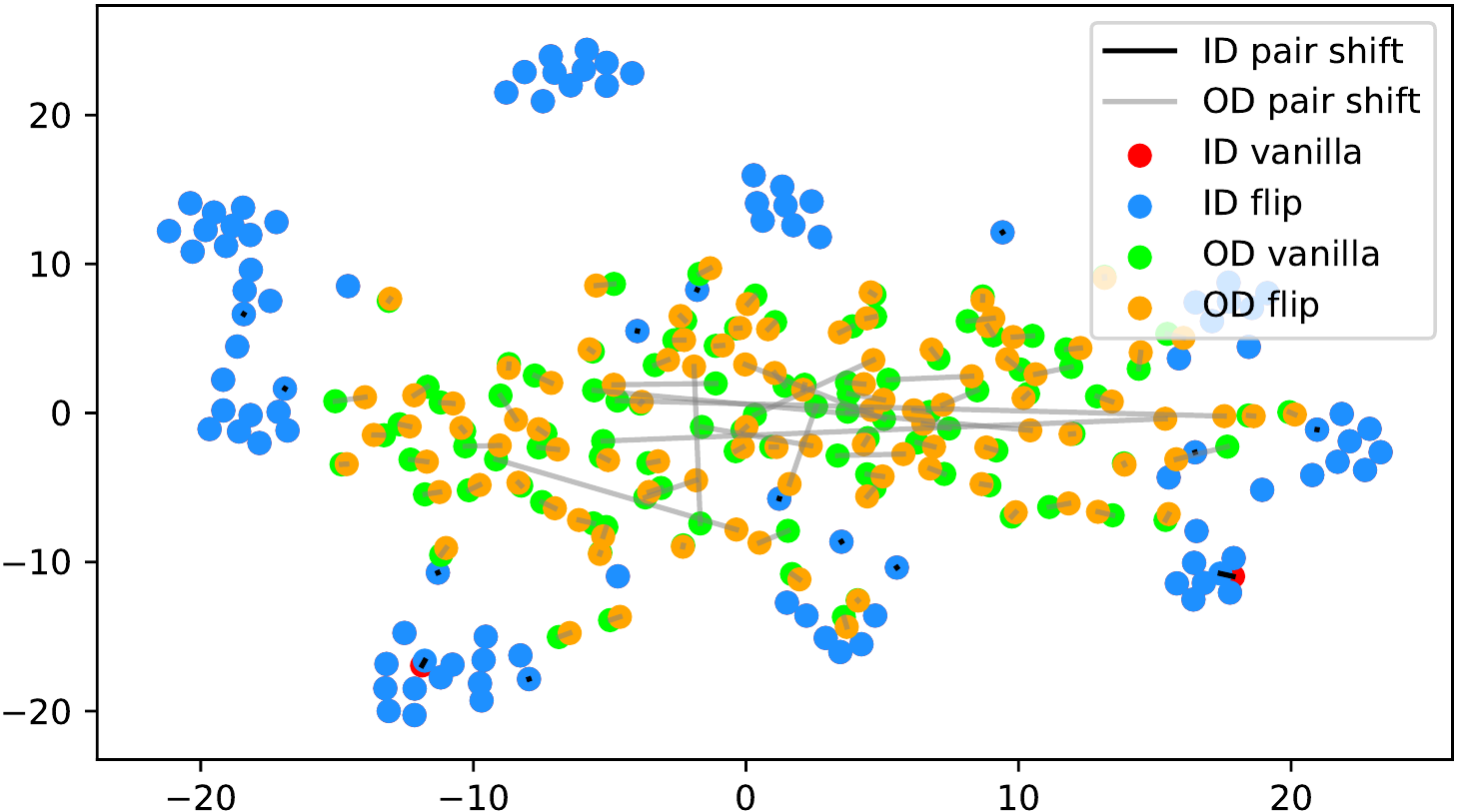}
			%\caption{fig1}
		\end{minipage}%
	}
	\subfigure[Flip transformantion on ResNet-50 for ImageNet Geirhos 16 setting]{
		\begin{minipage}[t]{0.32\linewidth}
			\centering
			\includegraphics[width=1.8in]{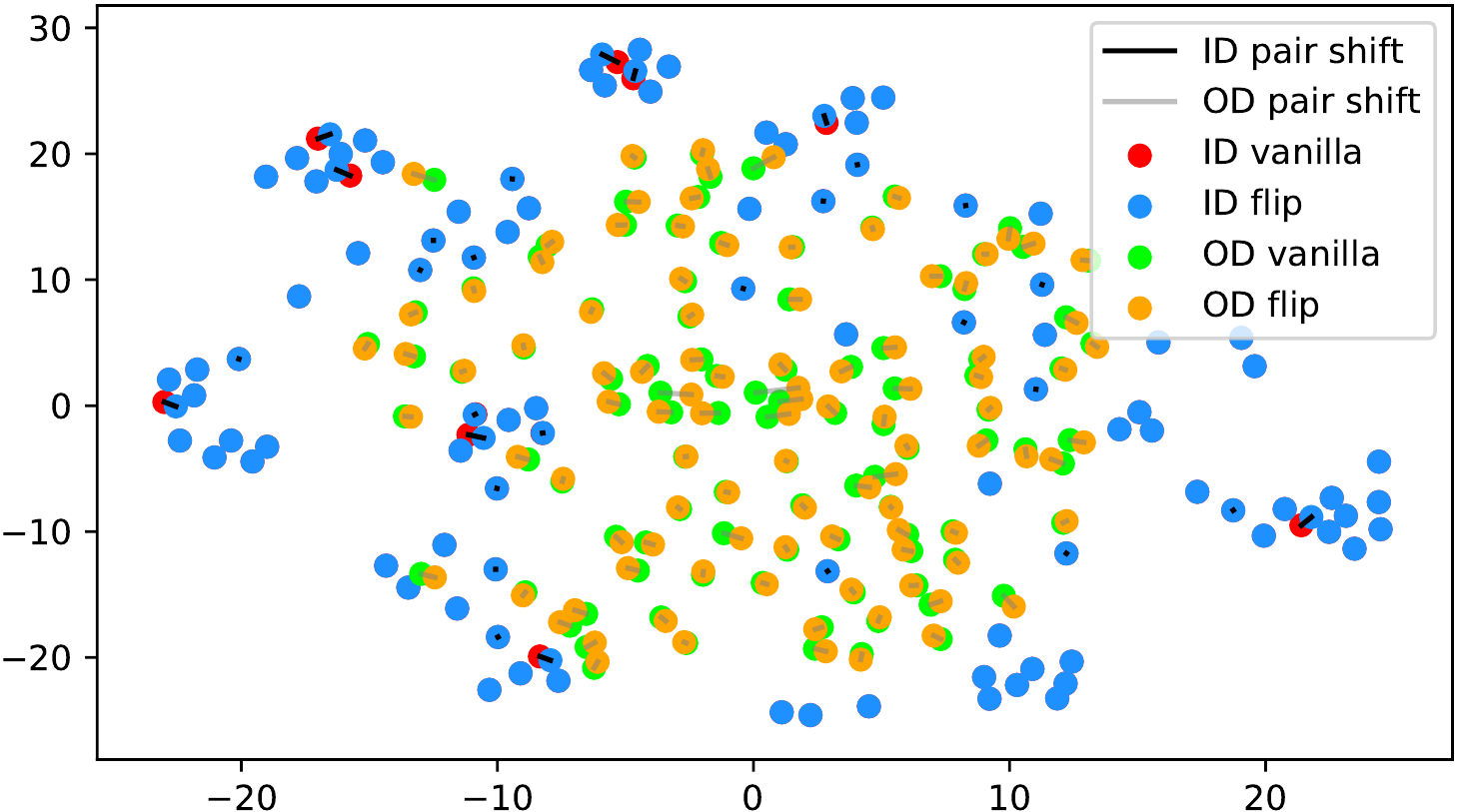}
			%\caption{fig1}
		\end{minipage}%
	}
	\subfigure[Flip transformantion on ResNet-50 for ImageNet Mixed 10 setting]{
		\begin{minipage}[t]{0.32\linewidth}
			\centering
			\includegraphics[width=1.8in]{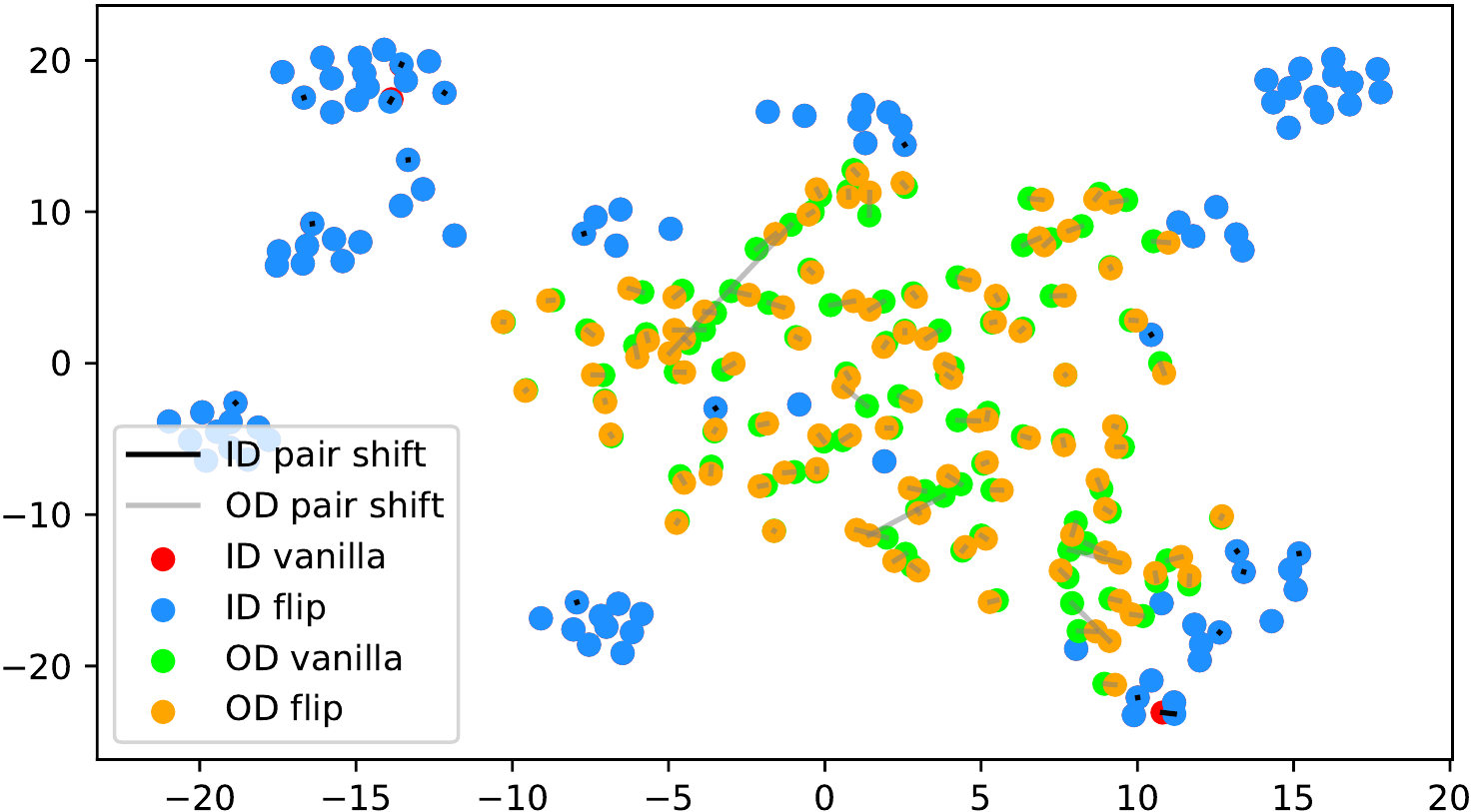}
			%\caption{fig1}
		\end{minipage}%
	}
	\\
	\vspace{-5px}
	\subfigure[$\text{FFT}_{40}$ transformantion on DenseNet-121 for ImageNet Living 9 setting]{
		\begin{minipage}[t]{0.32\linewidth}
			\centering
			\includegraphics[width=1.8in]{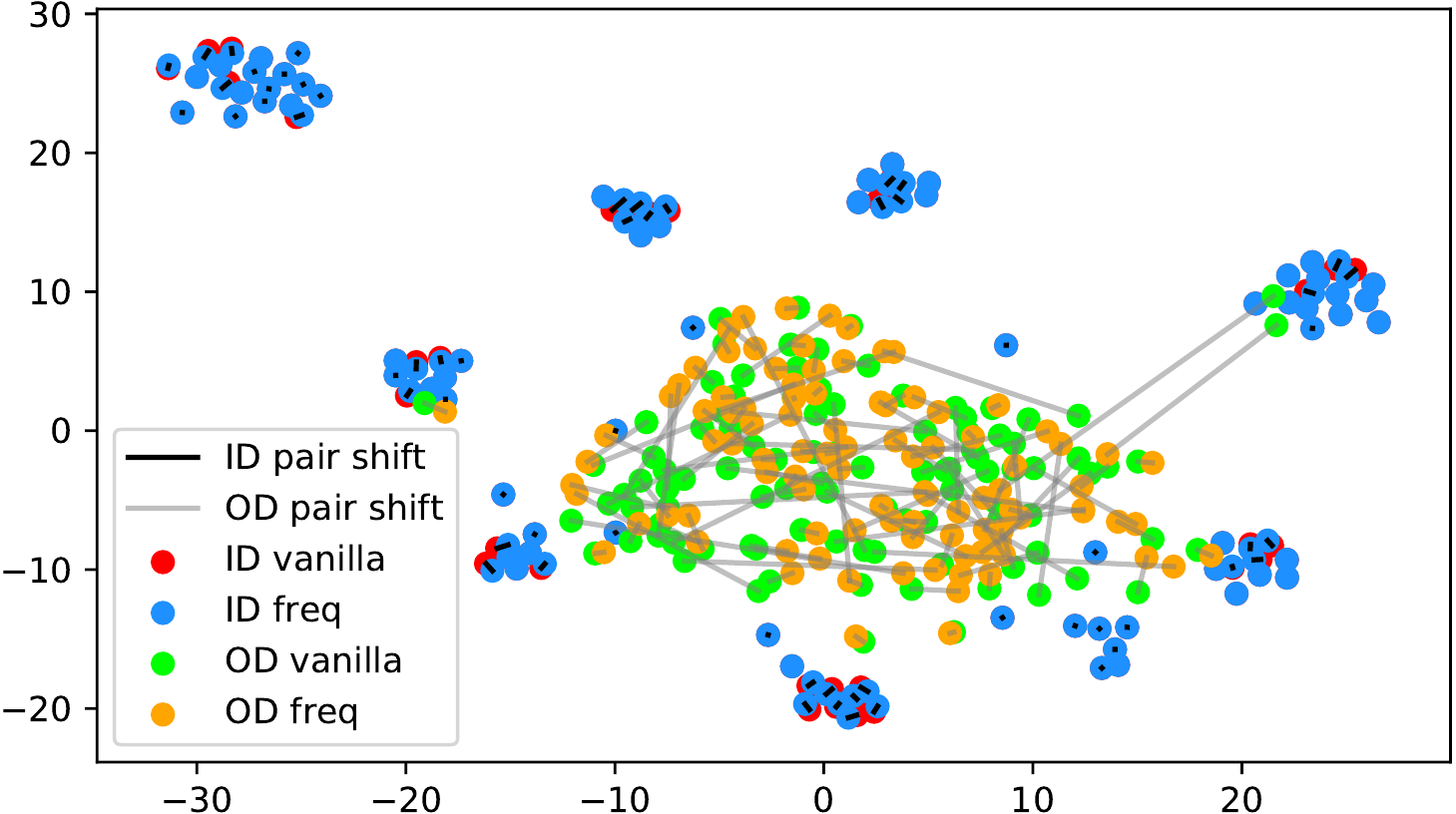}
			%\caption{fig1}
		\end{minipage}%
	}
	\subfigure[$\text{FFT}_{40}$ transformantion on DenseNet-121 for ImageNet Geirhos 16 setting]{
		\begin{minipage}[t]{0.32\linewidth}
			\centering
			\includegraphics[width=1.8in]{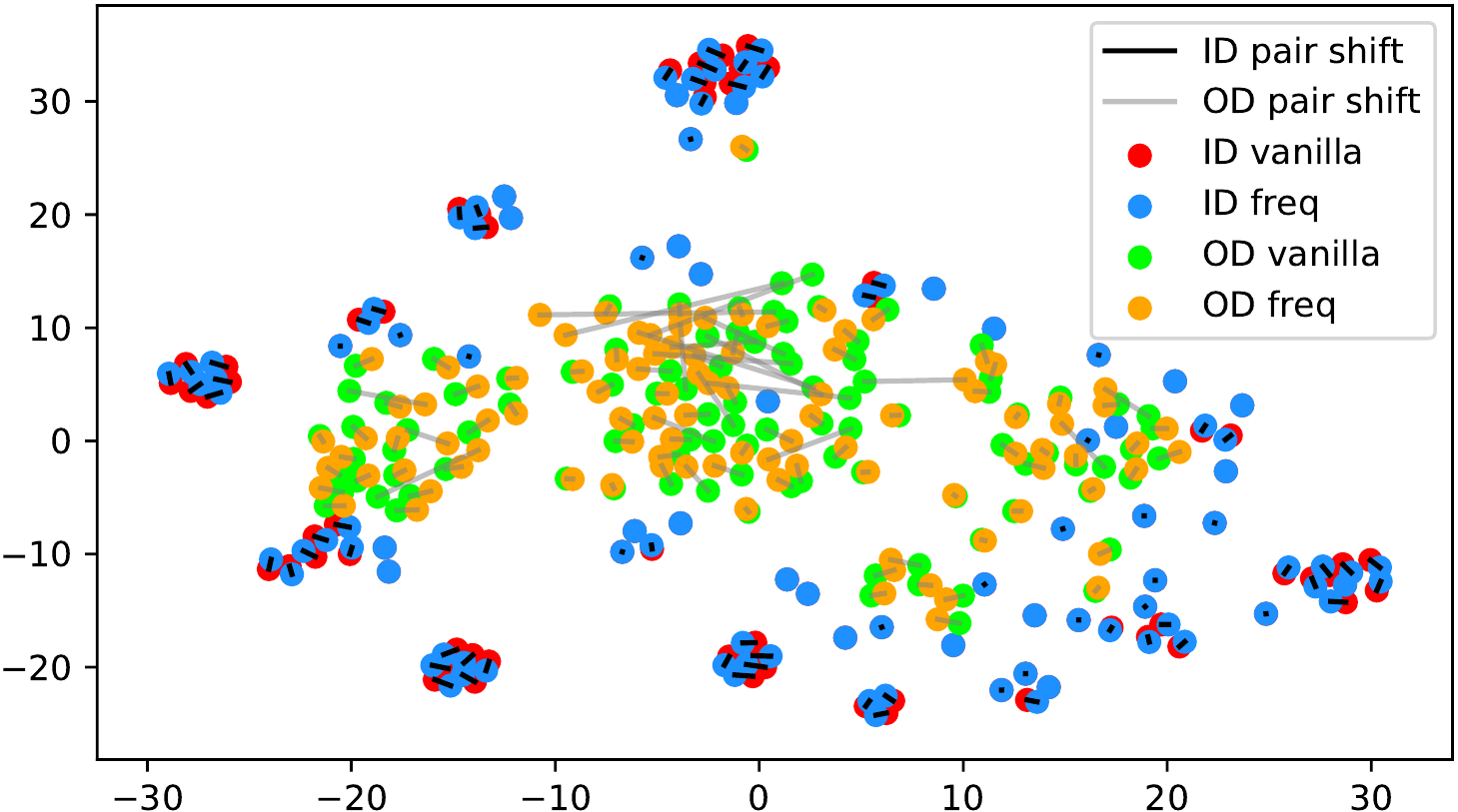}
			%\caption{fig1}
		\end{minipage}%
	}
	\subfigure[$\text{FFT}_{40}$ transformantion on DenseNet-121 for ImageNet Mixed 10 setting]{
		\begin{minipage}[t]{0.32\linewidth}
			\centering
			\includegraphics[width=1.8in]{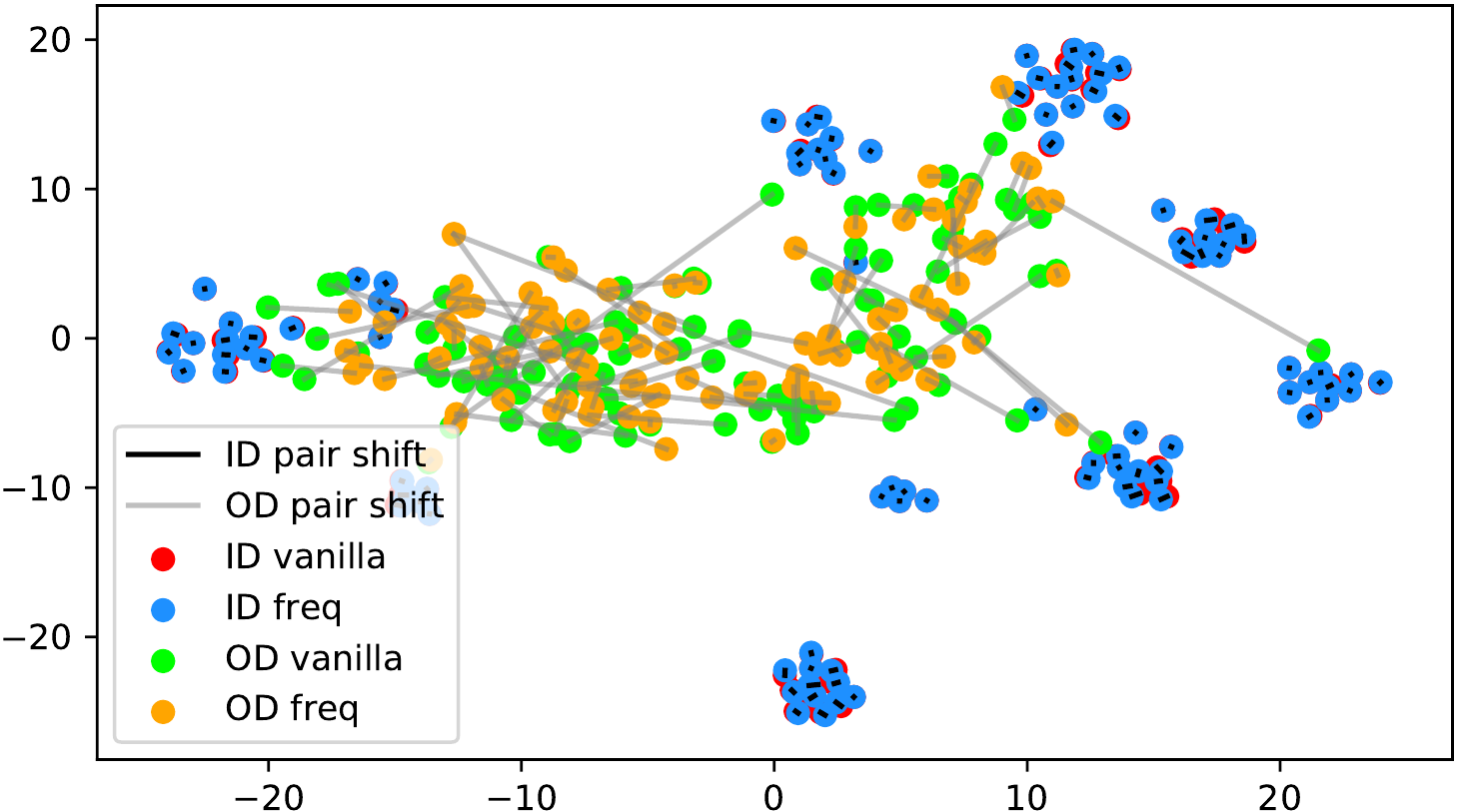}
			%\caption{fig1}
		\end{minipage}%
	}
	\\
	\vspace{-5px}
	\subfigure[Flip transformantion on DenseNet-121 for ImageNet Living 9 setting]{
		\begin{minipage}[t]{0.32\linewidth}
			\centering
			\includegraphics[width=1.8in]{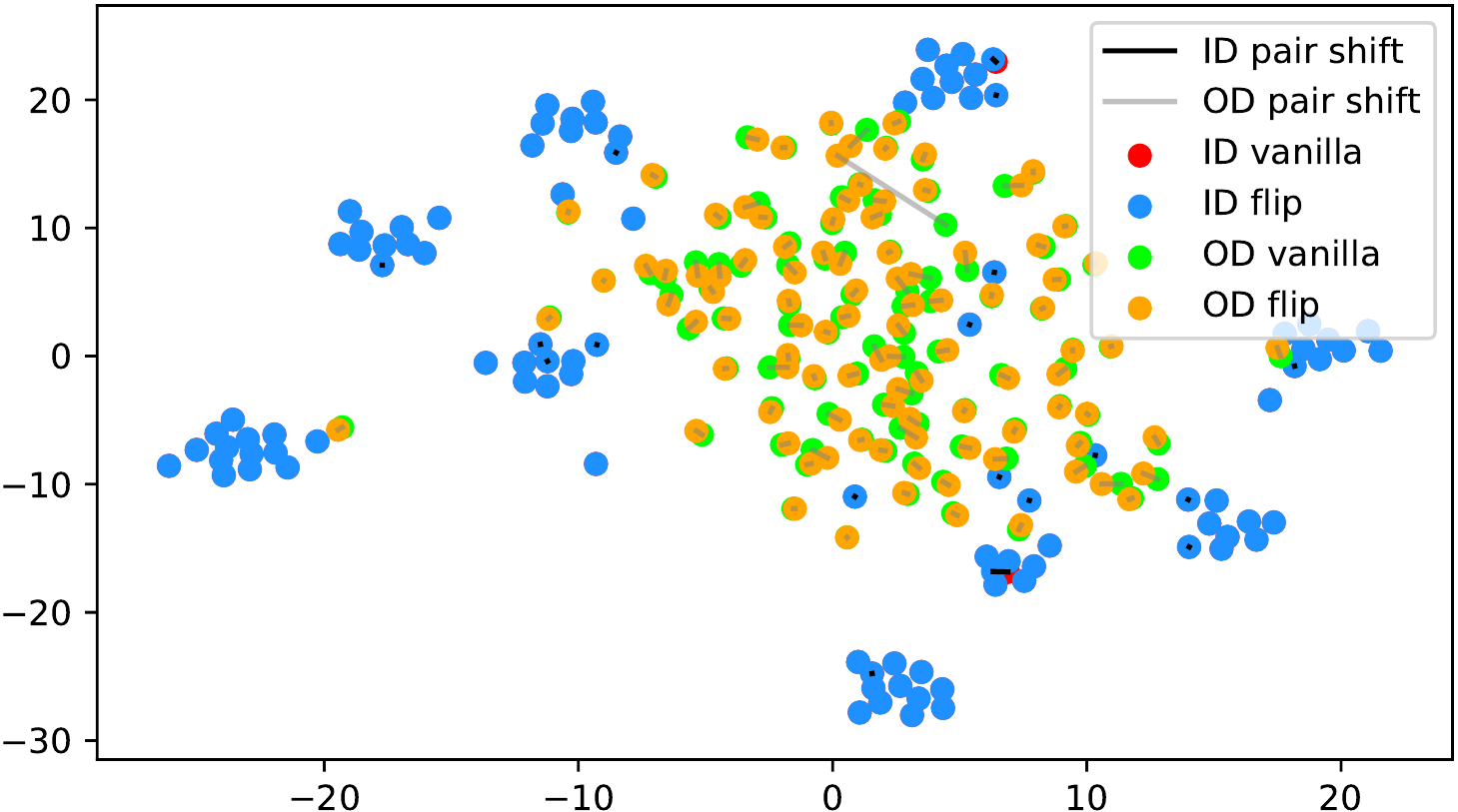}
			%\caption{fig1}
		\end{minipage}%
	}
	\subfigure[Flip transformantion on DenseNet-121 for ImageNet Geirhos 16 setting]{
		\begin{minipage}[t]{0.32\linewidth}
			\centering
			\includegraphics[width=1.8in]{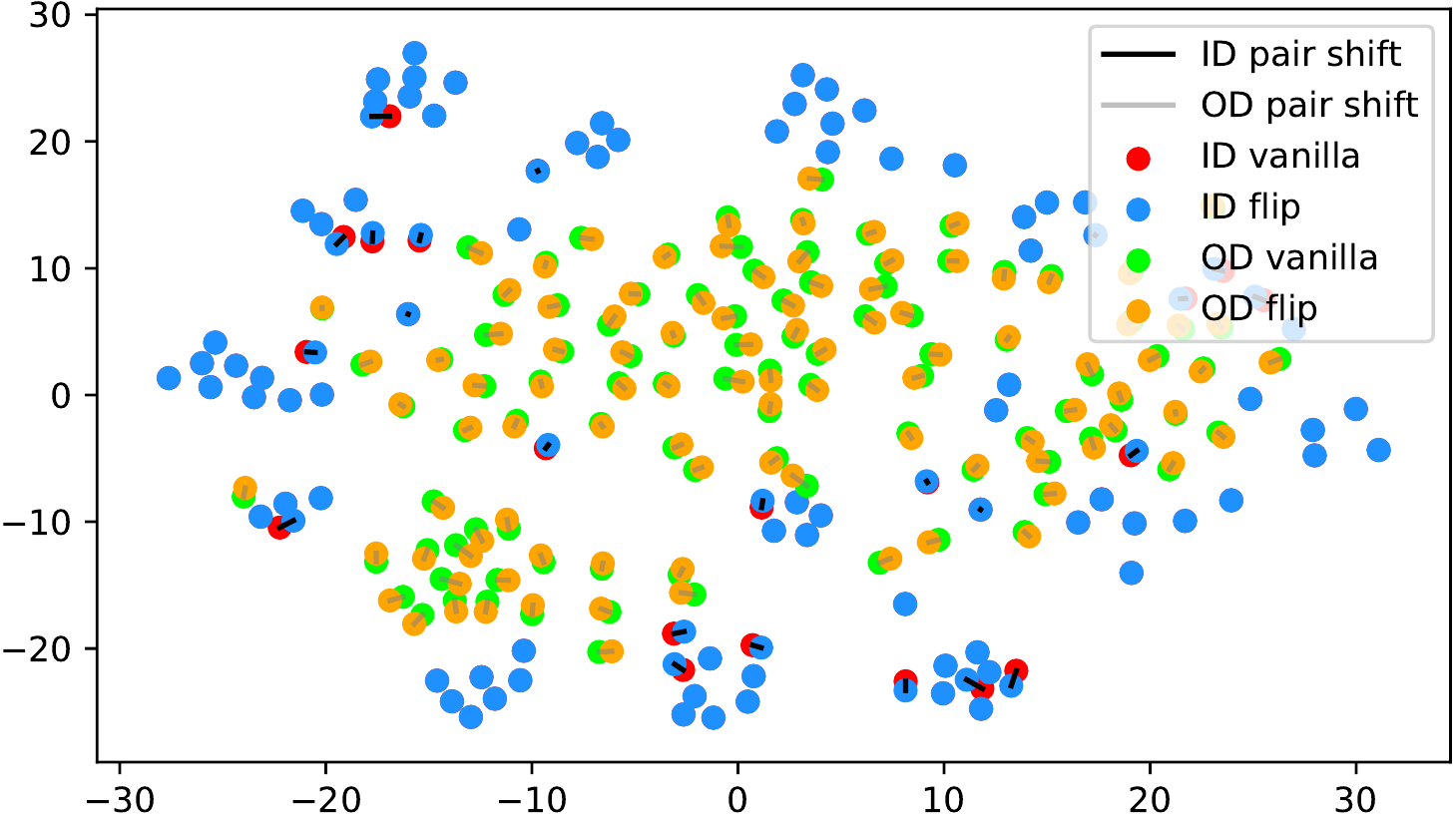}
			%\caption{fig1}
		\end{minipage}%
	}
	\subfigure[Flip transformantion on DenseNet-121 for ImageNet Mixed 10 setting]{
		\begin{minipage}[t]{0.32\linewidth}
			\centering
			\includegraphics[width=1.8in]{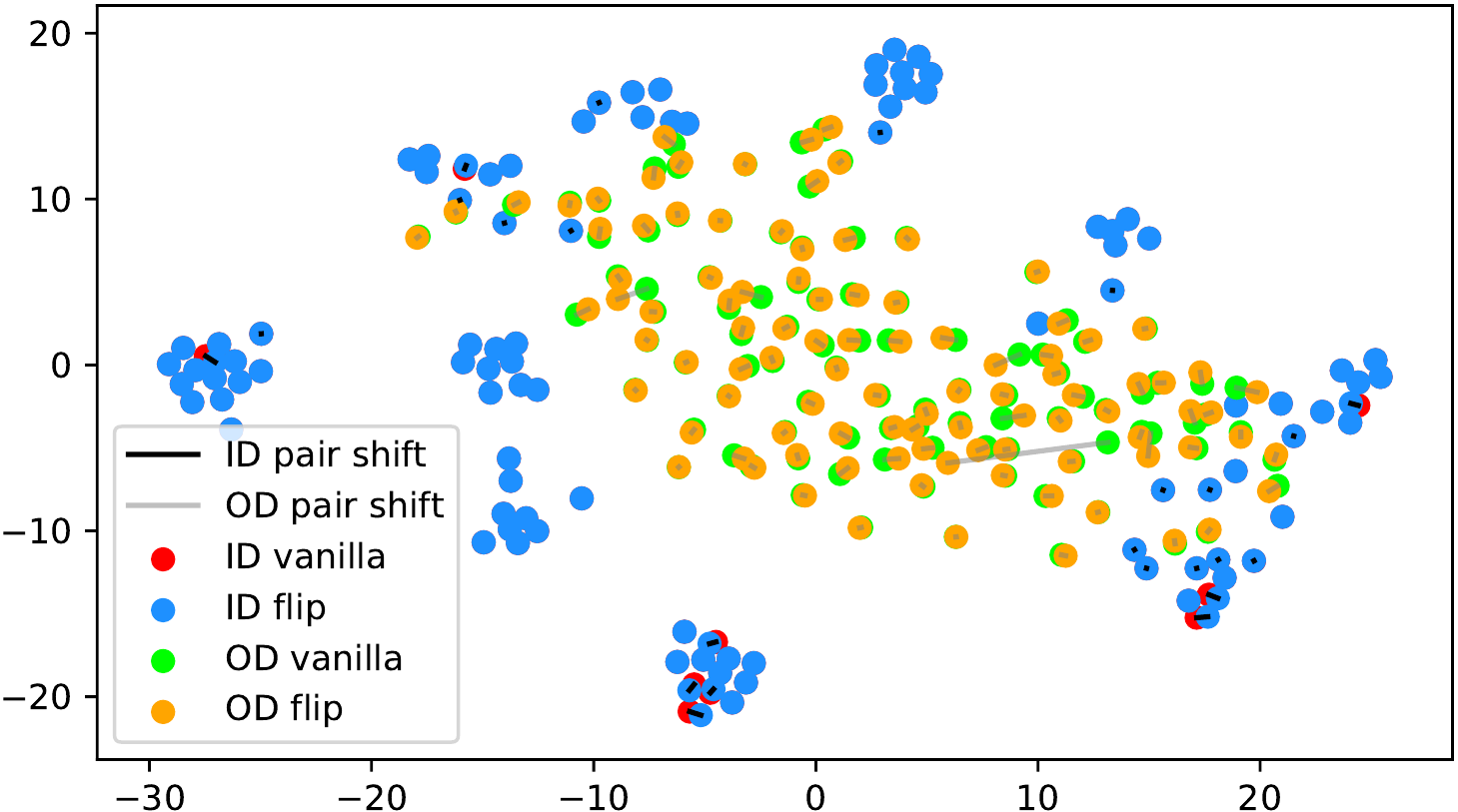}
			%\caption{fig1}
		\end{minipage}%
	}
	%	\subfigure[HARR WT]{ 
	%		\begin{minipage}[t]{0.2\linewidth}
	%			\centering
	%			\includegraphics[width=1in]{figure/img_haar.png}
	%			%\caption{fig1}
	%		
	%	}
	\caption{\label{fig:appendix_feature_space}More feature space projection examples across networks, datasets and augmentation methods.}
\end{figure*}

We verify the observation mentioned in Section~\ref{sec:intro} across architectures, dataset settings, and augmentation methods in Figure~\ref{fig:appendix_feature_space}. Due to the resolution problem, some images may be needed to zoom in to reflect the fine details we described. For each figure, we randomly pick 100 in-distribution samples and 100 out-distribution samples. For each pair of samples, a longer line segment means the distance in the projected feature space is large, leading to a smaller inner product value. In general, all the out-distribution samples in each subfigure have a statistically larger feature difference, compared to in-distribution samples. The dataset names, architectures, and transformation methods are annotated in each title of the subfigures. 

\section{Dataset construction and examples}
\label{app:datasetconstruction}
\begin{figure}[htbp]
	\centering
	\includegraphics[width=3in]{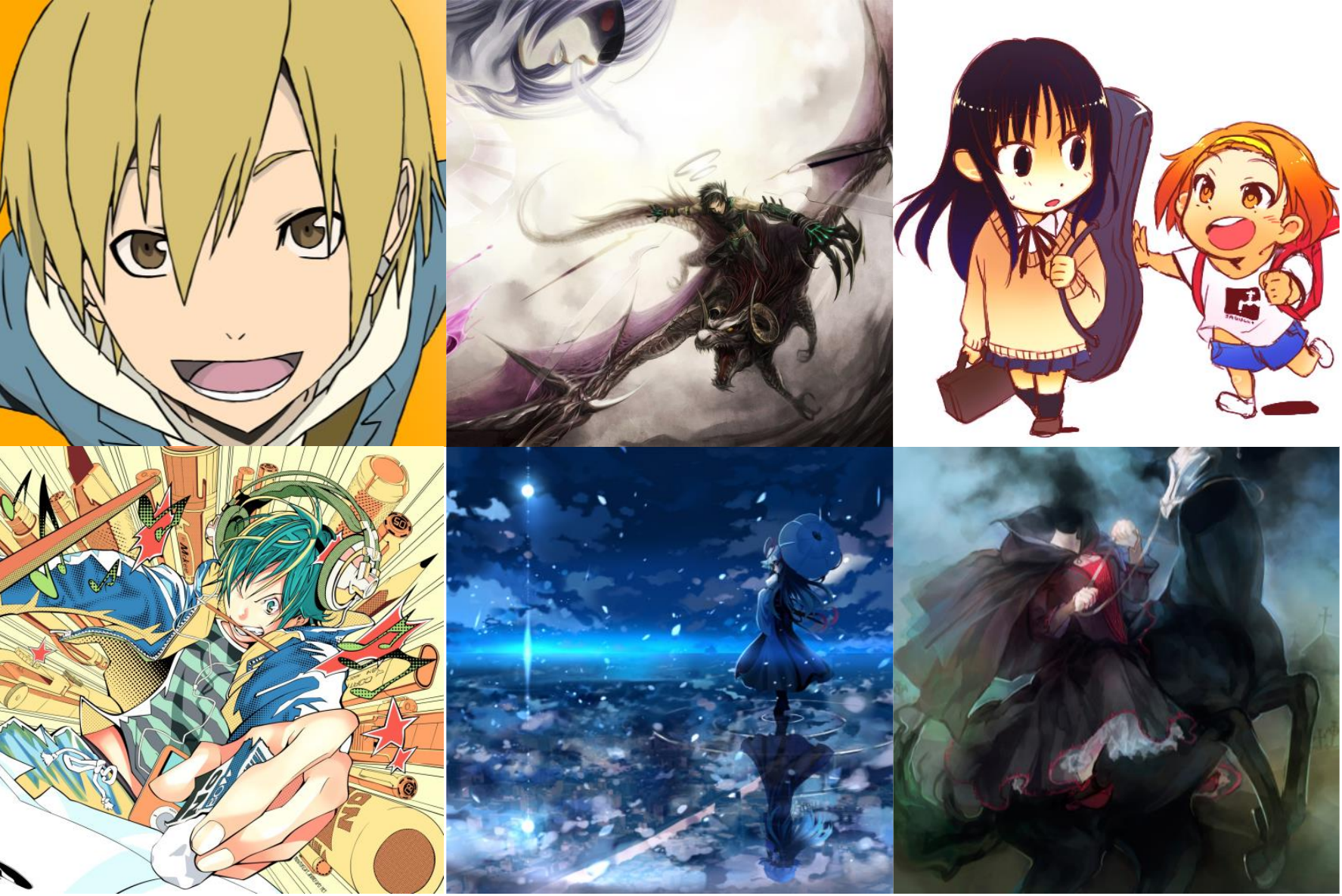}
	\caption{\label{fig:dataset}Some examples from the Anime dataset.}
\end{figure}

We show the detailed WordNet IDs and corresponding classes of each ImageNet subset in Table~\ref{table:datasetdetail}. And some examples of the Artificial dataset are shown in Figure~\ref{fig:dataset}.

\begin{table}[h]
	\caption{\label{table:datasetdetail}ImageNet subsets classes and the corresponding WordNet IDs.}
	\centering
	\begin{tabular}{|c|l|l|l|l|}
		\hline
		\multicolumn{1}{|l|}{Subsets}       & \multicolumn{2}{c|}{In-distribution classes \& WordNet IDs} & \multicolumn{2}{c|}{Out-distribution   classes \& WordNet IDs} \\ \hline
		\multirow{9}{*}{Living 9}    & dog                          & n02084071                    & furniture                          & n03405725                 \\ \cline{2-5} 
		& bird                         & n01503061                    & oven                               & n03862676                 \\ \cline{2-5} 
		& arthropod                    & n01767661                    & aircraft                           & n02686568                 \\ \cline{2-5} 
		& reptile                      & n01661091                    & bicycle                            & n02834778                 \\ \cline{2-5} 
		& primate                      & n02469914                    & musical instrument                 & n03800933                 \\ \cline{2-5} 
		& fish                         & n02512053                    &                                    &                           \\ \cline{2-5} 
		& feline                       & n02120997                    &                                    &                           \\ \cline{2-5} 
		& bovid                        & n02401031                    &                                    &                           \\ \cline{2-5} 
		& amphibian                    & n01627424                    &                                    &                           \\ \hline
		\multirow{16}{*}{Geirhos 16} & aircraft                     & n02686568                    & insect                             & n02159955                 \\ \cline{2-5} 
		& bear                         & n02131653                    & salamander                               & n01629276                  \\ \cline{2-5} 
		& bicycle                      & n02834778                    & clothing                           & n03623556                 \\ \cline{2-5} 
		& bird                         & n01503061                    & dophin                             & n02068974                 \\ \cline{2-5} 
		& boat                         & n02858304                    & reptile                            & n01661091                 \\ \cline{2-5} 
		& bottle                       & n02876657                    &                          &                  \\ \cline{2-5} 
		& car                          & n02958343                    &                                    &                           \\ \cline{2-5} 
		& cat                          & n02121808                    &                                    &                           \\ \cline{2-5} 
		& char                         & n03001627                    &                                    &                           \\ \cline{2-5} 
		& clock                        & n03046257                    &                                    &                           \\ \cline{2-5} 
		& dog                          & n02084071                    &                                    &                           \\ \cline{2-5} 
		& elephant                     & n02503517                    &                                    &                           \\ \cline{2-5} 
		& keyboard                     & n03614532                    &                                    &                           \\ \cline{2-5} 
		& knife                        & n03623556                    &                                    &                           \\ \cline{2-5} 
		& oven                         & n03862676                    &                                    &                           \\ \cline{2-5} 
		& truck                        & n04490091                    &                                    &                           \\ \hline
		\multirow{10}{*}{Mixed 10}   & dog                          & n02084071                    & furniture                          & n03405725                 \\ \cline{2-5} 
		& bird                         & n01503061                    & fish                               & n02512053                 \\ \cline{2-5} 
		& insect                       & n02159955                    & knife                              & n03623556                 \\ \cline{2-5} 
		& monkey                       & n02484322                    & keyboard                           & n03614532                 \\ \cline{2-5} 
		& car                          & n02958343                    &                            &                  \\ \cline{2-5} 
		& feline                       & n02120997                    &                                    &                           \\ \cline{2-5} 
		& truck                        & n04490091                    &                                    &                           \\ \cline{2-5} 
		& fruit                        & n13134947                    &                                    &                           \\ \cline{2-5} 
		& fungus                       & n12992868                    &                                    &                           \\ \cline{2-5} 
		& boat                         & n02858304                    &                                    &                           \\ \hline
	\end{tabular}
\end{table}

\section{Remaining score effect}
\label{app:remainingscore}

\begin{figure}[h]
	\centering
	\subfigure[\label{fig:den_img_max_probability}]{
		\begin{minipage}[t]{0.33\linewidth}
			\centering
			\includegraphics[width=\linewidth]{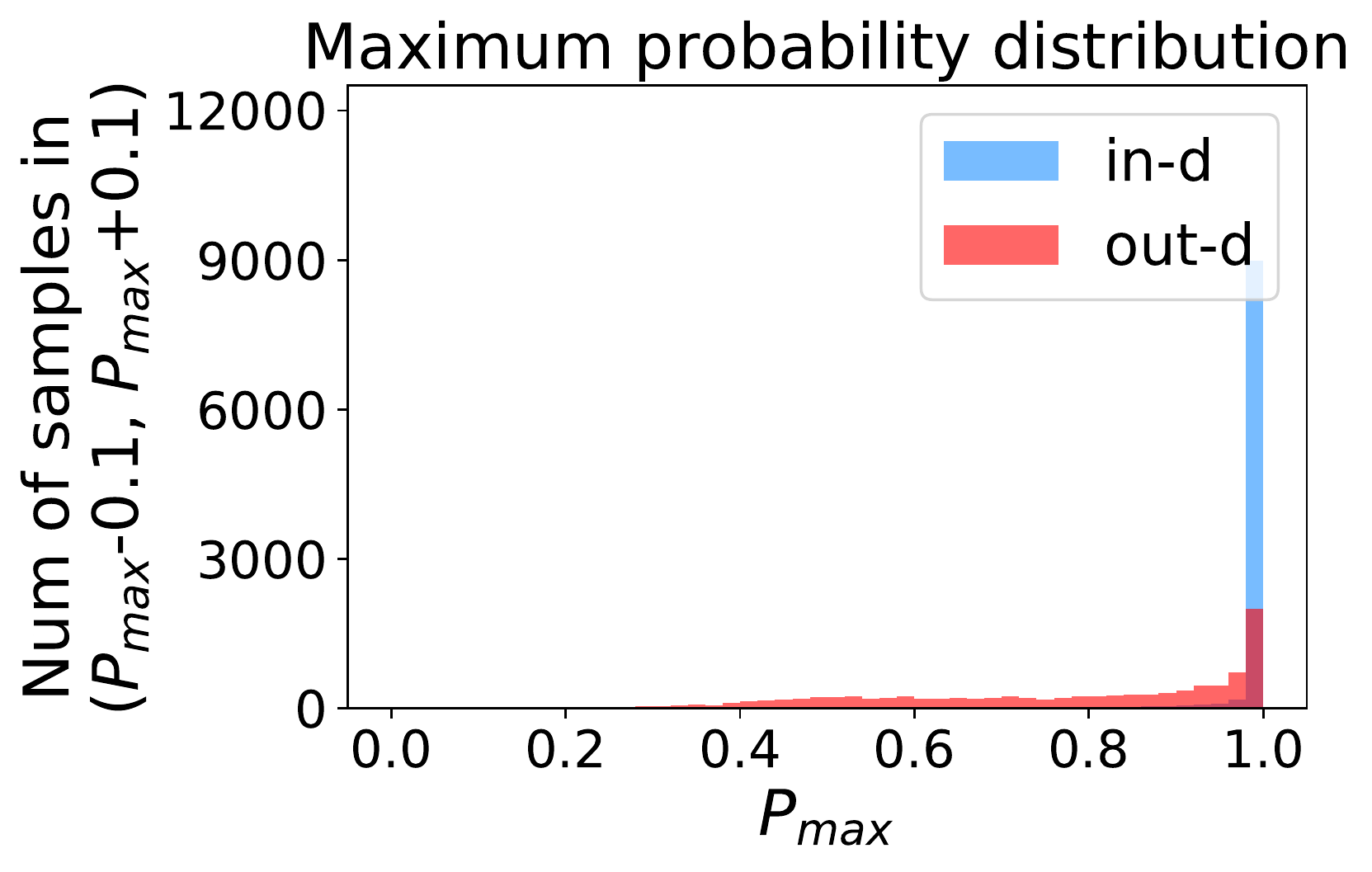}
			%\caption{fig1}
		\end{minipage}%
	}
	\subfigure[\label{fig:den_img_slice}]{
		\begin{minipage}[t]{0.29\linewidth}
			\centering
			\includegraphics[width=\linewidth]{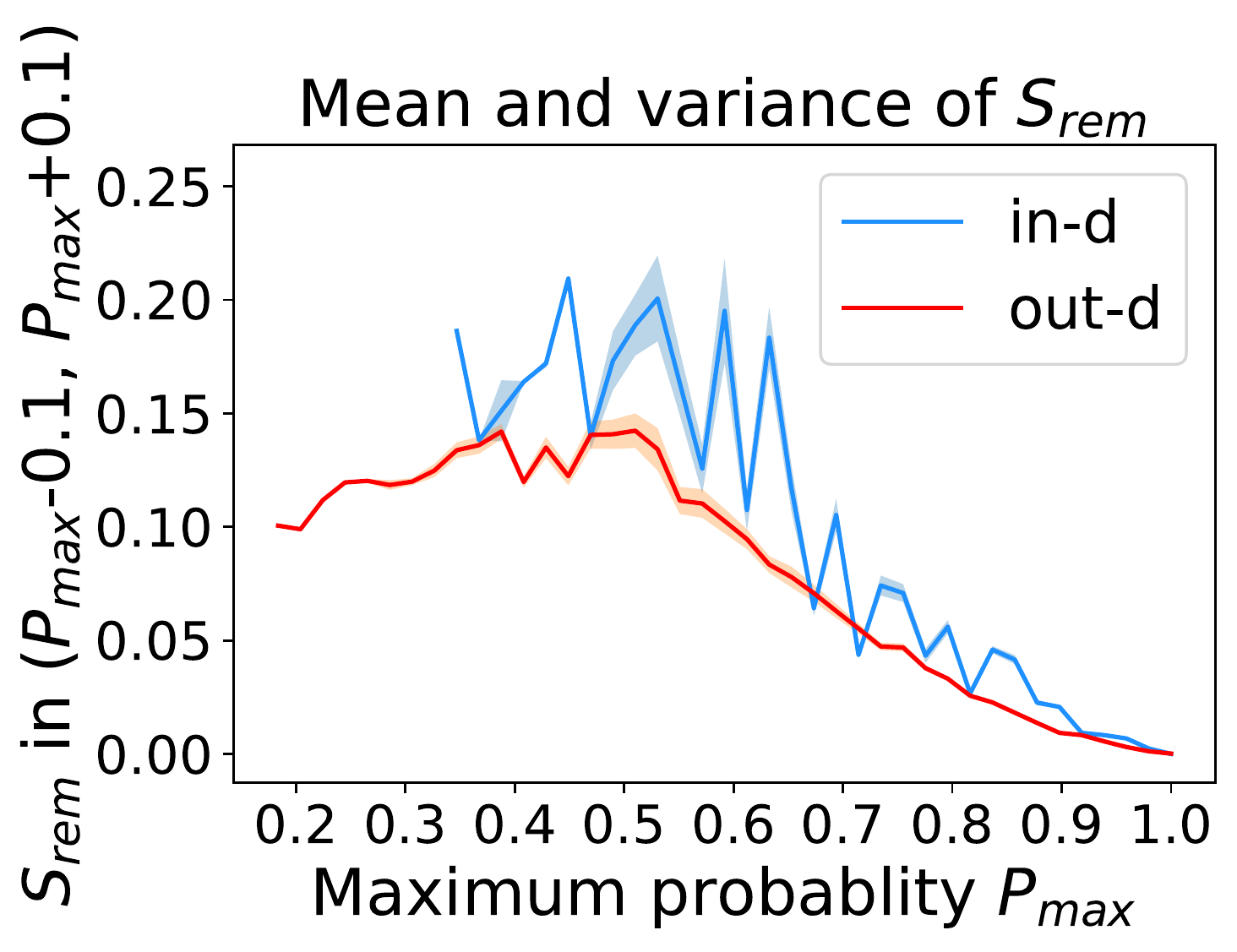}
			%\caption{fig1}
		\end{minipage}%
	}
	\subfigure[\label{fig:den_img_inner_product}]{
		\begin{minipage}[t]{0.33\linewidth}
			\centering
			\includegraphics[width=\linewidth]{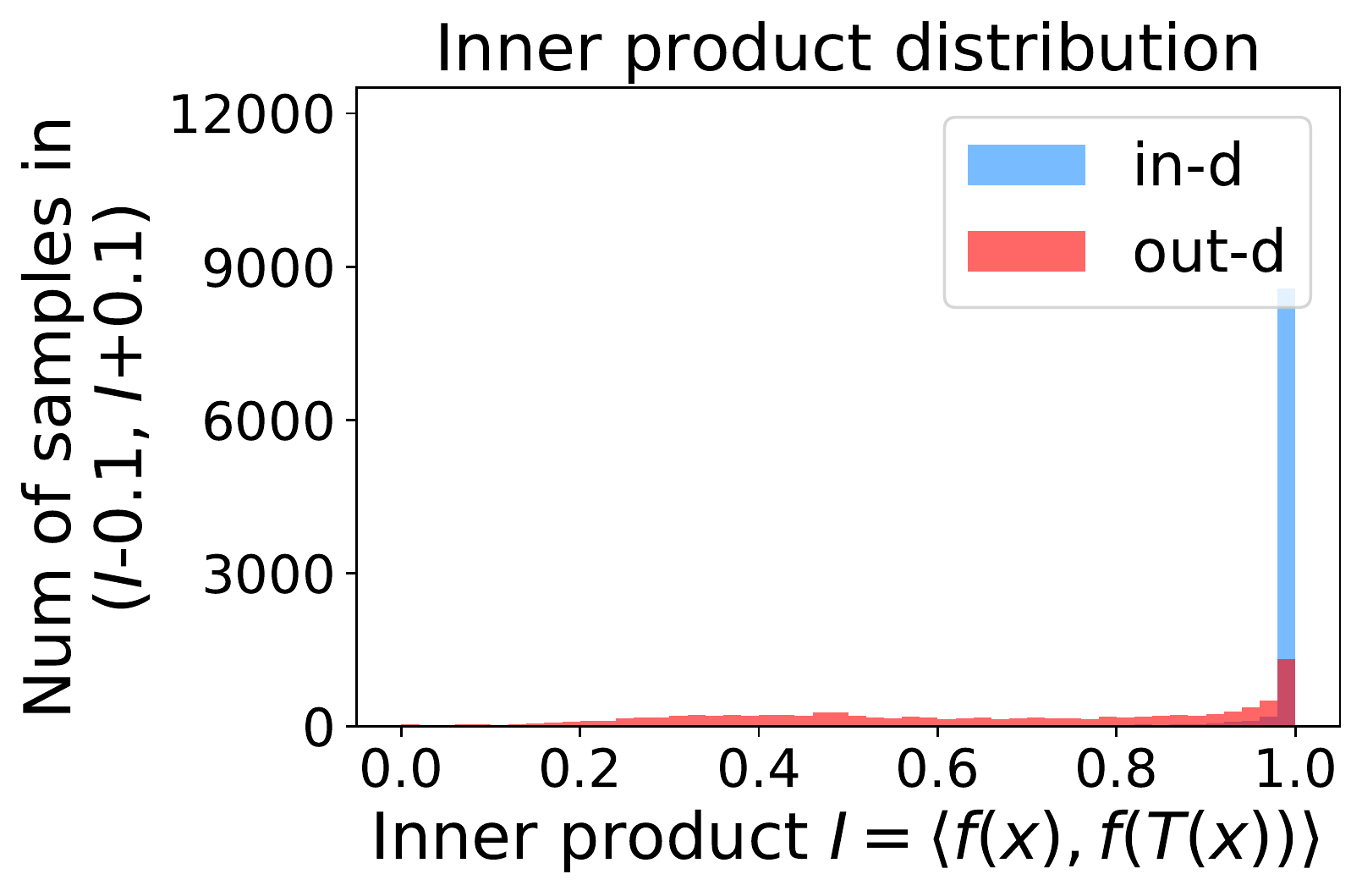}
			%\caption{fig1}
		\end{minipage}%
	}
	\\
	\vspace{-5px}
	\subfigure[\label{fig:res_svhn_max_probability}]{
		\begin{minipage}[t]{0.33\linewidth}
			\centering
			\includegraphics[width=\linewidth]{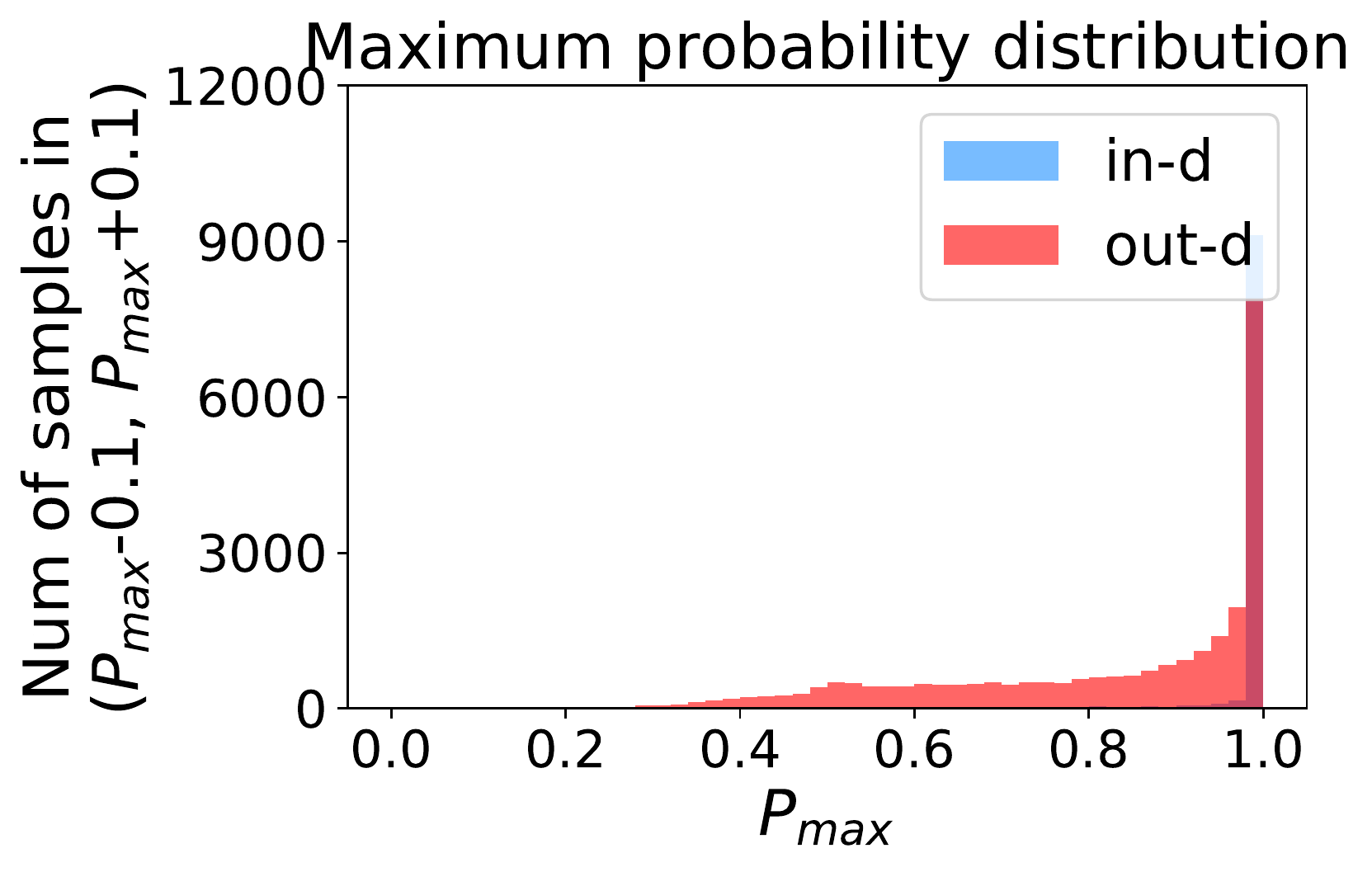}
			%\caption{fig1}
		\end{minipage}%
	}
	\subfigure[\label{fig:res_svhn_slice}]{
		\begin{minipage}[t]{0.29\linewidth}
			\centering
			\includegraphics[width=\linewidth]{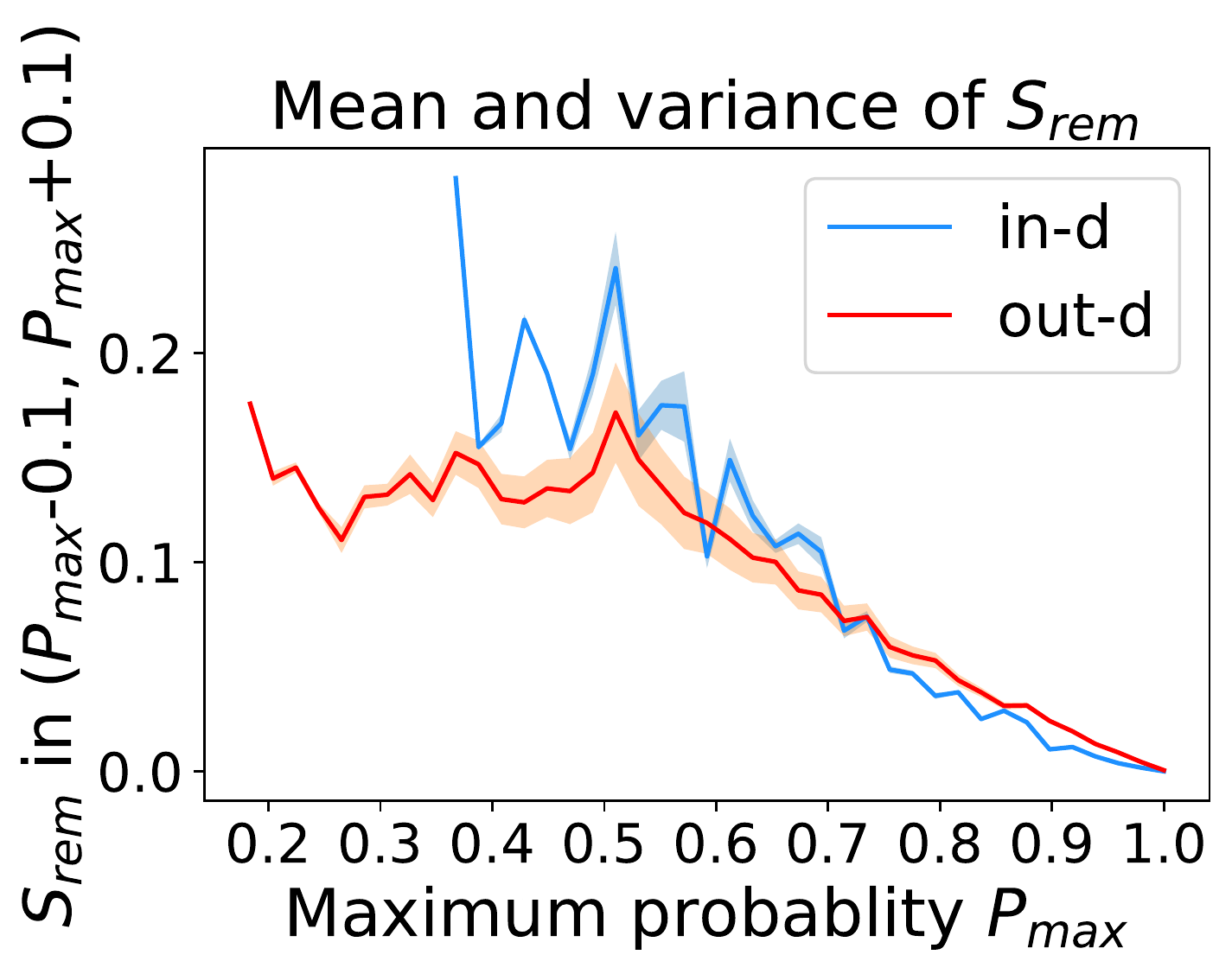}
			%\caption{fig1}
		\end{minipage}%
	}
	\subfigure[\label{fig:res_svhn_inner_product}]{
		\begin{minipage}[t]{0.33\linewidth}
			\centering
			\includegraphics[width=\linewidth]{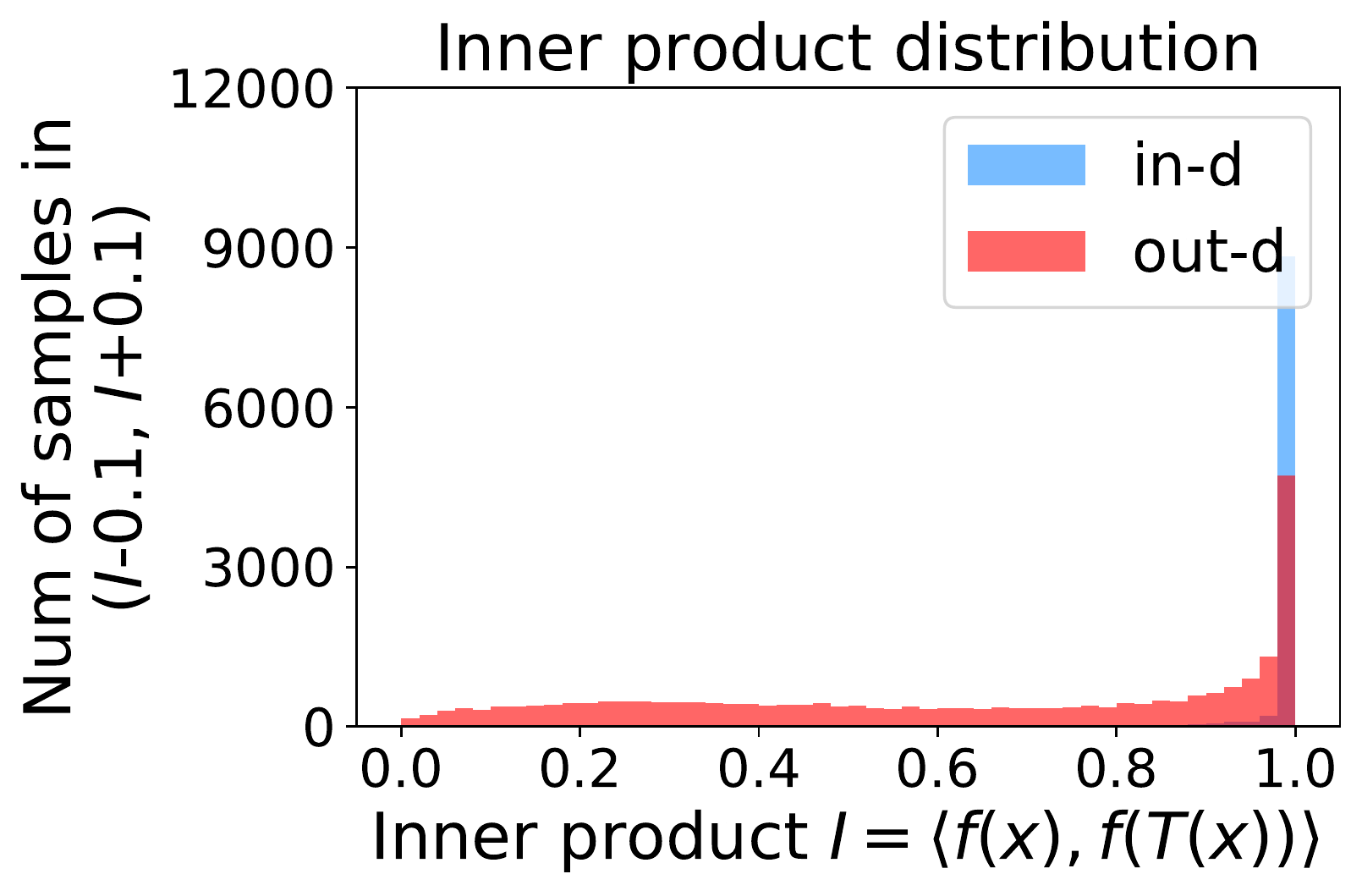}
			%\caption{fig1}
		\end{minipage}%
	}
	\\
	\vspace{-5px}
	\subfigure[\label{fig:den_lsun_max_probability}]{
		\begin{minipage}[t]{0.33\linewidth}
			\centering
			\includegraphics[width=\linewidth]{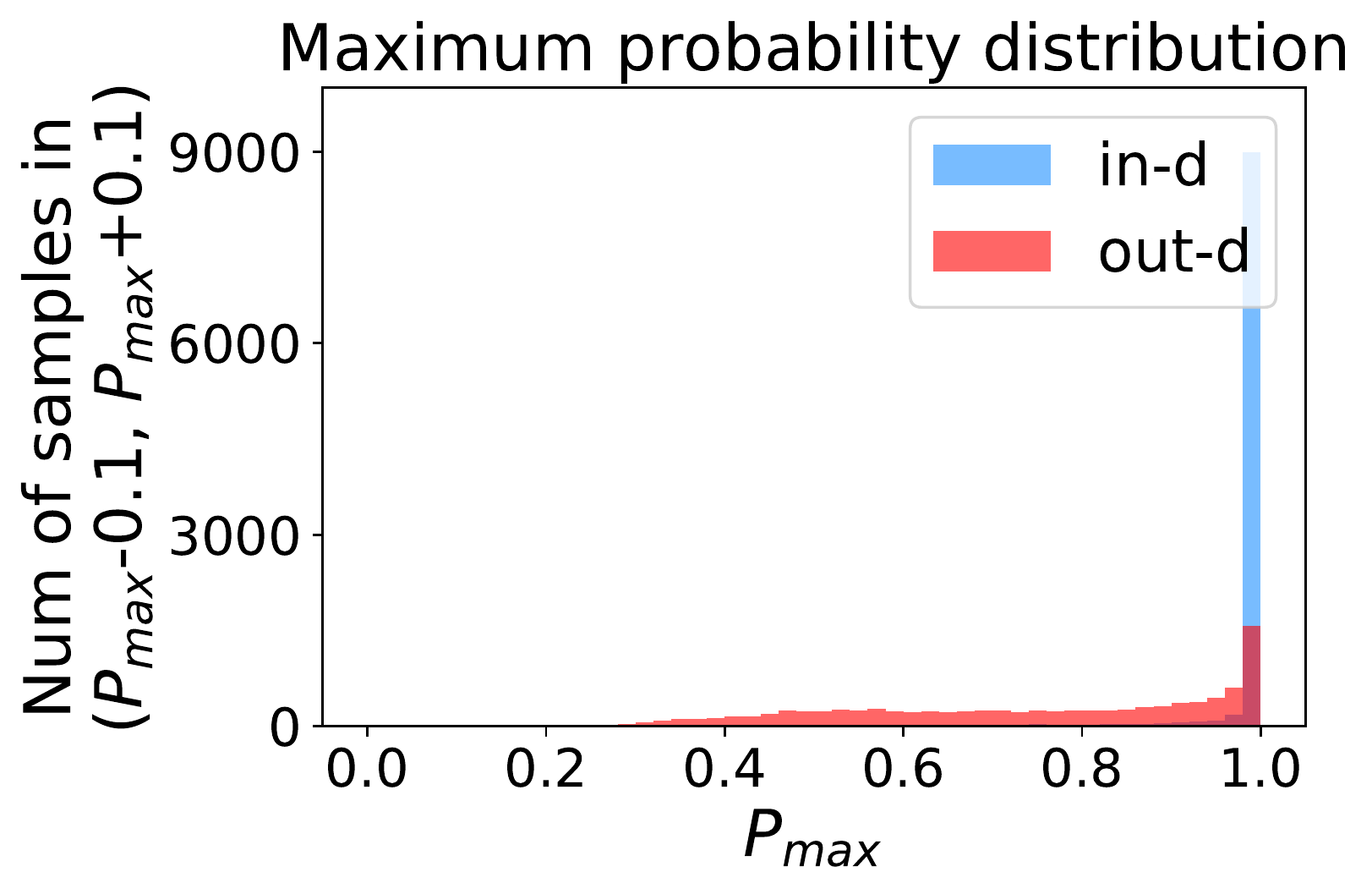}
			%\caption{fig1}
		\end{minipage}%
	}
	\subfigure[\label{fig:den_lsun_slice}]{
		\begin{minipage}[t]{0.29\linewidth}
			\centering
			\includegraphics[width=\linewidth]{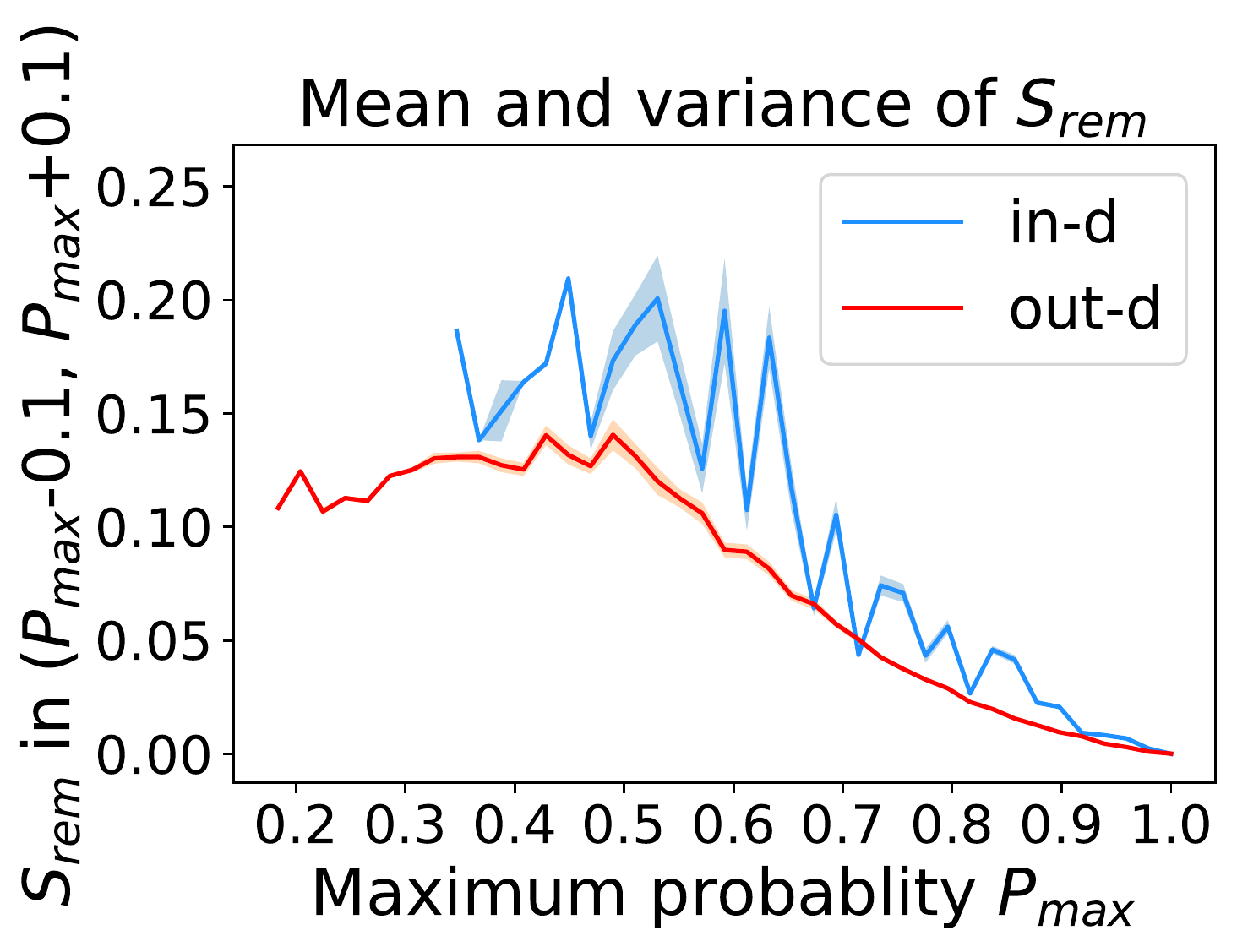}
			%\caption{fig1}
		\end{minipage}%
	}
	\subfigure[\label{fig:den_lsun_inner_product}]{
		\begin{minipage}[t]{0.33\linewidth}
			\centering
			\includegraphics[width=\linewidth]{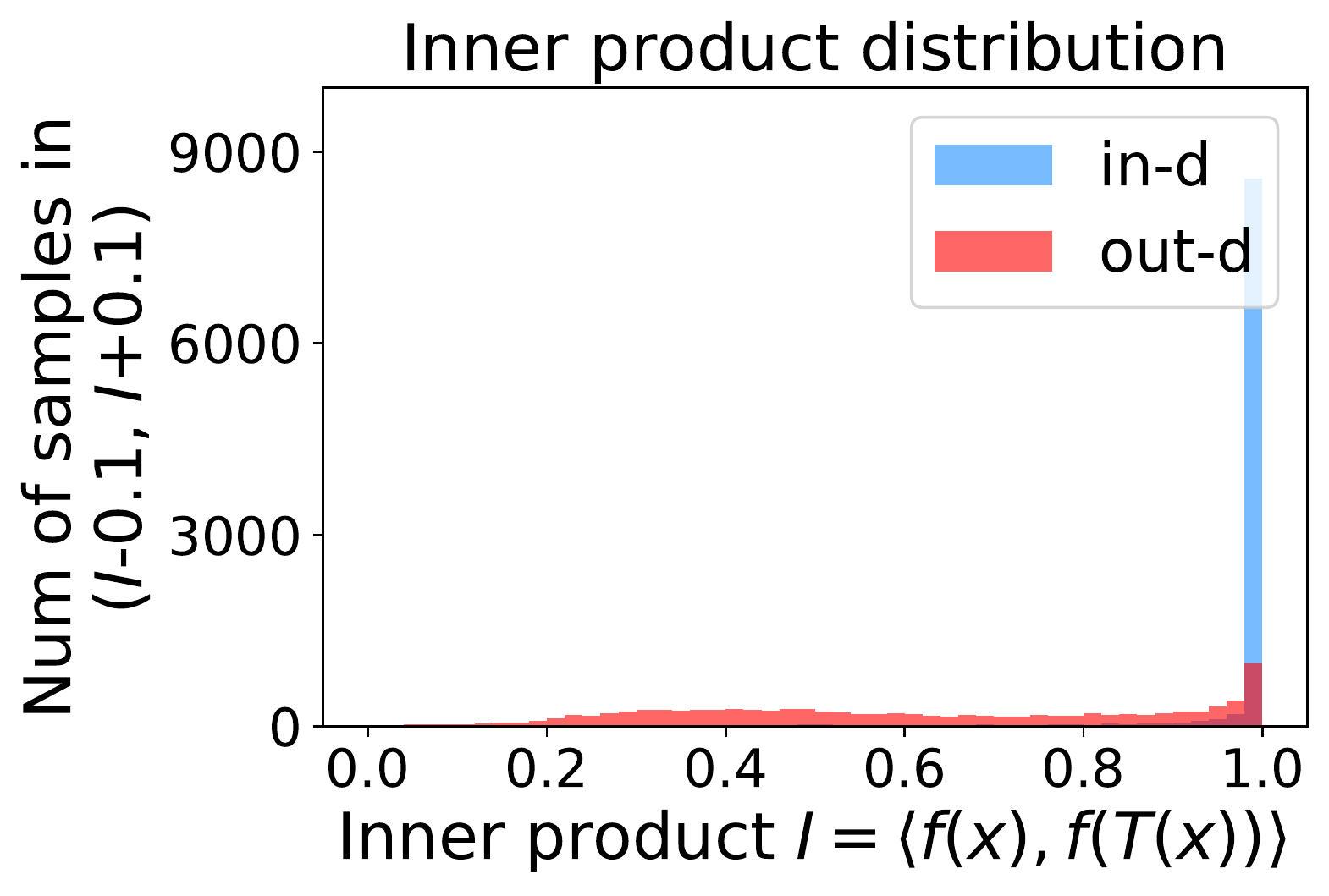}
			%\caption{fig1}
		\end{minipage}%
	}

	\centering
	
	\caption{\label{fig:cifar_empirical_evidence}Figures showing the effect of remaining classes of a pre-trained model. (a)-(c): CIFAR-10 vs. ImageNet\_resize on DenseNet-100. (d)-(f): CIFAR-10 vs. SVHN on ResNet-34. (g)-(i): CIFAR-10 vs. LSUN\_resize on DenseNet-100.}
\end{figure}

\begin{figure}[h]
	\centering
	\subfigure[\label{fig:res_liv_max_probability}]{
		\begin{minipage}[t]{0.33\linewidth}
			\centering
			\includegraphics[width=\linewidth]{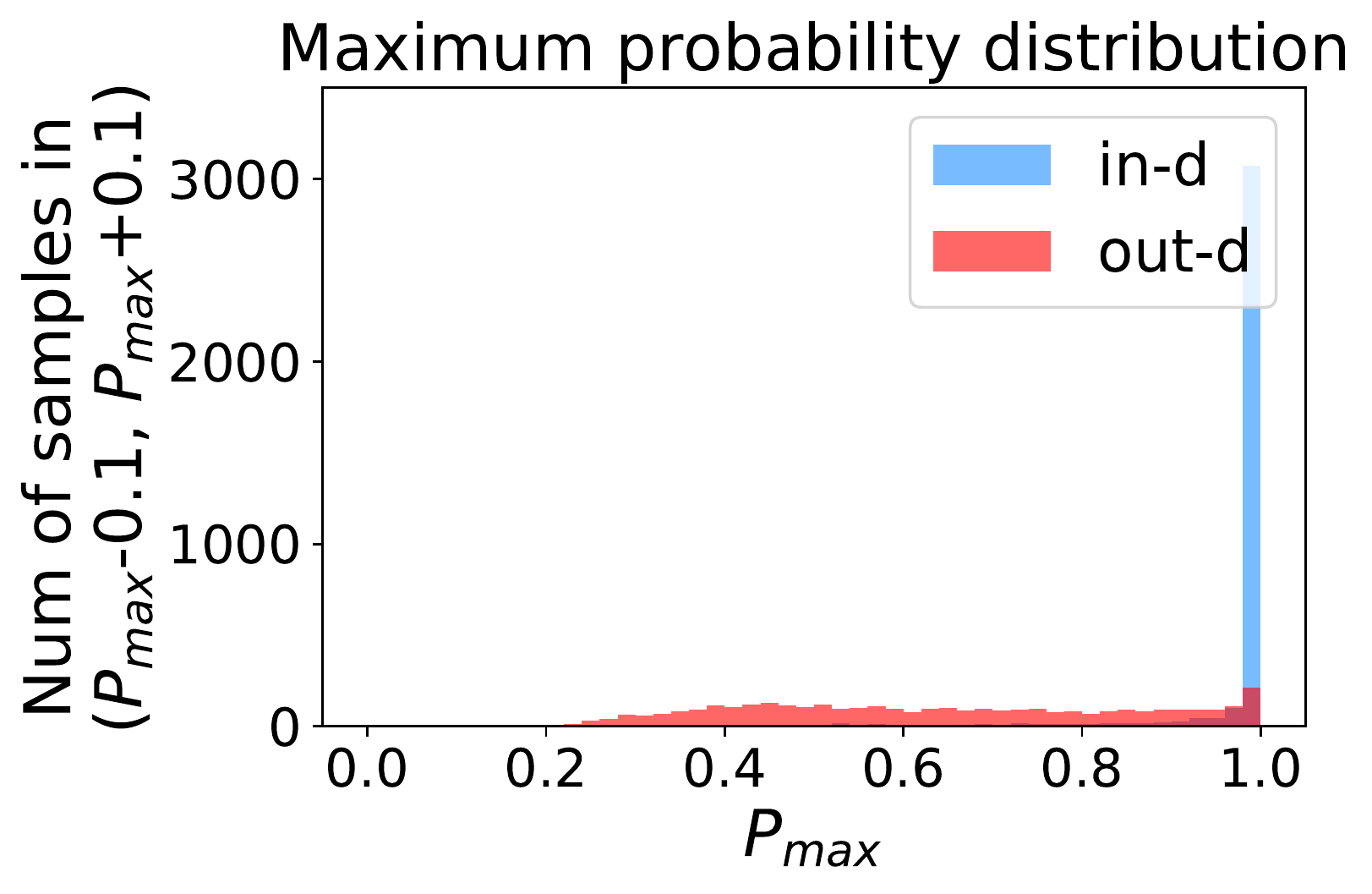}
			%\caption{fig1}
		\end{minipage}%
	}
	\subfigure[\label{fig:res_liv_slice}]{
		\begin{minipage}[t]{0.30\linewidth}
			\centering
			\includegraphics[width=\linewidth]{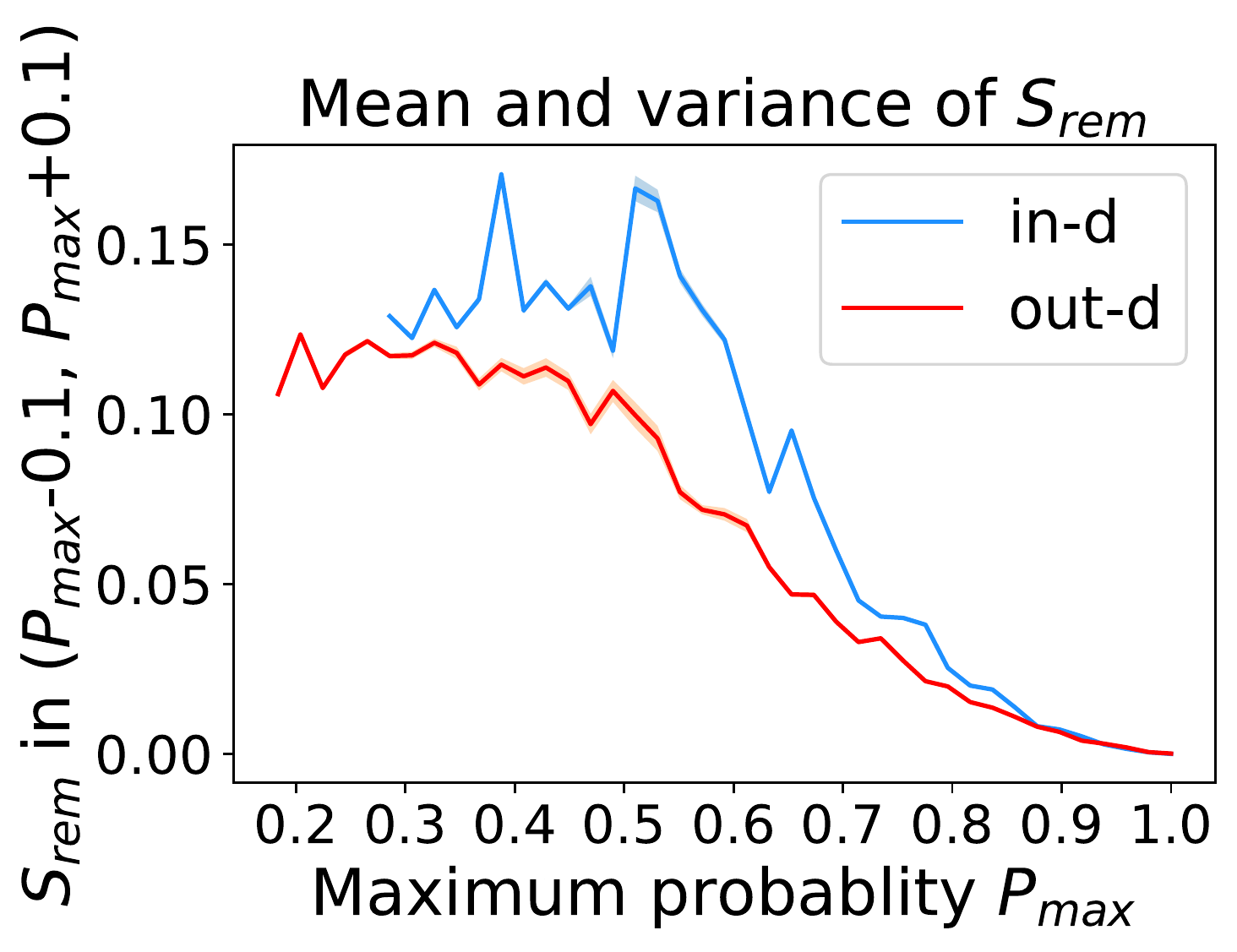}
			%\caption{fig1}
		\end{minipage}%
	}
	\subfigure[\label{fig:res_liv_inner_product}]{
		\begin{minipage}[t]{0.33\linewidth}
			\centering
			\includegraphics[width=\linewidth]{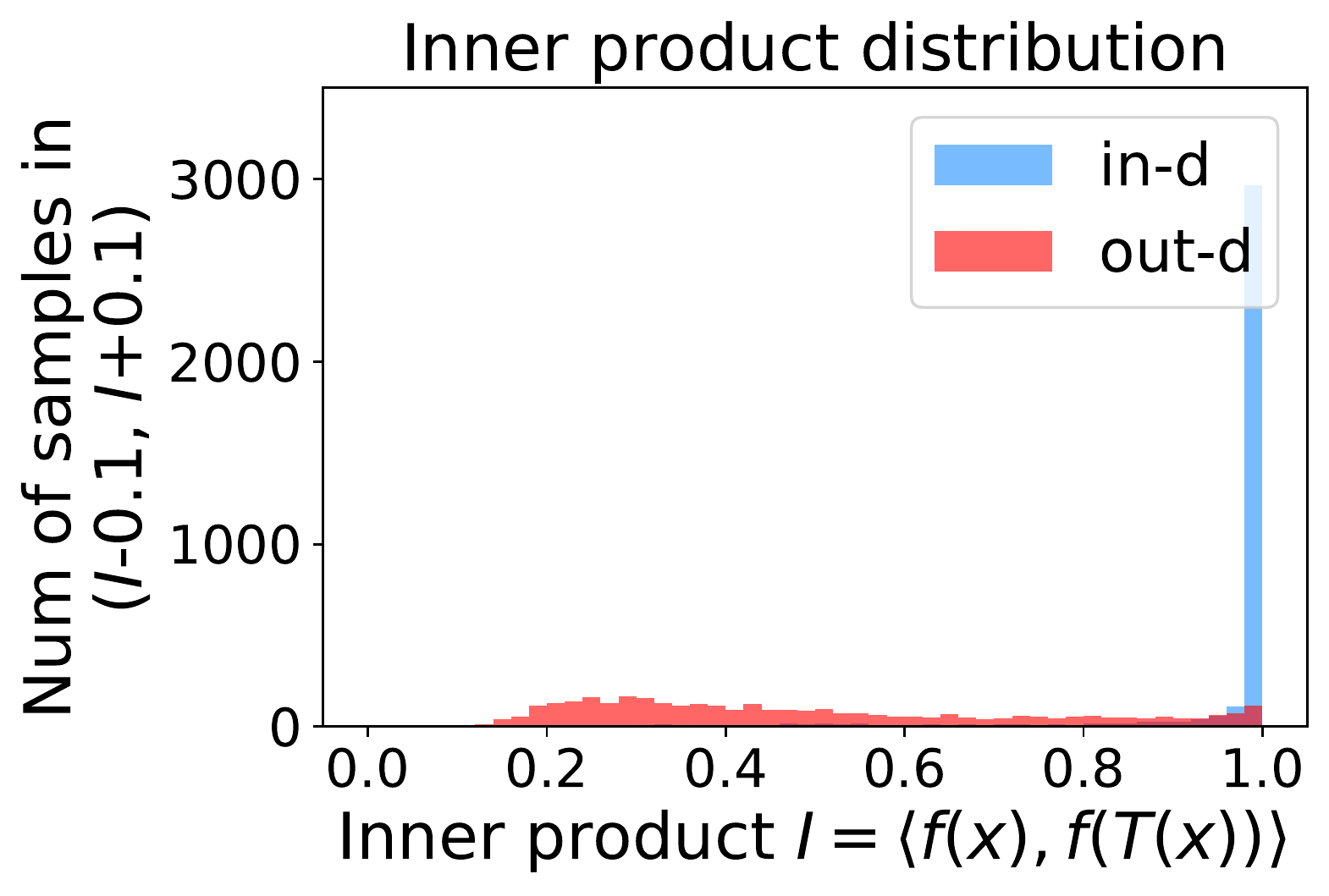}
			%\caption{fig1}
		\end{minipage}%
	}
	\\
	\vspace{-5px}
	\subfigure[\label{fig:den_ge_max_probability}]{
		\begin{minipage}[t]{0.33\linewidth}
			\centering
			\includegraphics[width=\linewidth]{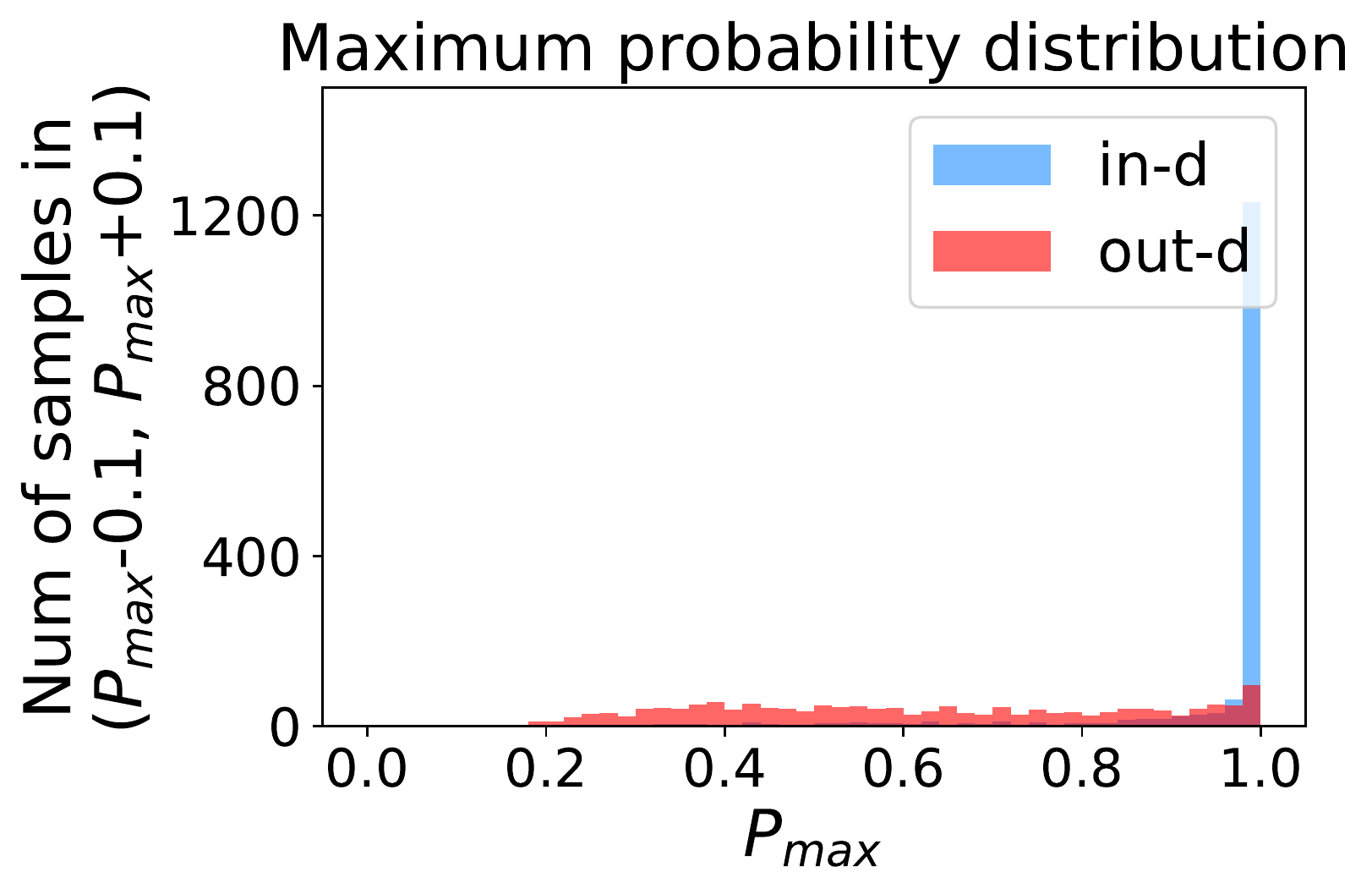}
			%\caption{fig1}
		\end{minipage}%
	}
	\subfigure[\label{fig:den_ge_slice}]{
		\begin{minipage}[t]{0.3\linewidth}
			\centering
			\includegraphics[width=\linewidth]{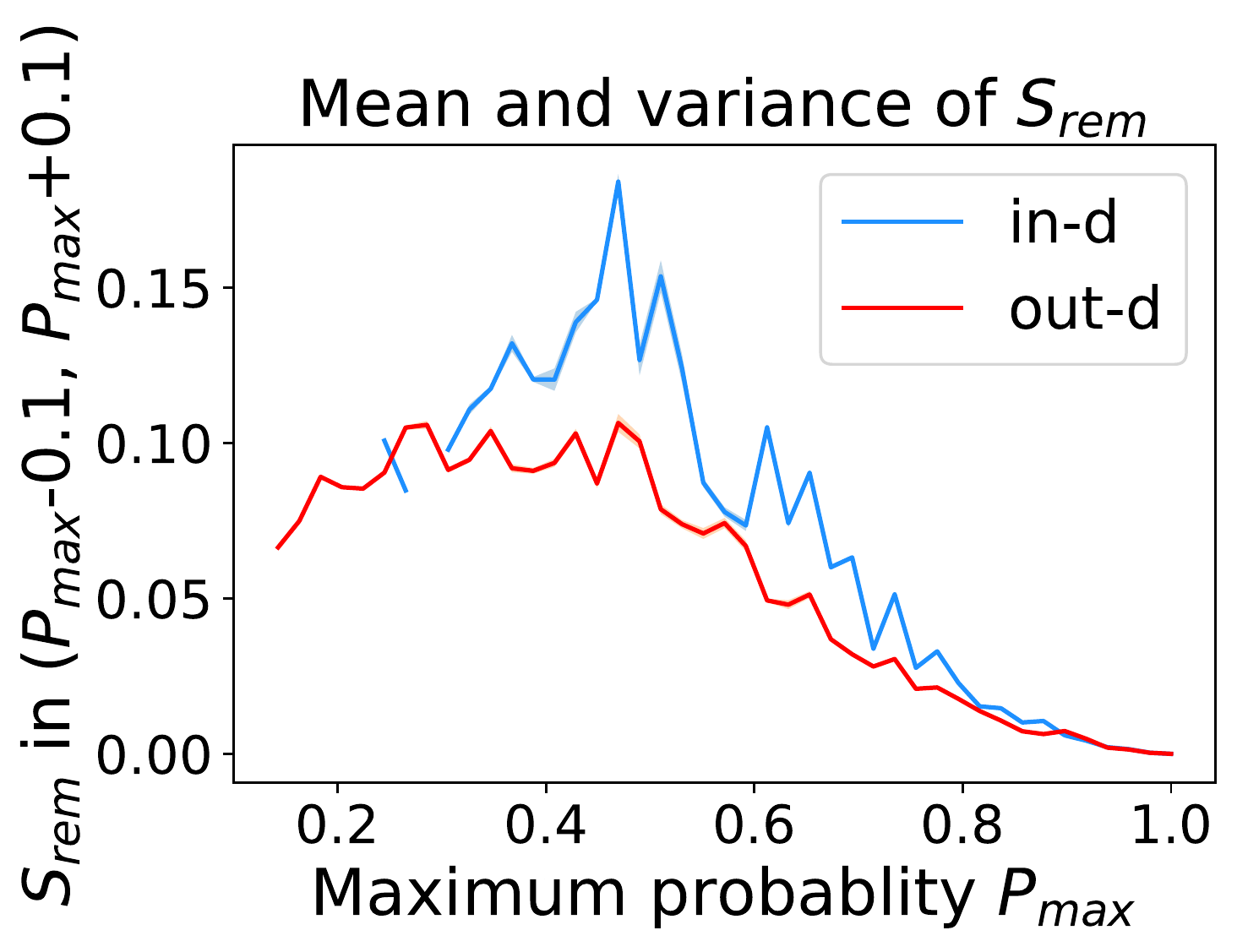}
			%\caption{fig1}
		\end{minipage}%
	}
	\subfigure[\label{fig:den_ge_inner_product}]{
		\begin{minipage}[t]{0.33\linewidth}
			\centering
			\includegraphics[width=\linewidth]{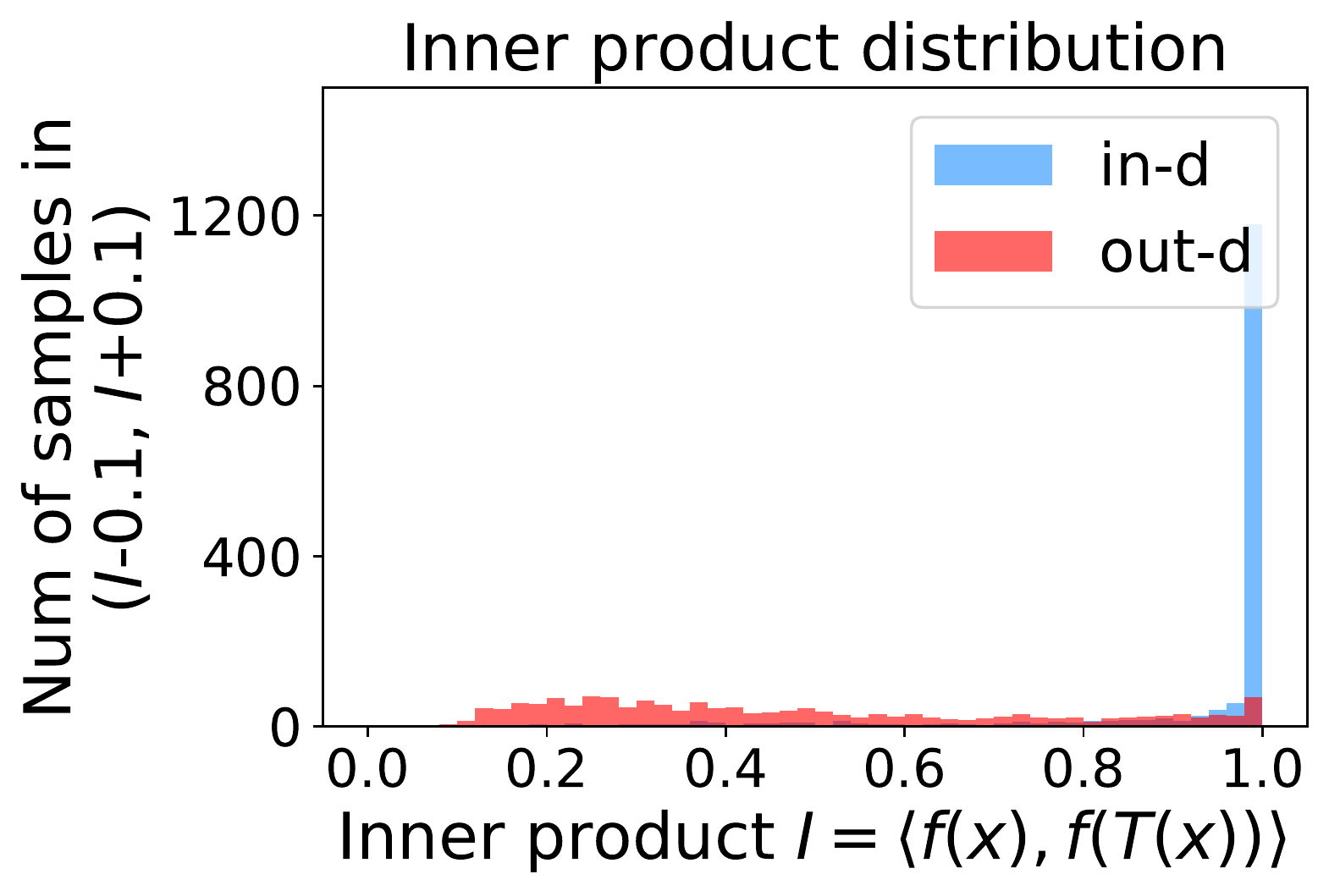}
			%\caption{fig1}
		\end{minipage}%
	}
	\\
	\vspace{-5px}
	\subfigure[\label{fig:res_mixed_max_probability}]{
		\begin{minipage}[t]{0.33\linewidth}
			\centering
			\includegraphics[width=\linewidth]{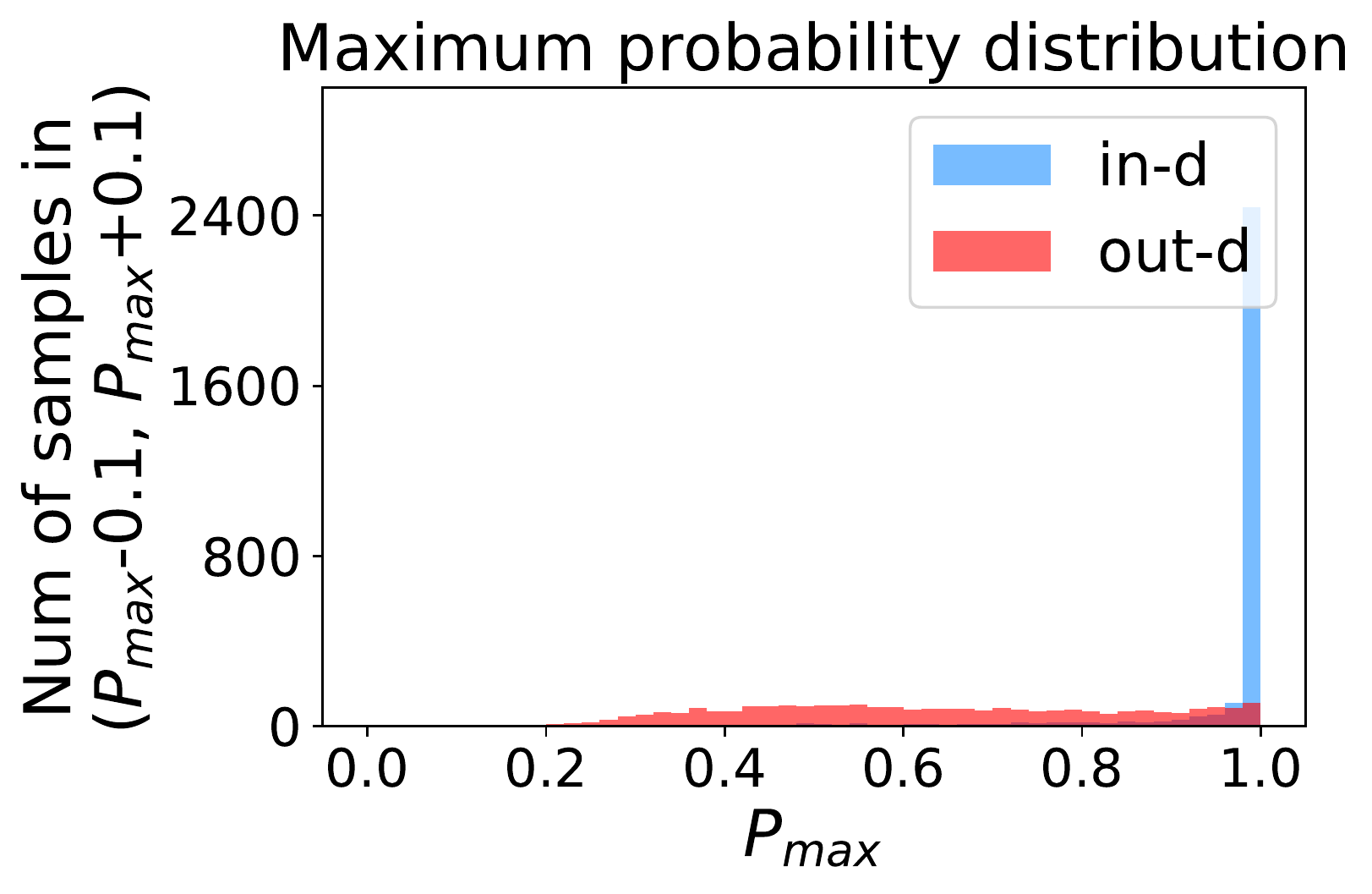}
			%\caption{fig1}
		\end{minipage}%
	}
	\subfigure[\label{fig:res_mixed_slice}]{
		\begin{minipage}[t]{0.3\linewidth}
			\centering
			\includegraphics[width=\linewidth]{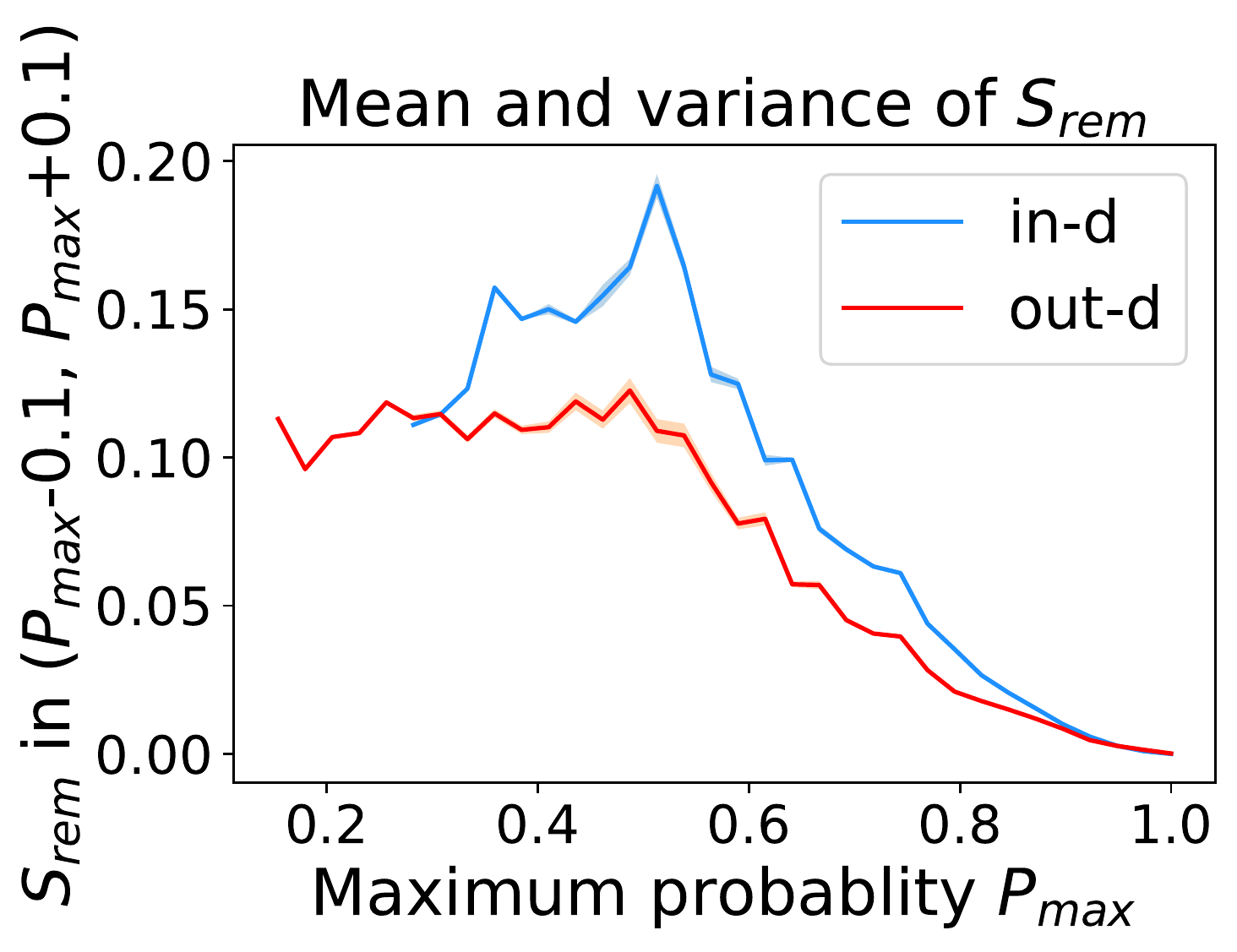}
			%\caption{fig1}
		\end{minipage}%
	}
	\subfigure[\label{fig:res_mixed_inner_product}]{
		\begin{minipage}[t]{0.33\linewidth}
			\centering
			\includegraphics[width=\linewidth]{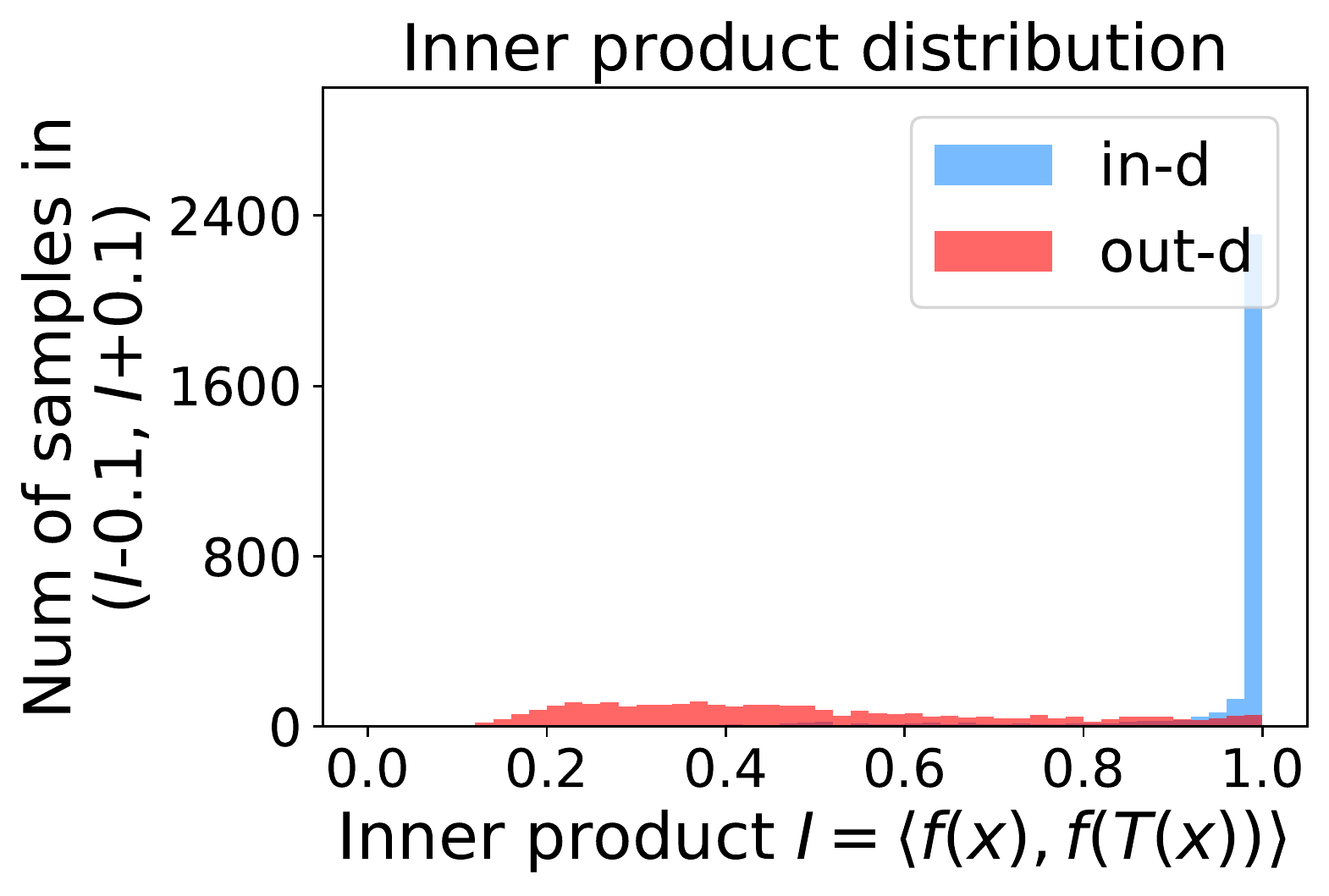}
			%\caption{fig1}
		\end{minipage}%
	}
	\caption{\label{fig:imagenet_subset_empirical_evidence}Figures showing the effect of remaining classes of a pre-trained model on ImageNet subsets. (a)-(c): The Living 9 subset setting on ResNet-50. (d)-(f): The Geirhos 16 subset setting on DenseNet-121. (g)-(i): The Mixed 10 subset setting on ResNet-50.}
\end{figure}

More examples showing the effect of the remaining classes across network architectures and datasets are shown in Figure~\ref{fig:cifar_empirical_evidence} and Figure~\ref{fig:imagenet_subset_empirical_evidence}. Similar to Figure~\ref{fig:empirical_evidence}, each row contains (a)~Maximum probability distributions of in- and out-distribution samples. (b)~Mean and variance of remaining scores within each slot $(P_{max}-\epsilon, P_{max}+\epsilon)$. (c)~Inner product~(ours) distributions of in- and out-distribution samples.
%We show some examples in Figure~\ref{fig:slice1}, \ref{fig:slice2} and \ref{fig:slice3}.

%\begin{figure}[h]
%	\centering
%	\begin{minipage}[t]{0.48\textwidth}
%		\centering            
%		\includegraphics[width=2.8in]{figure/slice/slice_img_ani_den.pdf}
%		\caption{\label{fig:slice1}The mean and variance of the remaining score within evenly divided intervals of DenseNet-121 on ImageNet vs Anime. }
%	\end{minipage}
%	\begin{minipage}[t]{0.48\textwidth}
%		\centering
%		\includegraphics[width=2.8in]{figure/slice/slice_cifar_img_den.pdf}
%		\caption{\label{fig:slice2}The mean and variance of the remaining score within evenly divided intervals of DenseNet-100 on CIFAR10 vs ImageNet\_resize. }
%	\end{minipage}
%\end{figure}

%\begin{figure}[h]
%	\centering            
%	\includegraphics[width=3.5in]{figure/slice/slice_img_ani_den.pdf}
%	\caption{\label{fig:slice1}The mean and variance of the remaining score within evenly divided intervals of DenseNet-121 on ImageNet vs Anime. }
%\end{figure}

%\begin{figure}[h]
%	\centering
%	\includegraphics[width=3.5in]{figure/slice/slice_cifar_img_den.pdf}
%	\caption{\label{fig:slice2}The mean and variance of the remaining score within evenly divided intervals of DenseNet-100 on CIFAR10 vs ImageNet\_resize. }
%\end{figure}

%\begin{figure}[h]
%	\centering
%	\includegraphics[width=3.5in]{figure/slice/slice_liv_res.pdf}
%	\caption{\label{fig:slice3}The mean and variance of the remaining score within evenly divided intervals of ResNet-50 on Living 9 setting. }
%\end{figure}

\section{Runs}
\label{app:runs}
%%% another note: the AUC/ROC comparison in test set might be not good enough? this is not the point

\subsection{Proof of Lemma~\ref{lem:expectedRuns}}

In this part, we validate the rationality of the expected runs.
Specifically, we first show that when the validation set is large enough (which usually holds in practice although we only have limited data at hand), the expected runs number can be calculated by the following Eqn~\eqref{Eqn:expectedRuns}.
We always assume that the random variables are bounded by $[-M, M]$.
\begin{equation}
	\label{Eqn:expectedRuns}
	\frac{\mathbb{E} R}{\sqrt{n_1n_2}} = \int_x \frac{\sqrt{n_1 n_2} f(x) g(x)}{n_1 f(x)+ n_2 g(x)} dx + o(1),
\end{equation}
where $f(x)$, $g(x)$ are the corresponding PDF of two random variables with $n_1, n_2$ samples, respectively.

%% The formal theorem

\textbf{Proof.}
Before the proof, we give the following properties about the runs.

\textbf{Property 1.} The runs will decrease if we delete some terms in the sequence.

\textbf{Property 2.} The runs will increase if we add some terms in the sequence.

We divide the support of $x$ into $|T|$ equal slices $x_0, x_1, \dots, x_{|T|+1}$, and consider the expected runs in the $i_{th}$ slice as $\int_{x_{i-1}}^{x_{i}} \frac{n_1 f(x) n_2 g(x)}{n_1 f(x) + n_2 g(x)}  dx$.
Note that in each slice, we do not calculate the last-one run, since when combining, the last run is considered separatable.

Therefore, there are two kinds of errors when calculating the expected runs.

\textbf{Combining errors.} When combining all slices, there are at most $|T|-1$ errors~(mismatch at the junction of two adjacent slices), since the expected runs within each slice will not affect each other when combining.

\textbf{Approximation errors.} Notice that in each slice, the distribution is not a ``uniform distribution'', it actually has some trend. 
Therefore, we use the above properties of runs to upper and lower bound the expected runs number in each slice.
The intuition is that, we can use the maximum position to calculate the upper bound (as if we add some terms to the sequence).

For the $i_{th}$ bar, denote the maximum/minimum value point of $f(x)$ as $v_i$, $r_i$, and the maximum/minimum value point of $g(x)$ as $u_i$, $s_i$, then clip the distribution to the maximum/minimum, so that the two distributions become uniform distributions.

\begin{equation*}
	\sum_i \frac{n_1 f(v_i) n_2 g(u_i)}{n_1 f(v_i) + n_2 g(u_i)} \Delta(x_i) \leq \mathbb{E} R^\prime \leq \sum_i \frac{n_1 f(r_i) n_2 g(s_i)}{n_1 f(r_i) + n_2 g(s_i)} \Delta x_i,
\end{equation*}
where $R^\prime$ is the expected runs number without considering the combining loss.
% A formal equation might be based on contional on some external samples.

We next prove that $\int_{x_{i-1}}^{x_{i}} \frac{n_1 f(x) n_2 g(x)}{n_1 f(x) + n_2 g(x)}  dx$ is close to both the upper and the lower bound.
First notice that there exists $\xi_i$ such that
\begin{equation*}
	\int_{x_{i-1}}^{x_{i}} \frac{n_1 f(x) n_2 g(x)}{n_1 f(x) + n_2 g(x)}  dx 
	=\frac{n_1 f(\xi_{i}) n_2 g(\xi_{i})}{n_1 f(\xi_{i}) + n_2 g(\xi_{i})}  \Delta x_i .
\end{equation*}

Then for $\forall a_i, b_i \in [x_{i-1}, x_{i}]$, we have
\begin{equation*}
	\begin{split}
		& |\int_{x_{i-1}}^{x_{i}} \frac{n_1 f(x) n_2 g(x)}{n_1 f(x) + n_2 g(x)}  dx - \frac{n_1 f(a_i) n_2 g(b_i)}{n_1 f(a_i) + n_2 g(b_i)} \Delta  x_i|\\
		=&[\frac{n_1 f(\xi_{i}) n_2 g(\xi_{i})}{n_1 f(\xi_{i}) + n_2 g(\xi_{i})} - \frac{n_1 f(a_i) n_2 g(b_i)}{n_1 f(a_i) + n_2 g(b_i)}] \Delta x_i \\
		=& [\frac{n_1 f(\xi_{i}) f(a_{i}) (g(\xi_{i})-g(b_{i})) + n_2 g(\xi_{i})g(b_{i}) (f(a_{i})-f(\xi_{i}))  }{(n_1 f(\xi_{i}) + n_2 g(\xi_{i}))   (n_1 f(a_i) + n_2 g(b_i))} ] n_1 n_2 \Delta x_i \\
		\leq & C n_1 n_2 (\Delta x_i)^2 [\frac{n_1 f(\xi_{i}) f(a_{i}) + n_2 g(\xi_{i})g(b_{i}) }{(n_1 f(\xi_{i}) + n_2 g(\xi_{i}))   (n_1 f(a_i) + n_2 g(b_i))} ] \\
		=&\mathcal{O}(\sqrt{n_1 n_2} (\Delta x_i)^2  C).
	\end{split}
\end{equation*}
%% the dependency of kappa?

By summation over $i$, we have

\begin{equation*}
	|\int_{x} \frac{n_1 f(x) n_2 g(x)}{n_1 f(x) + n_2 g(x)} dx - \sum_i \frac{n_1 f(a_i) n_2 g(b_i)}{n_1 f(a_i) + n_2 g(b_i)} \Delta  x_i|\\
	\leq \mathcal{O}(\sqrt{n_1 n_2} \Delta x_i M C).
\end{equation*}

Therefore, the upper and lower bounds are all close to the integral. 
We conclude that 
\begin{equation*}
	|\mathbb{E} R^\prime - \int_{x} \frac{n_1 f(x) n_2 g(x)}{n_1 f(x) + n_2 g(x)}  dx |
	\leq \mathcal{O}(\sqrt{n_1 n_2} \Delta x_i M C).
\end{equation*}

Therefore, we have
\begin{equation*}
	|\mathbb{E} R - \int_{x} \frac{n_1 f(x) n_2 g(x)}{n_1 f(x) + n_2 g(x)}  dx |\\
	\leq \mathcal{O}(\sqrt{n_1 n_2} \Delta x_i M C) + |T| .
\end{equation*}

By setting $|T| = (n_1 n_2)^{1/4} (M C)^{1/2}$, then 
\begin{equation*}
	|\mathbb{E} R - \int_{x} \frac{n_1 f(x) n_2 g(x)}{n_1 f(x) + n_2 g(x)}  dx |\\
	\leq \mathcal{O}((n_1 n_2)^{1/4} (M C)^{1/2}).
\end{equation*}

This leads to Lemma~\ref{lem:expectedRuns}.

\subsection{Lemma~\ref{lem: equ}}
\label{app:lemma2}

Runs number is an approximation of the Receiver Operating Characteristic~(ROC) which is used to evaluate the anomaly detection's performance in experiments. Lemma~\ref{lem: equ} illustrates this point from the perspective of the maximal runs number. 

\begin{lemma}[Maximal runs number]
	\label{lem: equ}
	Consider two distributions $f(x)$ and $g(x)$, then the expected runs number reaches the maximum when $f(x) = g(x)$ almost surely.
\end{lemma}

This is a fundamental property of the expected runs. Specifically, we prove that when $g(x) = f(x)$, the expected runs reach the maximum. Besides, Lemma~\ref{lem: equ} also validates the fact that the expected runs is a reasonable approximation to the ROC.

\textbf{Proof.} WLOG, we assume $n_1 > n_2$.
First, notice that with $f(x)$ and $g(x)$, the expected runs is 
$$\mathbb{E}R_1 = \int_x \frac{\inSample \outSample \inNoisePdf(x) \outNoisePdf(x)}{\inSample \inNoisePdf(x) + \outSample \outNoisePdf(x)}  dx.$$
when $f(x) = g(x)$ a.s., the expected runs is
$$\mathbb{E}R_2 = \int_x \frac{\inSample \outSample}{\inSample  + \outSample }  dx.$$
We calculate that 
$$
r = \frac{\mathbb{E}R_1}{\mathbb{E}R_2} = \int_x  \frac{\inNoisePdf(x) \outNoisePdf(x) (\inSample + \outSample)}{\inSample\inNoisePdf(x) +\outSample\outNoisePdf(x)} dx,
$$

where $\int_x \inNoisePdf(x) dx = \int_x \outNoisePdf(x) dx = 1$. 
We expect $r < 1$ when $f(x) \not = g(x)$ .

We can derive that 
$$
r = 1 + \frac{1}{2} \int_x \frac{\inSample \inNoisePdf(x) - \outSample \outNoisePdf(x)}{\inSample \inNoisePdf(x) + \outSample \outNoisePdf(x)} (\outNoisePdf(x) - \inNoisePdf(x))  dx.
$$

Therefore, it suffices to show that 
$$
\int_x \frac{\inSample \inNoisePdf(x) - \outSample \outNoisePdf(x)}{\inSample \inNoisePdf(x) + \outSample \outNoisePdf(x)} (\outNoisePdf(x) - \inNoisePdf(x))  dx \leq 0.
$$

We derive the left side as follows
\begin{equation}
	\label{Eqn: long}
	\begin{split}
		& \int_x \frac{\inSample \inNoisePdf(x) - \outSample \outNoisePdf(x)}{\inSample \inNoisePdf(x) + \outSample \outNoisePdf(x)} (\outNoisePdf(x) - \inNoisePdf(x)) dx \\
		=&  \int_x \frac{(\inSample-\outSample)(\inNoisePdf(x) + \outNoisePdf(x)) + (\inSample+\outSample)(\inNoisePdf(x) - \outNoisePdf(x))}{(\inSample+\outSample)(\inNoisePdf(x) + \outNoisePdf(x)) + (\inSample-\outSample)(\inNoisePdf(x) - \outNoisePdf(x))} (\outNoisePdf(x) - \inNoisePdf(x)) dx \\
		=&  \int_x \frac{(\inSample-\outSample)(\inNoisePdf(x) + \outNoisePdf(x))}{(\inSample+\outSample)(\inNoisePdf(x) + \outNoisePdf(x)) + (\inSample-\outSample)(\inNoisePdf(x) - \outNoisePdf(x))} (\outNoisePdf(x) - \inNoisePdf(x)) dx \\
		&+\int_x \frac{(\inSample+\outSample)(\inNoisePdf(x) - \outNoisePdf(x))}{(\inSample+\outSample)(\inNoisePdf(x) + \outNoisePdf(x)) + (\inSample-\outSample)(\inNoisePdf(x) - \outNoisePdf(x))} (\outNoisePdf(x) - \inNoisePdf(x)) dx.
	\end{split}
\end{equation}

For the first part, note that 
\begin{equation*}
	\begin{split}
		&\int_x \frac{(\inSample-\outSample)(\inNoisePdf(x) + \outNoisePdf(x))}{(\inSample+\outSample)(\inNoisePdf(x) + \outNoisePdf(x)) + (\inSample-\outSample)(\inNoisePdf(x) - \outNoisePdf(x))} (\outNoisePdf(x) - \inNoisePdf(x)) dx \\
		=& \int_x \frac{(\inSample-\outSample)(\inNoisePdf(x) + \outNoisePdf(x))}{(\inSample+\outSample)(\inNoisePdf(x) + \outNoisePdf(x)) + (\inSample-\outSample)(\inNoisePdf(x) - \outNoisePdf(x))} (\outNoisePdf(x) - \inNoisePdf(x)) dx \\
		&-  \frac{\inSample-\outSample}{\inSample+\outSample}\int_x \outNoisePdf(x) - \inNoisePdf(x)  dx \\
		=& \int_x \frac{\frac{(\inSample-\outSample)^2}{\inSample+\outSample}(\outNoisePdf(x) - \inNoisePdf(x))^2}{(\inSample+\outSample)(\inNoisePdf(x) + \outNoisePdf(x)) + (\inSample-\outSample)(\inNoisePdf(x) - \outNoisePdf(x))} dx.
	\end{split}
\end{equation*}

Therefore, we can derive from Equation~\eqref{Eqn: long} that 
\begin{equation*}
	\begin{split}
		&\int_x \frac{\inSample \inNoisePdf(x) - \outSample \outNoisePdf(x)}{\inSample \inNoisePdf(x) + \outSample \outNoisePdf(x)} (\outNoisePdf(x) - \inNoisePdf(x))  dx \\
		=& \int_x \frac{\frac{(\inSample-\outSample)^2}{\inSample+\outSample}(\outNoisePdf(x) - \inNoisePdf(x))^2}{(\inSample+\outSample)(\inNoisePdf(x) + \outNoisePdf(x)) + (\inSample-\outSample)(\inNoisePdf(x) - \outNoisePdf(x))}  dx \\
		&-\int_x \frac{(\inSample+\outSample)(\inNoisePdf(x) - \outNoisePdf(x))^2}{(\inSample+\outSample)(\inNoisePdf(x) + \outNoisePdf(x)) + (\inSample-\outSample)(\inNoisePdf(x) - \outNoisePdf(x))}  dx \\
		=& \int_x \frac{\frac{(\inSample-\outSample)^2 - (\inSample+\outSample)^2}{\inSample+\outSample}(\outNoisePdf(x) - \inNoisePdf(x))^2}{(\inSample+\outSample)(\inNoisePdf(x) + \outNoisePdf(x)) + (\inSample-\outSample)(\inNoisePdf(x) - \outNoisePdf(x))}  dx \\
		\leq& 0.
	\end{split}
\end{equation*}
The proof is done!

\subsection{Proof of Lemma~\ref{lem: change}}
\label{app:lemma3}
\begin{figure*}[htbp]
	\centering
	\subfigure[\label{fig:beta_inner_product_anime_res}Real distribution of ResNet-50]{
		\begin{minipage}[t]{0.45\linewidth}
			\centering
			\includegraphics[width=2.8in]{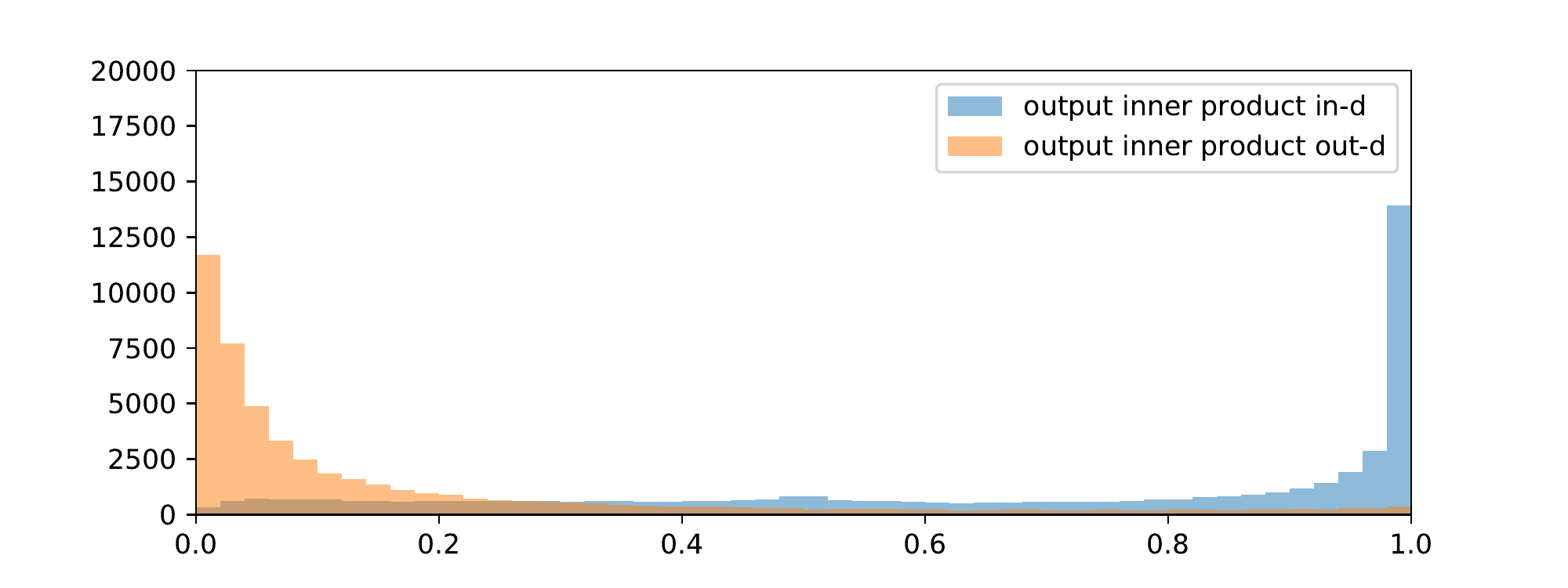}
			%\caption{fig1}
		\end{minipage}%
	}
	\subfigure[\label{fig:beta_inner_product_anime_dense}Real distribution of DenseNet-121]{
		\begin{minipage}[t]{0.45\linewidth}
			\centering
			\includegraphics[width=2.8in]{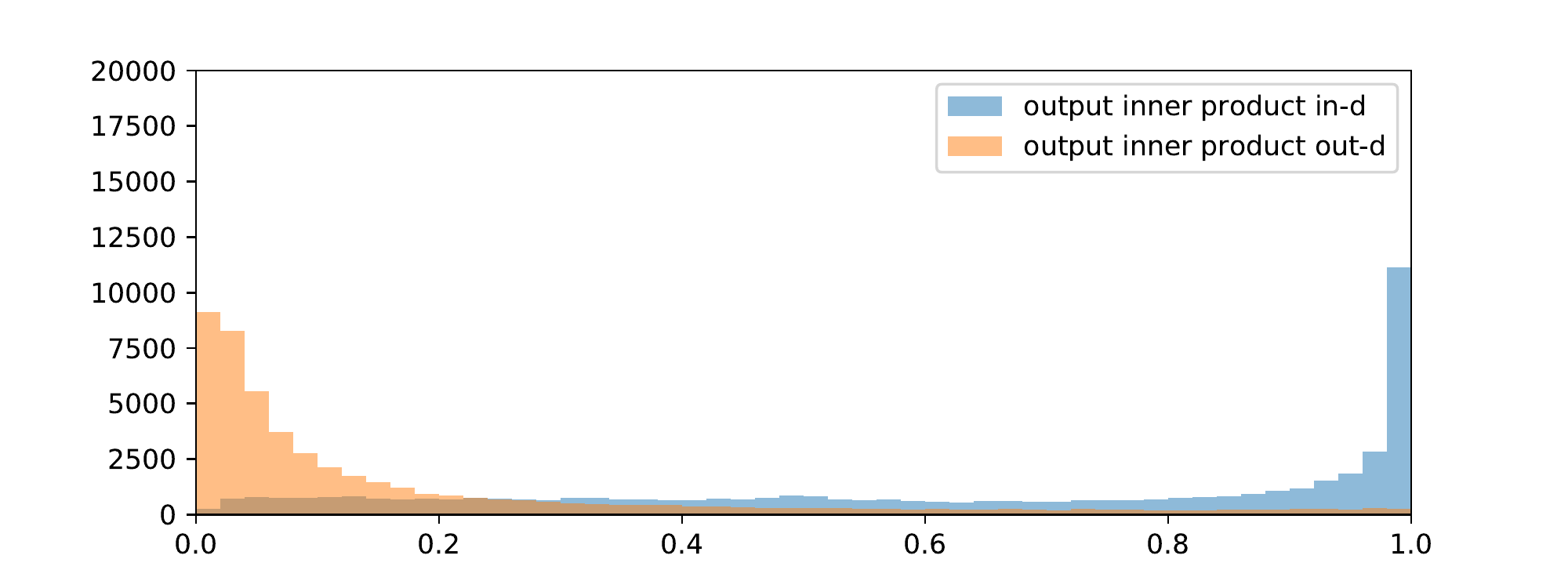}
			%\caption{fig1}
		\end{minipage}%
	}
	\\
	\subfigure[\label{fig:beta_inner_product_anime_beta_res}Beta distribution fitting for ResNet-50]{
		\begin{minipage}[t]{0.45\linewidth}
			\centering
			\includegraphics[width=2.8in]{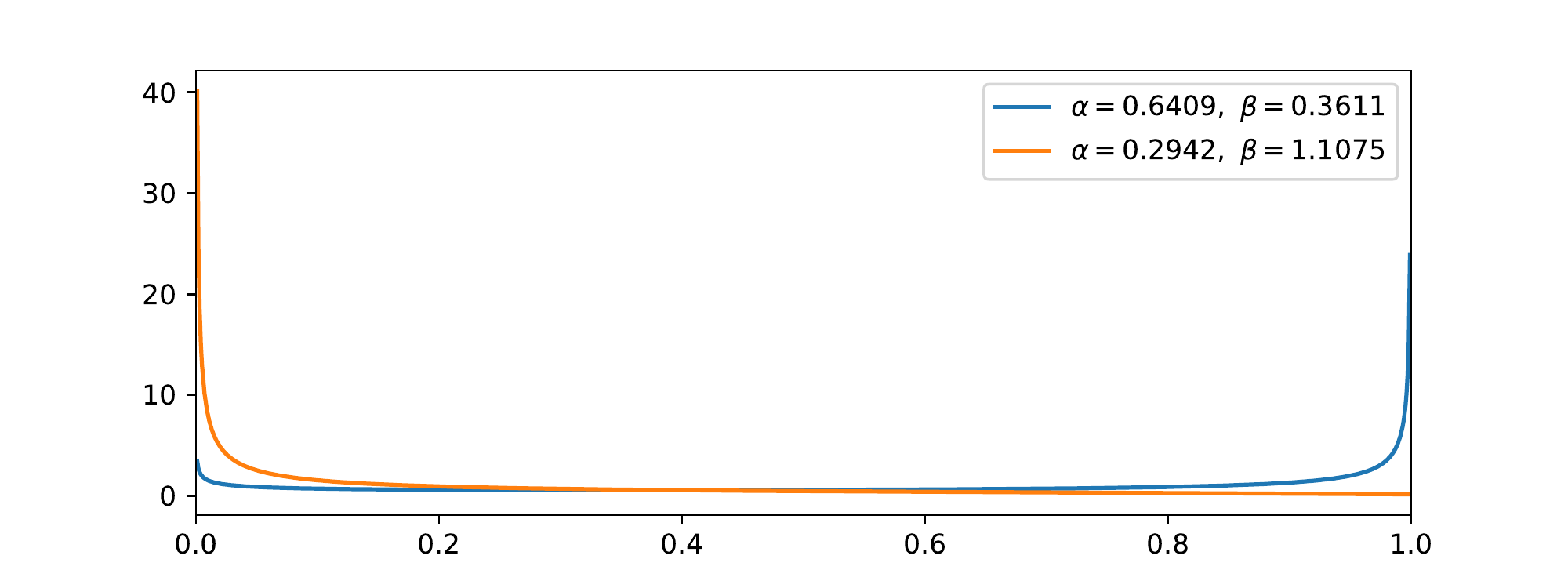}
			%\caption{fig1}
		\end{minipage}%
	}
	\subfigure[\label{fig:beta_inner_product_anime_beta_dense}Beta distribution fitting for DenseNet-121]{
		\begin{minipage}[t]{0.45\linewidth}
			\centering
			\includegraphics[width=2.8in]{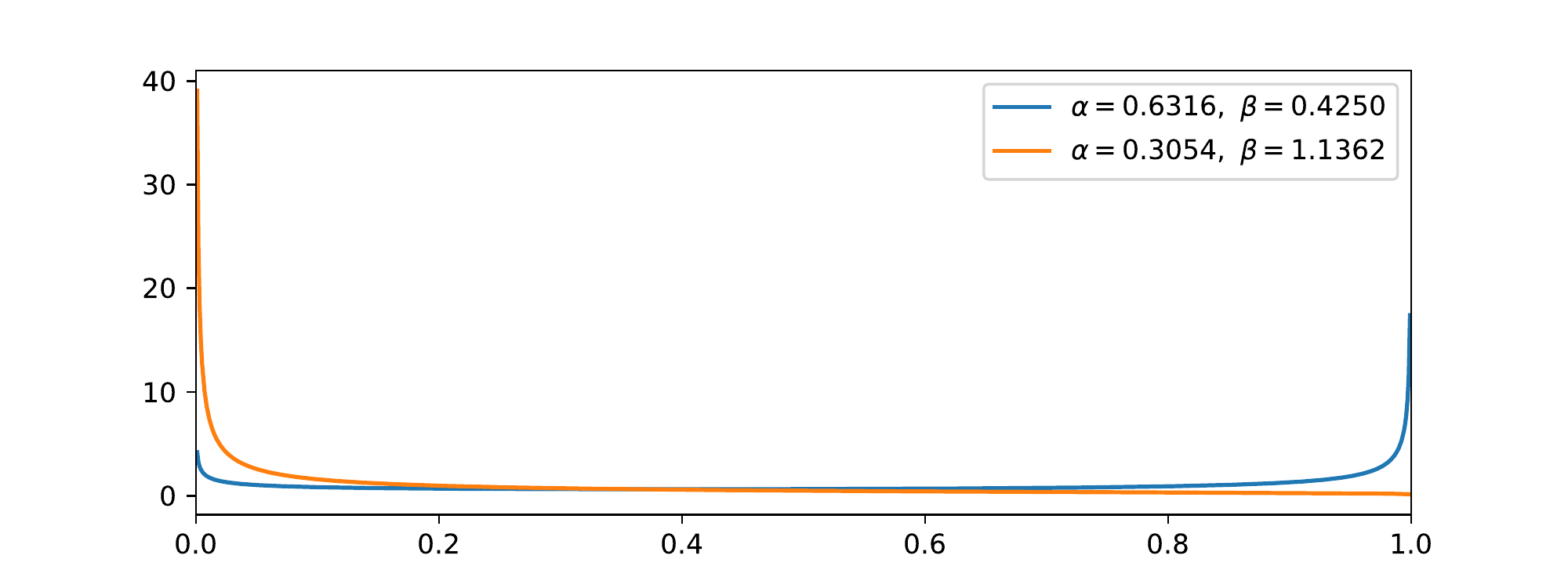}
			%\caption{fig1}
		\end{minipage}%
	}
	\centering
	\caption{\label{fig:beta_inner_product}The in-distribution~(ImageNet) and out-distribution~(Artificial dataset) trained on two different architectures.}
\end{figure*}

First, let us see the real in-distributions and out-distributions. Here we show two examples of \textit{ImageNet vs. Artificial} dataset in Figure~\ref{fig:beta_inner_product_anime_res} and Figure~\ref{fig:beta_inner_product_anime_dense}. Then we run Beta fitting on the two examples and get Figure~\ref{fig:beta_inner_product_anime_beta_res} and Figure~\ref{fig:beta_inner_product_anime_beta_dense}. From the figures, we can see it is reasonable to model the in- and out- distributions as Beta distributions. 

Next, let us focus on Lemma~\ref{lem: change}. 
Set $f(x) = \frac{\Gamma(\alpha_1)\Gamma(\beta_1)}{\Gamma(\alpha_1+\beta_1)} x^{\alpha_1-1} (1-x)^{\beta_1 - 1}$ and $g(x) = \frac{\Gamma(\alpha_2)\Gamma(\beta_2)}{\Gamma(\alpha_2+\beta_2)} x^{\alpha_2-1} (1-x)^{\beta_2 - 1}$.
We rewrite the expected runs as follows (we omit the small-o term):
\begin{equation*}
	\begin{split}
		&\mathbb{E} R (\alpha_1, \beta_1, \alpha_2, \beta_2) \\
		=& \int_{x\in\mathcal{X}} \frac{n_1 n_2 f(x)g(x)}{n_1 f(x) + n_2 g(x)}  dx \\
		=& \int_0^1 \frac{\frac{\Gamma(\alpha_2)\Gamma(\beta_2)}{\Gamma(\alpha_2+\beta_2)} x^{\alpha_2-1} (1-x)^{\beta_2 - 1}}{\kappa + \frac{\Gamma(\alpha_2)\Gamma(\beta_2)}{\Gamma(\alpha_2+\beta_2) \Gamma(\beta_1)} (1-x)^{\beta_2 - \beta_1} \frac{\Gamma(\alpha_1+\beta_1) }{\Gamma(\alpha_1)}x^{\alpha_2-1} } dx.
	\end{split}
\end{equation*}

We make a derivation on $\alpha_1$ and get
\begin{equation*}
	\frac{\partial \mathbb{E} R (\alpha_1, \beta_1, \alpha_2, \beta_2)}{\partial \alpha_1}\\
	=\int_0^1 C(x, \alpha_1, \beta_1, \alpha_2, \beta_2, n_1, n_2) [(\alpha_2 - \alpha_1) - (\psi_0(\alpha_1 + \beta_1) - \psi_0(\alpha_1) )x] dx.
\end{equation*}
where $C(x, \alpha_1, \beta_1, \alpha_2, \beta_2, n_1, n_2) \geq 0$, and $\psi_0(z)$ is defined as Digamma function, which is $\psi_0(z) = \Gamma^\prime(z) / \Gamma(z)$.

We make some discussions on the Digamma function $\psi_0$.

\textbf{Property 1.} $\psi_0(z)$ is increasing with respect to the variable $z$.

According to its explict formula $\psi_0(z+1) = -\gamma + \int_0^1 \frac{1-t^{z-1}}{1-t}$, and $\gamma$ is the Euler–Mascheroni constant, we reach the above conclusion.

\textbf{Property 2.} $\psi_0(z+1) - \psi_0(z) = 1/z$.

Therefore,
$ (\alpha_2 - \alpha_1) - (\psi_0(\alpha_1 + \beta_1) - \psi_0(\alpha_1)) \leq (\alpha_2 - \alpha_1) - (\psi_0(\alpha_1 + \beta_1) - \psi_0(\alpha_1) )x \leq (\alpha_2 - \alpha_1)$.

Therefore, when $ \alpha_2 - \alpha_1 \leq 0$, we have $\frac{\partial \mathbb{E} R (\alpha_1, \beta_1, \alpha_2, \beta_2)}{\partial \alpha_1} \leq 0$; when $ \alpha_2 - \alpha_1 \geq  \sum_{k=0}^{\lfloor \beta_1\rfloor} \frac{1}{\alpha_1 + k}$, we have $\frac{\partial \mathbb{E} R (\alpha_1, \beta_1, \alpha_2, \beta_2)}{\partial \alpha_1} \geq 0$.

\section{Transformation examples}
\label{app:transformationexamples}
\begin{figure*}[htbp]
	\centering
	\subfigure[Raw input]{
		\begin{minipage}[t]{0.96\linewidth}
			\centering
			\includegraphics[width=1.2in]{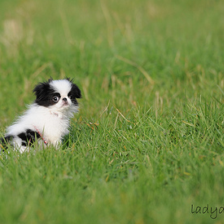}
			\includegraphics[width=1.2in]{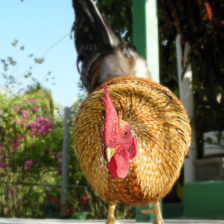}
			\includegraphics[width=1.2in]{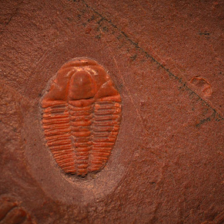}
			\includegraphics[width=1.2in]{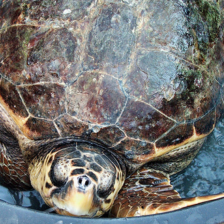}
			%\caption{fig1}
		\end{minipage}%
	}
	\subfigure[$\text{FFT}_{40}$]{
		\begin{minipage}[t]{0.96\linewidth}
			\centering
			\includegraphics[width=1.2in]{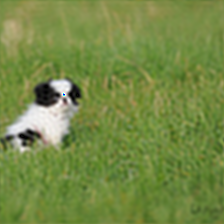}
			\includegraphics[width=1.2in]{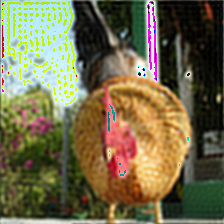}
			\includegraphics[width=1.2in]{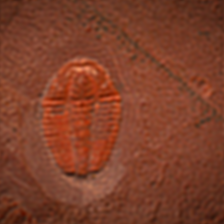}
			\includegraphics[width=1.2in]{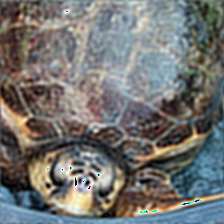}
			%\caption{fig1}
		\end{minipage}%
	}
	\subfigure[$\text{FFT}_{100}$]{
		\begin{minipage}[t]{0.96\linewidth}
			\centering
			\includegraphics[width=1.2in]{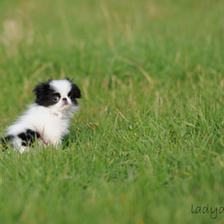}
			\includegraphics[width=1.2in]{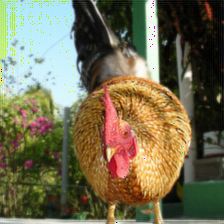}
			\includegraphics[width=1.2in]{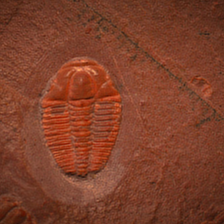}
			\includegraphics[width=1.2in]{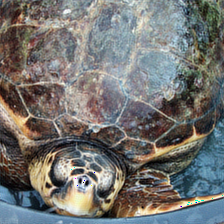}
			%\caption{fig1}
		\end{minipage}%
	}
	\subfigure[Horizontal flip]{
		\begin{minipage}[t]{0.96\linewidth}
			\centering
			\includegraphics[width=1.2in]{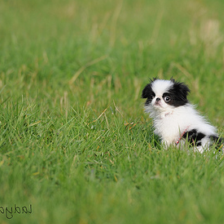}
			\includegraphics[width=1.2in]{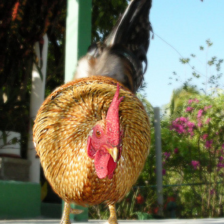}
			\includegraphics[width=1.2in]{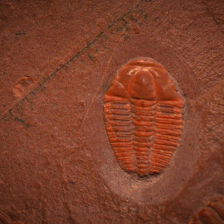}
			\includegraphics[width=1.2in]{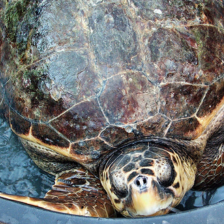}
			%\caption{fig1}
		\end{minipage}%
	}
	
	\caption{\label{fig:disentanglingmethod}ImageNet data transformation examples. For the FFT method, different filter radii bring different degrees of filtering to image details. For example, the grass details in the first row and the texture details of the turtle shells in the fourth row.}
\end{figure*}

Here we show several transformation examples from the ImageNet dataset in Figure~\ref{fig:disentanglingmethod}.

\section{Detailed computational cost description of highly related algorithms}
\label{app:computational cost}
In Table~\ref{table:comparemethods}, we briefly summarize the extra training and testing computational cost of highly related~(standard classifier-based) algorithms. But we did not provide details because of space limitation. In this section, we will describe the details to clarify. 

\subsection{ODIN}
After obtaining a standard classification model, ODIN requires input pre-processing, which first needs a forward pass to compute the loss, then a backpropagation to modify the input, and finally a forward pass again to calculate the final predicted probability values. The searching process of hyperparameters like noisy magnitude and temperature demands prior out-distribution knowledge.

\subsection{Mahalanobis}
Mahalanobis method assumes that the features of the in-distribution dataset follow a Gaussian distribution. It first needs forward all training samples to compute feature mean and variance. And then, it adopts a similar idea from input pre-processing, thus requires two forward passes and one backward pass to record the internal feature at test time. Meanwhile, the Mahalanobis method does a grid search across k (pre-determined)~internal layers and several noise magnitude values. The best setting is selected with the help of prior out-distribution knowledge. After recording all the features of the test samples, it finally trains a regression model and gives final anomaly scores of test samples. 

\subsection{Ours}
The computational cost of TTA-AD is almost the same as doubling testing a test point. We only need two forward passes per sample to compute the anomaly score. The pipeline is shown in Algorithm~\ref{alg:Framwork}.

\subsection{Running time of classifier based algorithms}
\label{app:runningtime}
\begin{figure}[htbp]
	\centering
	\includegraphics[width=2.3in]{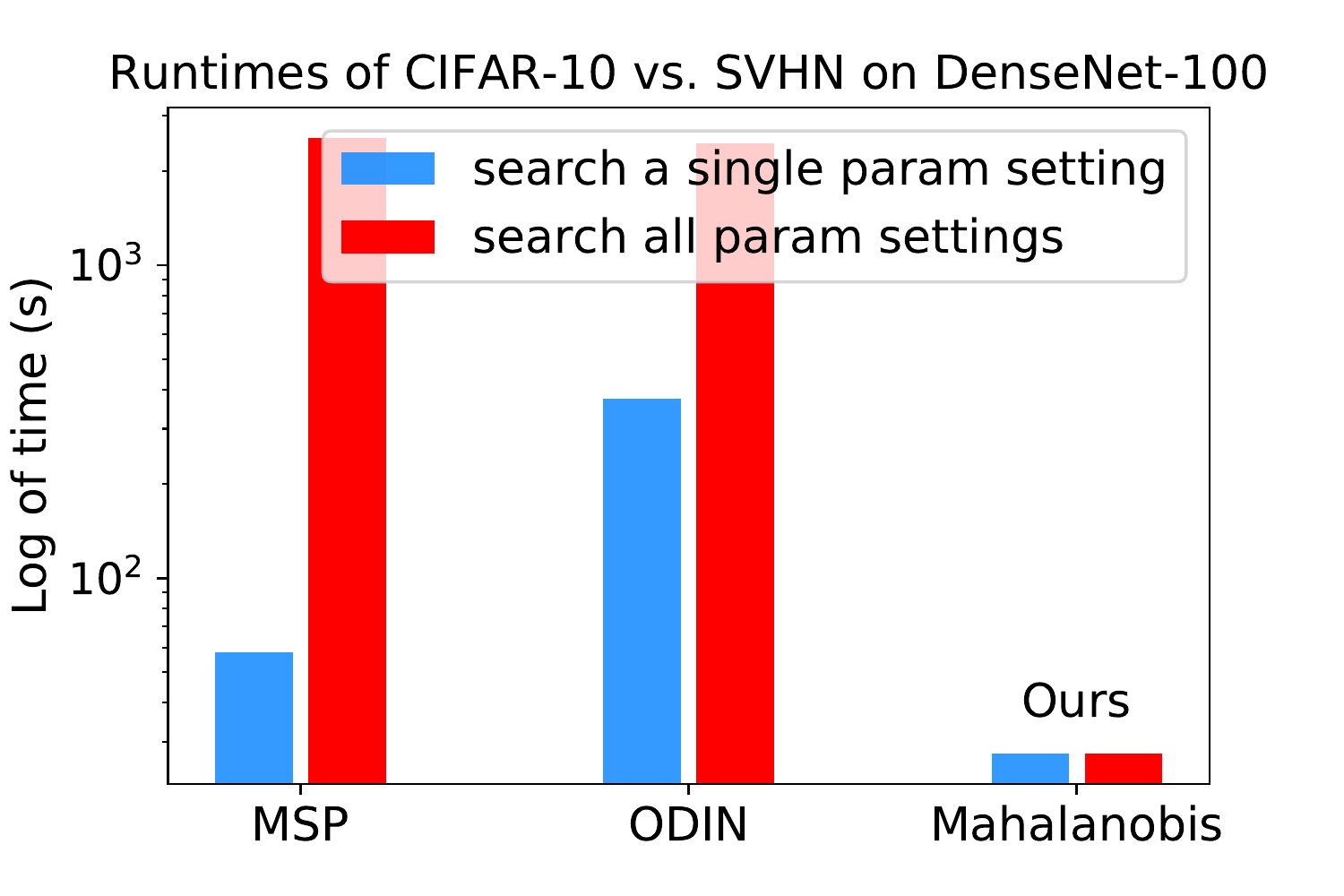}
	\caption{\label{fig:runtime_den}Average algorithm running time on DenseNet-100. }
\end{figure}

For experiments in this section, we run all the running time experiments on an RTX 2080Ti graphic card. We only consider the CIFAR-10 vs. SVHN setting since the proportion of running time does not change with different datasets. For DenseNet-100, we conduct experiments similarly and the result is shown in Figure~\ref{fig:runtime_den}. And we omit the common classifier training time. 

%The running time figures~(Figure~\ref{fig:runtime} and \ref{fig:runtime_den}) only consider classifier-based algorithms: baseline~\cite{oodbaseline}, ODIN~\cite{ODIN}, and Mahalanobis~\cite{Ma}, because these three algorithms are the most relevant. All the three algorithms, plus ours, depend on the pre-trained classifier, so we omit the classifier training time. For the DenseNet-100 running time, see Figure~\ref{fig:runtime_den}. 

For ODIN, we slightly simplify the search space. According to the original ODIN paper, there are 210 times of searches in total. We only choose 44 among the 210 settings, which are useful in the CIFAR-10 vs. SVHN setting~\cite{ODIN}. The Mahalanobis algorithm demands recording the mean and variance of the training samples, which takes about 19 seconds for the CIFAR-10 dataset in ResNet-34. This process will take more time if the training sample size increases. We follow the hyperparameter searching settings, which include 7 independent searches. 

We do not mention other algorithms because they adopt other training ways which are significantly different from supervised training. For example, GAN-based algorithms require GAN training, which highly depends on the GAN algorithm. And obtaining a satisfying GAN usually takes a longer time than supervised classifier training. Contrastive learning aided algorithms~\cite{CSI} use contrastive methods~(SimCLR~\cite{simclr}) to train a model, which takes much more time than standard classifier training. For example, SimCLR still gains performance improvement after 800 epochs training~\cite{simclr} and the computation of contrastive losses is more complicated. Considering the significant difference in pre-training time, we only compare the running time of our algorithm with these relatively lightweight standard classifier-based algorithms. 

\section{Attempts for adapting previous algorithm}
\label{app:attempts_to_improve}
\subsection{Network architecture}
For CIFAR classifiers, we simply follow the common architectures which are also used by Liang et al.~\cite{ODIN} and Lee et al.~\cite{Ma}. 

CIFAR models can not be directly used in ImageNet images. Following common configurations, we resize the ImageNet figures into 224x224, and use \textit{torchvision}'s official ImageNet models: ResNet-50 and DenseNet-121 from \url{https://pytorch.org/vision/stable/models.html}. Standard normalization is added when pre-processing the images. For ImageNet subset training, we only change the output size of the last layer to the corresponding class numbers. 

\subsection{OC-SVM~(One Class-SVM)}
OC-SVM~\cite{OCSVM} is a general anomaly detection algorithm that trains a one-class SVM on a given~(feature) dataset. The input of OC-SVM can be obtained from autoencoders, GANs, and plain classifiers. In our case, the classifier's final output feature is used to train the OC-SVM. The only hyperparameter is $\nu$ bounded between 0 and 1, which acts as an upper bound on the fraction of margin errors and a lower bound of the fraction of support vectors relative to the total number of training examples. For both CIFAR and ImageNet settings, we linearly search the $\nu$ from 0.1 to 0.9 in intervals of 0.1. 

The failure of OC-SVM on complex tasks is not surprising since it is usually used to perform anomaly detection at a more fine level, i.e., one-class-vs-all anomaly detection. For example, in the CIFAR-10 dataset, one of the classes is treated as in-distribution data while all other nine classes are considered as out-distribution data. But in our setting, multiple classes are treated as in-distribution data on the whole. One possible explanation for the failure is that we use larger models and harder classification tasks. So the features are more complicated than that of one-class-vs-all CIFAR, making the OC-SVM hard to find a good hyperplane. 

\subsection{MSP}
The MSP algorithm~\cite{oodbaseline} can be directly adopted to different datasets because its criteria is the maximum probability value. And stable performance across low-resolution datasets and high-resolution datasets can be observed in Table~\ref{table:imgnet}, Table~\ref{table:imagenetresult} and~Table~\ref{table:cifarresult}, .

\subsection{ODIN}
The major difference between ODIN~\cite{ODIN} and the MSP method is that ODIN uses two techniques, input-preprocessing and temperature scaling, to enlarge the gap between the in-distribution and out-distribution samples in maximum predicted value. For both CIFAR and ImageNet subset settings, we do a grid search for noise value from [0, 0.0005, 0.001, 0.0014, 0.002, 0.0024, 0.005, 0.01, 0.05, 0.1, 0.2], and temperature value from [1, 10, 100, 1000]. 

ODIN and the MSP method do not tune the feature space of deep neural networks, and reach good performance in both the ImageNet subset settings~(Table~\ref{table:imagenetresult}) and CIFAR~(Table~\ref{table:cifarresult}). 

\subsection{Mahalanobis algorithm}
\label{app:ma_algorithm_tune}
For the CIFAR setting, we directly use the official setting mentioned in~\cite{Ma}. For the ImageNet subset settings, we use the official ResNet-50 and DenseNet-121 architecture from torchvision. The choice of internal output features is the same as the CIFAR setting: we choose the output of the first convolution layer and output of every transition layer for DenseNet, and the input of the first residual block and the output of every residual block for ResNet. 

We also expand the hyperparameter search space. Specifically, we search the noise magnitude of the Mahalanobis method from [0.0, 0.5, 0.2, 0.1, 0.05, 0.02, 0.01, 0.005, 0.002, 0.0014, 0.001, 0.0005], which expands the original search space a lot. The validation dataset size is defined as 10\% of the whole test set size, which is the same as the original paper.

\section{Classifier model training}
\label{app:classifier_training}

%resnet train test 100.000 94.710  99.998 95.000   100.000 94.800
%densenet train test 99.996 94.78   99.992 94.470   99.998 94.230 

Since we find the performance of classifier-based algorithms is highly based on the pre-trained classifier, e.g. different random seeds, we train three models for all settings and report the mean and variance. Standard training data augmentation methods are adopted, including resizing, padding and flipping. For full ImageNet training, we directly use pre-trained models from torchvision\footnote{\url{https://pytorch.org/docs/stable/torchvision/models.html}}. The validation accuracy is $76.01\%$ for ResNet-50 and $74.47\%$ for DenseNet-121. 

For ImageNet subset training, we use the official implementation of ResNet-50 and DenseNet-121 models from \textit{torchvision}. We train 200 epochs using SGD with a stepping learning rate.  The training and test accuracy is shown in Table~\ref{table:imagenetsubsettrain}.

\begin{table*}[h]
	\caption{\label{table:imagenetsubsettrain} Classifier training and test accuracy~(\%) on ImageNet subset.}
	\centering
	\begin{tabular}{ccccc}
		\toprule[1pt]
		\multirow{2}{*}{Subset} & \multicolumn{2}{c}{ResNet-50}           & \multicolumn{2}{c}{DenseNet-121}        \\ \cline{2-5} 
		& Training           & Test               & Training           & Test               \\ \toprule[1pt]
		Living 9                & $96.00\pm0.35$ & $75.91\pm0.70$ & $96.23\pm0.05$ & $76.86\pm0.35$ \\
		Geirhos 16              & $93.56\pm2.79$ & $64.61\pm0.10$ & $96.39\pm0.11$ & $67.06\pm0.44$ \\
		Mixed 10                & $95.37\pm0.16$ & $80.37\pm0.13$ & $95.90\pm0.26$ & $80.24\pm0.46$ \\ \toprule[1pt]
	\end{tabular}
\end{table*}

For the CIFAR-10 dataset, we follow the training configuration from  Lee et al.~\cite{Ma}. And the training and test result is shown in Table~\ref{table:cifartrain}. 

\begin{table}[h]
	\caption{\label{table:cifartrain}Classifier training and test accuracy~(\%) on CIFAR-10.}
	\centering
	\begin{tabular}{ccc}
		\toprule[1pt]
		& Training acc.      & Test acc.         \\ \toprule[1pt]
		ResNet-34    & 100.00$\pm$0.00 & 94.84$\pm$0.12 \\
		DenseNet-100 & 100.00$\pm$0.00 & 94.49$\pm$0.23 \\ \toprule[1pt]
	\end{tabular}
\end{table}

\section{Generalization of anomaly detection and CIFAR results}
\label{app:cifar_results}

As mentioned by~\cite{zhou2021step}, when exposed to a new different out-distribution dataset, some previous algorithms suffer from severe performance degradation due to a severe bias introduced by validation bias. In the main text, TTA-AD has shown better average performance on advanced dataset settings like ImageNet. To verify whether our data augmentation and consistency evaluation pipeline has similar bias to anomaly detection datasets, we conduct CIFAR experiments to test whether there is a generalization problem of TTA-AD. 

The performance on CIFAR-10 settings are shown in Table~\ref{table:cifarresult}. Specifically, the CIFAR-10 dataset is treated as in-distribution data, while the SVHN~\cite{svhn}, TinyImageNet~\cite{tinyimagenet}, and LSUN~\cite{lsun} datasets are out-distribution datasets. TTA-AD reaches the second best results with other algorithms on most settings. At the same time, we should notice that the best algorithm of CIFAR setting in Table~\ref{table:cifarresult}, i.e. the Mahalanobis, suffers from sever performance degradation in ImageNet settings Table~\ref{table:imagenetresult}, which is consistent with the observations from~\cite{zhou2021step}. One possible reason is that the features of large network architectures and complicated datasets are not easy to be modeled as Gaussian distributions, which is the main assumption of the Mahalanobis algorithm. Another possible reason is that for ImageNet subset training, the test set size is not as large as CIFAR, so the hyperparameter searching process is difficult on ImageNet settings. 

At the same time, our TTA-AD not only reaches higher average detection performance, but also shows stable generalization ability to low-resolution dataset settings.

\begin{table*}[h]
	\caption{\label{table:cifarresult}AUROC~(\%) of CIFAR results. Best results are in \textbf{bold}. Second best results are \underline{underlined}. }
	\centering
	\begin{tabular}{clccc}
		\toprule[1pt]
		\multirow{2}{*}{Architecture} & \multirow{2}{*}{Algorithm} & \multicolumn{3}{c}{Out-distribution data}                                                                                            \\ \cline{3-5} 
		&                            & SVHN                                       & ImageNet\_resize                           & LSUN\_resize                               \\ 
		\toprule[1pt]
		\multirow{5}{*}{ResNet-34}    & MSP                   & $88.86\pm0.96$                         & $86.98\pm2.01$                         & $90.28\pm1.13$                         \\
		& OC-SVM                       & $90.34\pm0.45$  &\underline{$89.64\pm0.99$}   & $91.07\pm0.50$	           \\
		& ODIN                       & $90.84\pm1.71$                         & $88.64 \pm2.77 $                       & $91.87 \pm2.31 $                       \\
		& Mahalanobis                & \bm{$98.50 \pm0.12 $} & \bm{$99.39 \pm0.06 $} & \bm{$99.68 \pm0.07 $} \\
		& $\text{TTA-AD}_{\text{FFT}(t=1)}$               & \underline{$92.95 \pm0.89 $}                       & $89.15 \pm1.93 $                       & \underline{$92.53 \pm0.85 $}                       \\
		& $\text{TTA-AD}_{\text{Flip}}$                     & $88.59\pm0.84$                         & $88.01 \pm1.57 $                       & $90.67\pm0.71$                         \\ \hline
		\multirow{5}{*}{DenseNet-100} & MSP                   & $86.80 \pm1.64 $                       & $92.92 \pm0.87 $                       & $93.90 \pm0.66 $                       \\
		& OC-SVM                       & $77.54\pm2.45$
		& $80.93\pm1.86$
		& $82.22\pm0.63$
		\\
		& ODIN                       & $90.76\pm0.34$                         & \bm{$97.73 \pm0.55 $}                       & \bm{$98.57 \pm0.34} $                       \\
		& Mahalanobis                & \bm{$95.61 \pm2.23 $} & $91.20 \pm1.29 $                       & $80.49 \pm17.67 $                      \\
		& $\text{TTA-AD}_{\text{FFT}(t=1)}$                 & \underline{$91.15 \pm3.17 $}                       & \underline{$96.48 \pm0.94 $}                       & \underline{$97.88 \pm0.52 $}                       \\
		& $\text{TTA-AD}_{\text{Flip}}$                       & $86.64 \pm1.62 $                       & $92.64 \pm0.77 $                       & $93.48 \pm0.56 $                       \\ 
		\toprule[1pt]
	\end{tabular}
\end{table*}

\section{Sensitivity to filter radius}
\label{app:sensitivitytofilterradius}
\begin{figure}[h]
	\centering
	\begin{minipage}[t]{0.48\textwidth}
		\centering
		\includegraphics[width=2.6in]{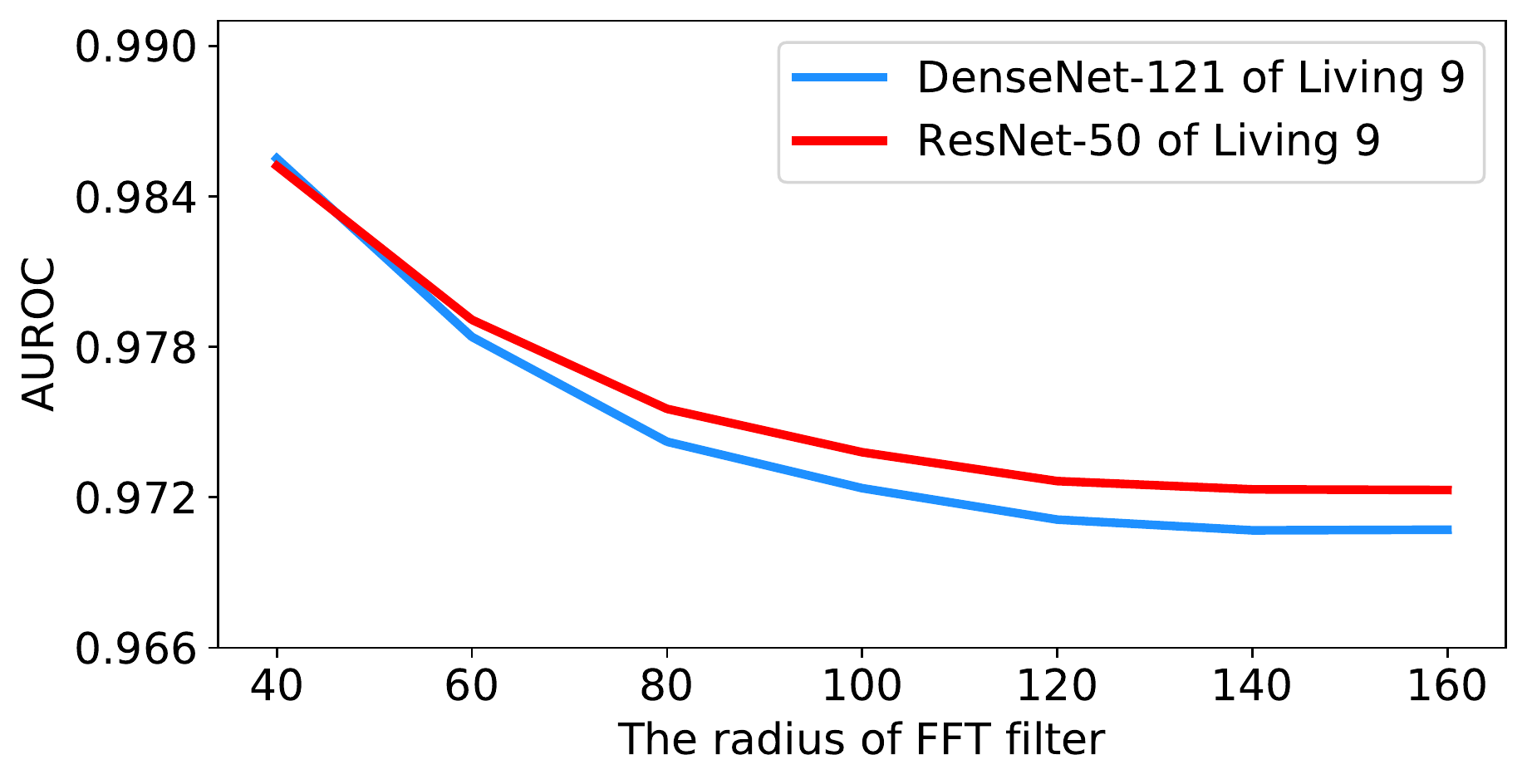}
		\caption{\label{fig:freqsensitivity1}Detection AUROC of Living 9 setting with different radius of FFT filter.}
	\end{minipage}
	%	\begin{minipage}[t]{0.48\textwidth}
	%		\centering
	%		\includegraphics[width=2.6in]{figure/temp/freq_effect_ge16.pdf}
	%		\caption{\label{fig:freqsensitivity2}Detection AUROC of Geirhos 16 setting with different radius of FFT filter.}
	%	\end{minipage}
	\begin{minipage}[t]{0.48\textwidth}
		\centering
		\includegraphics[width=2.6in]{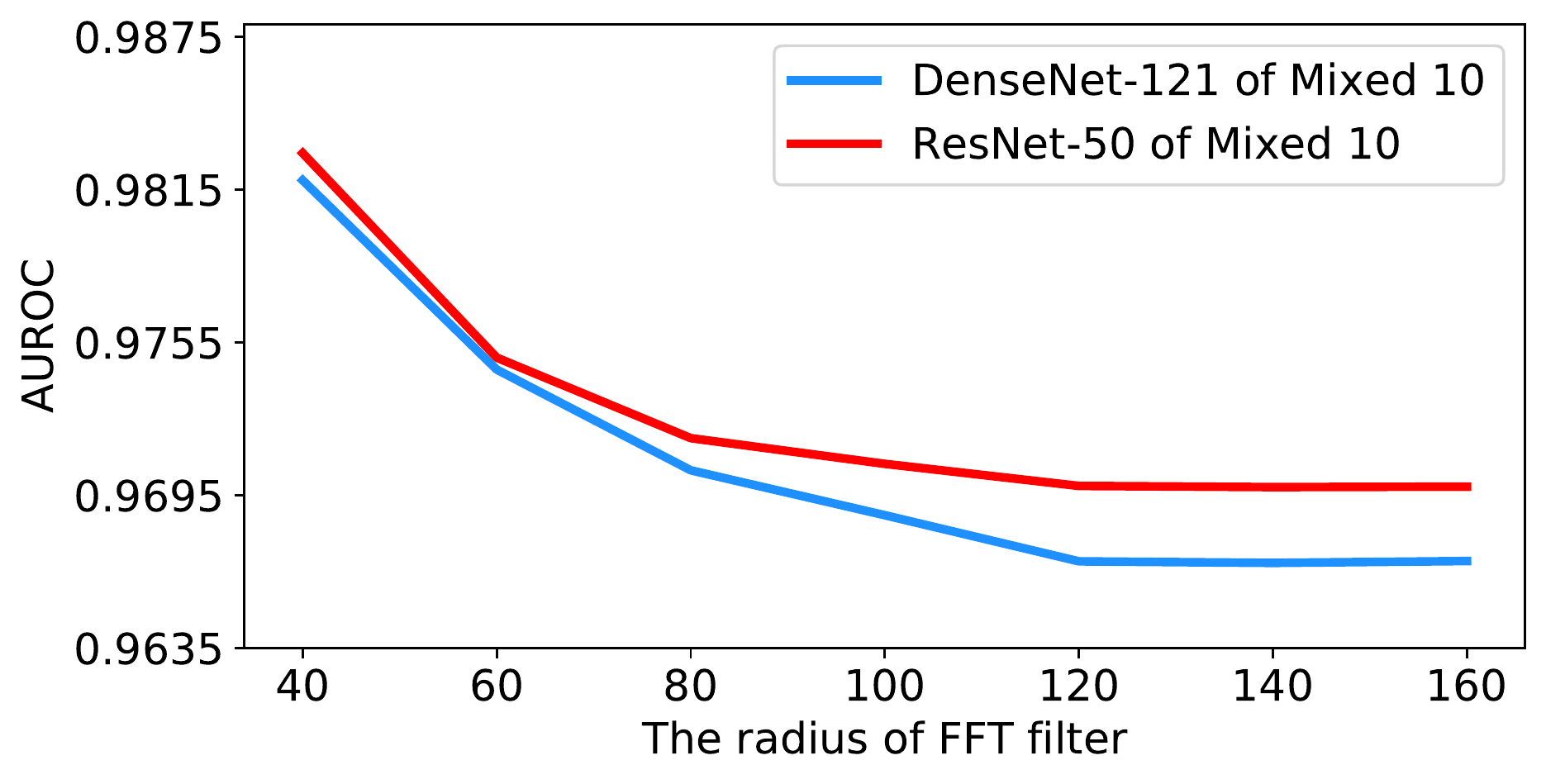}
		\caption{\label{fig:freqsensitivity3}Detection AUROC of Mixed 10 setting with different radius of FFT filter.}
	\end{minipage}
\end{figure}

More sensitivity results for ImageNet subsets are in Figure~\ref{fig:freqsensitivity1} and Figure~\ref{fig:freqsensitivity3}. In these figures, flatter curves mean less sensitivity to filter radius. For ImageNet subset settings, the optimal FFT filter radius seems to be much lower than that of the full ImageNet setting. A possible explanation is that the frequency domain characteristic of the Artificial dataset is different from the ImageNet dataset because artificial images are drawn by humans which lack fine-grained changes~(high frequency signals). Meanwhile, we can see from Table~\ref{table:imagenetresult} that no matter which radius is used, the performance is still good enough to outperform other algorithms. 

%\begin{figure}[h]
%	\centering
%	\includegraphics[width=3.5in]{figure/temp/freq_effect_living9.pdf}
%	\caption{\label{fig:freqsensitivity1}Detection AUROC of Living 9 setting with different radius of FFT filter.}
%\end{figure}
%\begin{figure}[h]
%	\centering
%	\includegraphics[width=3.5in]{figure/temp/freq_effect_ge16.pdf}
%	\caption{\label{fig:freqsensitivity2}Detection AUROC of Geirhos 16 setting with different radius of FFT filter.}
%\end{figure}

%\begin{figure}[h]
%	\centering
%	\includegraphics[width=3.5in]{figure/temp/freq_effect_mixed10.pdf}
%	\caption{\label{fig:freqsensitivity3}Detection AUROC of Mixed 10 setting with different radius of FFT filter.}
%\end{figure}

\section{Sensitivity to temperature}
\label{app:sensitivitytotemperature}
\begin{figure}[!t]
	\vspace{-450pt}
	\centering
	\begin{minipage}[h]{0.48\textwidth}
		\centering
		\includegraphics[width=2.6in]{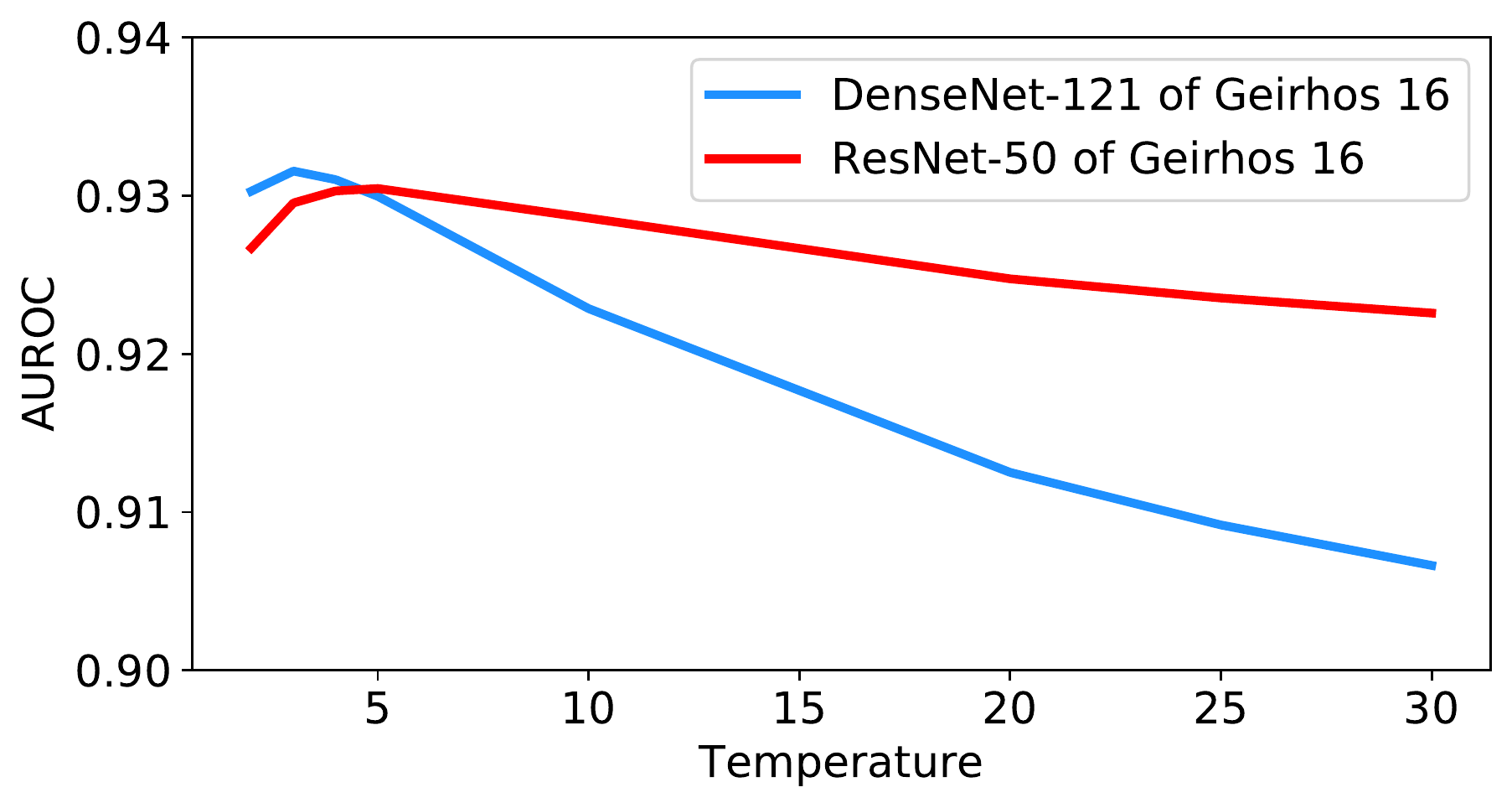}
		\caption{\label{fig:tempsensitivity2}Detection AUROC of Geirhos 16 setting with different temperature.}
	\end{minipage}
	\begin{minipage}[h]{0.48\textwidth}
		\centering
		\includegraphics[width=2.6in]{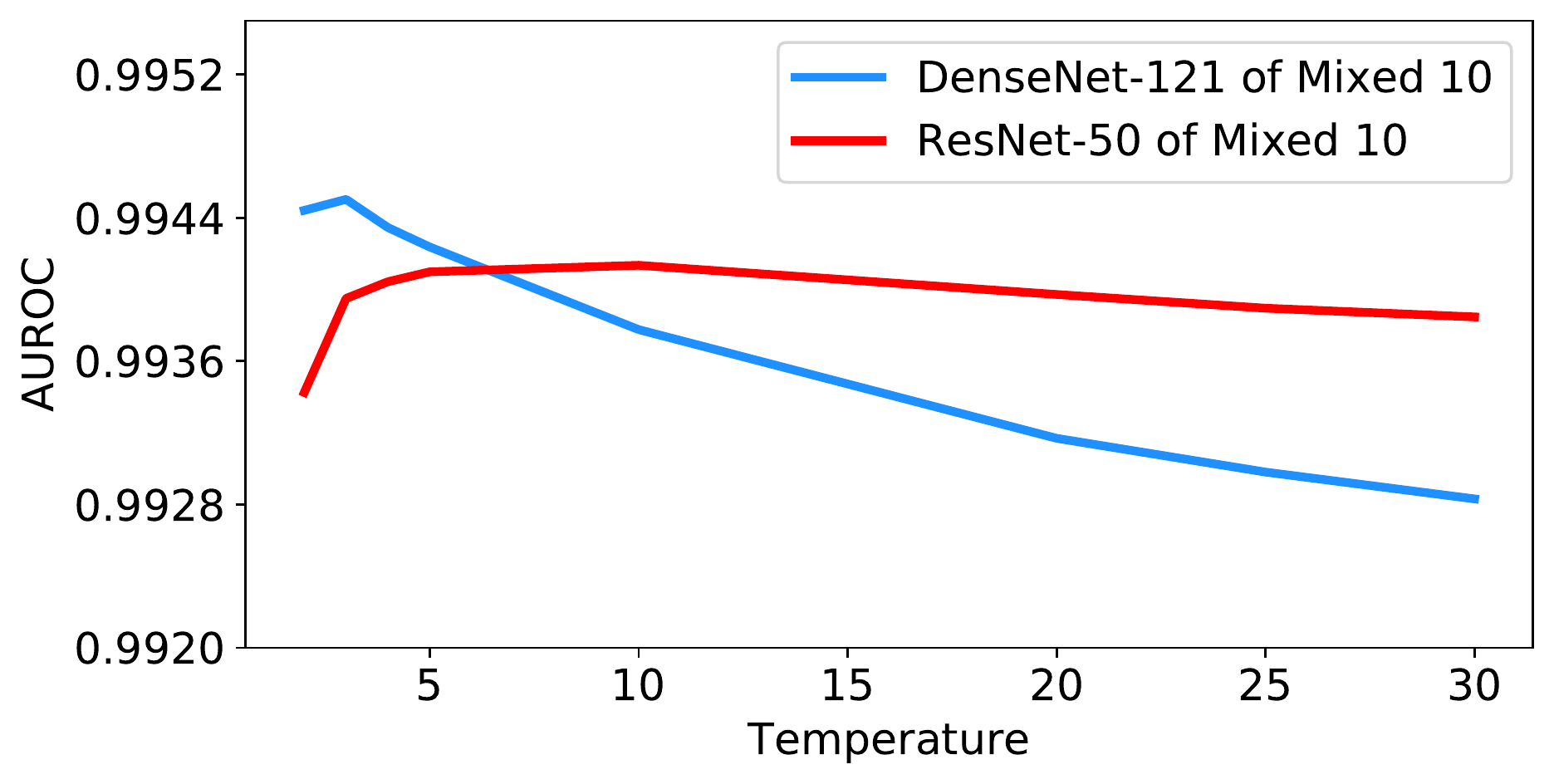}
		\caption{\label{fig:tempsensitivity3}Detection AUROC of Mixed 10 setting with different temperature.}
	\end{minipage}
\end{figure}

See Figure~\ref{fig:tempsensitivity2} and Figure~\ref{fig:tempsensitivity3}. Notice that for the \textit{ImageNet vs. Artificial} setting, the dimension of the output is 1000. When the class number is big, temperature effect is enlarged a lot thus large temperature value leads to a relatively fast decreasing in the AUROC result(A brief explanation: Since the maximum output is much bigger than others, when calculating softmax, a large class number will lead to a big denominator value in the softmax equation). And for those whose output dimension is under 20, we can see that the AUROC results are not sensitive to temperature value. 

%\begin{figure}[htbp]
%	\centering
%		\includegraphics[width=3.5in]{figure/temp/temp_effect_img_ani.pdf}
%		\caption{\label{fig:tempsensitivity1}Detection AUROC of ImageNet vs Anime dataset with different temperature.}
%\end{figure}

\end{document}